\documentclass[twoside,11pt]{article}
\usepackage[preprint]{jmlr2e}
\usepackage{url}            
\usepackage{booktabs}       
\usepackage{nicefrac}       
\usepackage{microtype}      
\usepackage{xcolor}         
\usepackage{graphics}       
\usepackage{graphicx}
\usepackage{amsmath}
\usepackage{amsfonts}       
\usepackage{mathrsfs}
\usepackage{mathtools}
\usepackage{natbib}
\usepackage{amsthm}
\usepackage{algorithm, algpseudocode}
\usepackage[font=small, labelfont=bf]{caption}
\usepackage[font=small, labelfont=bf]{subcaption}
\usepackage[bottom]{footmisc}

\usepackage{hyperref}
\hypersetup{colorlinks, citecolor=blue,linkcolor=red}

\theoremstyle{exampstyle} 
\newtheorem{theorem}{Theorem}
\newtheorem{lemma}[theorem]{Lemma}
\newtheorem{proposition}{Proposition}
\newtheorem{assumption}{Assumption}

\newtheorem{condition}{Condition}
\newtheorem{remark}{Remark}

\mathchardef\mhyphen="2D
\DeclareMathOperator*{\argmin}{argmin}
\DeclareMathOperator*{\argmax}{argmax}

\DeclareMathOperator*{\minsubsub}{\vphantom{argmin}min}

\newcommand{\newinf}{\mathop{\mathrm{inf}\vphantom{\mathrm{sup}}}}

\makeatletter
\def\@fnsymbol#1{\ensuremath{\textcolor{black}{\ifcase#1\or *\or \dagger\or \ddagger\or
   \mathsection\or \mathparagraph\or \|\or **\or \dagger\dagger
   \or \ddagger\ddagger\else\@ctrerr\fi}}}
\makeatother

\usepackage{lastpage}

\ShortHeadings{Stage-Aware Learning for Dynamic Treatments}{}
\firstpageno{1}

\usepackage[toc,page,header]{appendix}
\usepackage{minitoc}

\begin{document}

\doparttoc 
\faketableofcontents 

\title{Stage-Aware Learning for Dynamic Treatments}

\author{\name Hanwen Ye \email hanweny@uci.edu \\
       \addr Department of Statistics\\
       University of California\\
       Irvine, CA 92617, USA
       \AND
       \name Wenzhuo Zhou \email wenzhuz3@uci.edu \\
       \addr Department of Statistics\\
       University of California\\
       Irvine, CA 92617, USA
       \AND
       \name Ruoqing Zhu \email rqzhu@illinois.edu \\
       \addr Department of Statistics\\
       University of Illinois\\
       Urbana-Champaign, IL 61820, USA 
       \AND
       \name Annie Qu \email aqu2@uci.edu \\
       \addr Department of Statistics\\
       University of California\\
       Irvine, CA 92617, USA
}

\editor{}

\maketitle

\begin{abstract}
        Recent advances in dynamic treatment regimes (DTRs) facilitate the search for optimal treatments, which are tailored to individuals' specific needs and able to maximize their expected clinical benefits. However, existing algorithms relying on consistent trajectories, such as inverse probability weighting estimators (IPWEs), could suffer from insufficient sample size under optimal treatments and a growing number of decision-making stages, particularly in the context of chronic diseases. To address these challenges, we propose a novel individualized learning method which estimates the DTR with a focus on prioritizing alignment between the observed treatment trajectory and the one obtained by the optimal regime across decision stages. By relaxing the restriction that the observed trajectory must be fully aligned with the optimal treatments, our approach substantially improves the sample efficiency and stability of IPWE-based methods. In particular, the proposed learning scheme builds a more general framework which includes the popular outcome weighted learning framework as a special case of ours. Moreover, we introduce the notion of stage importance scores along with an attention mechanism to explicitly account for heterogeneity among decision stages. We establish the theoretical properties of the proposed approach, including the Fisher consistency and finite-sample performance bound. Empirically, we evaluate the proposed method in extensive simulated environments and a real case study for the COVID-19 pandemic.
\end{abstract}


\section{Introduction}
\label{sec:intro}

There has been great interest and demand for individualized modeling and personalized prediction, with applications ranging from medicine to education programs and marketing \citep{goetz2018personalized, tetzlaff2021developing, vesanen2006building}. For instance, the outbreak of COVID-19 in recent years has highlighted the growing demand for developing an effective time-varying treatment which can be tailored to individual patients \citep{jin2020gender, balzanelli2022new}. Dynamic treatment regime (DTR) \citep{tsiatis2019dynamic}, as an emerging individualized treatment strategy in multi-stage decision-making, has thus found much attention in the medical field. In contrast to traditional, one-size-fits-all medical treatments, DTRs continuously adapt a patient's treatment plan based on their response to previous treatments and changes in their medical condition. The main goal in the precision medicine research is to estimate the optimal treatment regime which maximizes expected long-term benefits for each patient \citep{rubin1974estimating, robins1986new}.

However, estimation of an optimal regime in practice is not a straightforward task, as there are often limited clinical data, complex heterogeneity among patients, and the number of treatment sequences grows exponentially with the number of decision points. Current estimation algorithms aim to tackle these empirical challenges and can be typically categorized into two main frameworks: indirect and direct value-search. The indirect methods, such as  Q-learning \citep{watkins1992q,nahum2012q}, A-learning \citep{AL-murphy, AL-blatt, shi2018high}, and tree-based methods \citep{laber2015tree, tao2018tree} primarily model the conditional distribution of a clinical outcome given the patients' past health status and treatment information, and choose the treatment that maximizes the modeled outcome as optimal. However, as the selection process follows backward induction, an optimal regime might not be recovered if one of the outcome models is not correctly specified \citep{QALReview}. The situation only gets worse in a chronic disease setting which involves a long sequence of decision stages. Though one can formulate the outcome model via a semi- or non-parametric approach \citep{NPQL-1, NPQL-2, zhao2009reinforcement} to allow more model flexibility and mitigate the risk of model-misspecification, the fitted models oftentimes are hard to interpret and thus less appealing for clinicians to apply.

On the other hand, value search methods include outcome weighed learning (OWL) \citep{OWL, SOWL}, residual weighted learning (RWL) \citep{zhou2017residual}, robust estimators \citep{zhang2012estimating, zhang2012robust,zhang2013robust, zhao2019efficient, schulz2021doubly}, non-parametric Q-learning based policy searches \citep{zhang2015using, zhang2018c, zhang2018estimation}, distributional learning \citep{mo2021learning}, and angle-based learning \citep{qi2020multi, xue2022multicategory}. These methods directly posit a class of DTRs and maximize the expected cumulative benefit within this posited class to estimate the optimal regime. However, performance can vary significantly depending on the importance sampling technique used to estimate the cumulative benefits. For instance, OWL targets to maximize the expected reward via inverse-probability weighted estimators (IPWEs), which is known to be unstable and sample inefficient when the number of treatment trajectories induced by the optimal regime is small in the dataset \citep{zhang2012estimating, zhang2012robust}. Unfortunately, in real-world applications, such a dilemma can be commonly found where the optimal treatment are under-studied for new diseases or inaccessible due to side effects or drug scarcity \citep{kao2008clinical,pawlik2005feasibility}. To improve the stability of the IPWEs, robust estimators and RWL augment mean-zero terms or estimate outcome residuals to capture information from non-optimal treatments and reduce variance. Alternatively, non-parametric Q-learning based policy searches combines the regression-based framework with the reward importance weighting to avoid IPWE instability. While these approaches address the stability issues associated with IPWEs, and models such as efficient augmentation and relaxation learning (EARL) \citep{zhao2019efficient} aim to improve computational efficiency, they still require additional estimations of the augmented outcomes or residuals. This introduces an additional computational burden not present in IPWEs, which becomes more pronounced as the number of decision steps in the regime increases.

In this paper, we propose a novel DTR estimation method, namely Stage Aware Learning (SAL). This method relaxes the main source of sample inefficiency and instability in IPWEs, i.e., the strict alignment requirement between observed and optimal treatment trajectories at all decision stages, which we refer to as the \textit{curse of full-matching}. The key idea is that treatment mismatches between the observed and the optimal regime are allowed. Specifically, instead of only seeking optimal regimes among patients assigned to optimal treatments at all decision stages, our approach includes patients treated under all strategies, and emphasizes those whose observed treatments align more closely with the optimal ones. Furthermore, to better capture the difference in treatment effectiveness at varying stages, we introduce the notion of stage importance scores and further propose the Stage Weighted Learning (SWL) method based on SAL, taking stage heterogeneity into account.

The main contributions of our paper are summarized as follows. First, to the best of our knowledge, our work is among the first to break the \textit{curse of full-matching}. By incorporating all clinical samples into the treatment rule estimation regardless of how well their assigned treatments align with the optimal regime, we bridge the gap between the DTR algorithms and practical challenges. Specifically, in settings with limited sample sizes and large number of decision stages, expecting all decisions to be optimal is unrealistic, but encountering one or more non-optimal decisions at some stages is more likely. In addition, unlike robust estimators, RWL, and regression-based methods, our method builds directly on IPWEs without the need for additional outcome models. These result in a more straightforward optimization, improved sample efficiency, and more stable IPWE estimation.

Second, our approach provides a more general framework which, for the first time, combines a number of IPWEs in value search algorithms to estimate optimal regimes. The flexibility of weighing each IPWE greatly increases the generalizability of DTR estimation methods to various treatment matching scenarios. In particular, the general framework includes the popular IPWE-based OWL and its variants as special cases of our approach. Furthermore, we propose stage importance scores along with an attention-based estimation procedure, which naturally inherits interpretability in non-parametric function approximation scenarios, to additionally account for stage heterogeneity and facilitate DTR estimation for our developed method.

Third, our work builds the theoretical connection between multi-stage DTR search problems and multi-label classification problems \citep{dembczynski2012label}. This simplifies the multi-stage optimal regime estimation procedure into a single-stage maximization problem. Specifically, we thoroughly investigate the theoretical properties of the proposed algorithms, including the Fisher consistency and the finite-sample performance error bound. Notably, our theoretical results work for both parametric and non-parametric model classes, and are comparable with the fastest convergence rates in the existing literature.

The remainder of the paper is structured as follows. In Section \ref{sec:background}, we introduce the notations and background of IPWEs. In Section \ref{sec:methodology}, we propose a novel $k$-IPWE estimator and introduce the SAL and SWL methods to account for stage heterogeneity. Section \ref{sec:theory} presents the Fisher consistencies and finite-sample performance error bound of the proposed SWL method, and Section \ref{sec:algo} explains the implementation details. In Section \ref{sec:simulation}, extensive simulation results are presented to illustrate the empirical performance advantages of our proposed methods. In Section \ref{sec:data}, we apply the proposed methods to the COVID-19 data from UC hospitals. Lastly, we conclude with discussion in Section \ref{sec:discussion}.

\section{Background} \label{sec:background}
In this section, we introduce the necessary notations and assumptions used in the paper, and formulate the multi-stage DTR estimation procedure under the IPWE \citep{horvitz1952generalization, robins1994estimation} framework. In addition, we describe the strict full-matching requirement induced by the IPWEs and provide a brief overview of related works which aim to enhance the stability of IPWEs.

\subsection{Notation and Preliminary}

Consider a balanced multistage decision setting where all patients in the study have a total number of T stages (visits). For each patient at their $j^{th}$ clinic visit, where $j=1,\ldots,T$, a set of time-varying variables $X_j \in \mathcal{X}_j$ are recorded to collect individual health status. Consequently, a new treatment assignment $A_j$ is delivered based upon the patient's longitudinal historical information from their first visit to the $j^{th}$ visit, denoted by $H_j = S_j(X_1, A_1, X_2, A_2,\ldots, A_{j-1}, X_j) \in \mathcal{H}_j$, where $S_j$ is a deterministic summary function. In this paper, we consider binary treatments, i.e., $A_j \in \mathcal{A} = \{1, -1\}$, where the interpretation of the treatment options are nested depending on the treatments assigned in previous stages. After the final visit, a clinical outcome $R=\sum_{j=1}^T r_{j}$, also known as the total reward from all immediate rewards, is obtained to reflect the benefits of the allocated treatment assignment.

A DTR is a sequence of decision rules $\mathscr{D} = \{D_j: \mathcal{H}_j \mapsto \mathcal{A}\}_{j=1}^T$ that map patients' historical information onto treatment space. The decision rules $\mathscr{D}$ could also be represented by a composite function of the real-valued functions $\mathbf{f} = \{f_j \in \mathcal{F}: \mathcal{H}_j \mapsto \mathbb{R}\}_{j=1}^T$, where a realization of the decision that follows the treatment regime at the $j^{th}$ visit is $d_j = D_j(H_j) = \text{sign}\{f_j(H_j)\}$. Now, assume that the full trajectory of an observation sequence $\{X_1, A_1, X_2, A_2,\ldots, X_T, A_T, R\}$ follows a data distribution $P$. Our goal is to seek the optimal treatment regime $\mathscr{D}^{*}$ which yields the largest expected rewards among all regimes:
\begin{equation}
    \mathscr{D}^{*} \in \argmax_\mathscr{D}{\mathbb{E}^\mathscr{D}\{R\}}.
    \label{optimalDTR}
\end{equation}
Note that the expectation operator $\mathbb{E}^\mathscr{D}$ in equation \eqref{optimalDTR} is taken with respect to an unknown restricted distribution $\{X_1, A_1=d_1, X_2, A_2=d_2,..., X_t, A_{T}=d_T, R\} \sim P^\mathscr{D}$, which describes the probability distribution when the treatments are assigned according to the regime $\mathscr{D}$. By convention, we call the corresponding $\mathscr{D}$ the target regime. Since the historical information and potential outcome under $P^{\mathscr{D}}$ are unobservable, to infer the decision rules from the observed data while avoiding confounding issues between the assignments and expected rewards, we adopt the following three standard assumptions: 

\begin{assumption}
(Stable Unit Treatment Value Assumption, SUTVA \citep{rubin1980randomization}) Potential outcomes of each subject are not affected by the treatments assigned to other subjects (no interference); there is no different form of each treatment level (no hidden variations)
\label{A-SUTVA}
\end{assumption}

\begin{assumption}
(Sequential Ignorability \citep{robins1986new}) Treatment assignments are independent of potential future outcomes, conditional on the history up to the current time $j$, i.e., $R_{j:T} \perp A_j  \;|\;H_j$
\label{A-seq}
\end{assumption}

\begin{assumption}
(Positivity) Any subjects are possibly assigned for all treatments, i.e., $P(A_j=a_j \;|\;H_j) > 0$ for any $a_j \in \mathcal{A}$ and $1 \le t \le T$.
\label{A-pos}
\end{assumption}

\subsection{Inverse Probability Weighted Estimators}

Under the aforementioned assumptions, it can be shown that the expected total reward under the target regime $\mathscr{D}$ is estimable by inverse probability weighting \citep{qian2011performance}:
\begin{equation}
\mathscr{D}^{*} = \argmax_{\mathscr{D}} \mathbb{E}^\mathscr{D}_{\text{IPW}}\{R\} =\argmax_{\mathscr{D}} \mathbb{E}\left\{ \frac{R \cdot \prod_{j=1}^T \mathbb{I}(A_{j} = D_j(H_j))}{\prod_{j=1}^T \pi_j(A_j | H_j)}\right\},
\label{eq:VF-IPWE}
\end{equation}
where $\pi_{j}$ is the propensity score function and the density corrected expected reward is referred to as the IPWE. Provided that the propensity functions are correctly specified, the optimal treatment regime $\mathscr{D}^*$ is the maximizer of the IPWE. Accordingly, learning schemes that directly estimate the optimal regime by maximizing the IPWE estimator could be categorized as IPWE-based approaches \citep{OWL,SOWL,laha2024finding}. 

However, due to the density ratio corrections, IPWE-bsaed approaches only incorporate a reward when an individual's treatments are fully matched with the target regime. Such a full-matching requirement can be reflected in the conditional expectation form of the weigthed total rewards,  
\begin{equation}
\small
\mathbb{E}_{\text{IPW}}^{\mathscr{D}}\{R\} =
 \underbrace{\mathbb{E}\left\{ \frac{R}{\prod_{j=1}^T \pi_j(A_j | H_j))} \;\middle|\;  \prod_{j=1}^T \mathbb{I}(A_{j} = D_j(H_j)) = 1\right\}}_{\text{Full-matching expected reward}} \cdot \underbrace{P \left(\prod_{j=1}^T \mathbb{I}(A_{j} = D_j(H_j)) = 1\right)}_{\text{Target regime assignment rate}},
    \label{eq:RN-cond}
\end{equation}
which is determined by two major factors: first, the expected reward among the population whose observed treatments are fully matched with the target regime $\mathscr{D}$, i.e., \textit{full-matching expected reward}; and second, the probability of assigning treatments conforming with the target regime at all stages, i.e., \textit{target regime assignment rate}. At the population level, the probability of assigning any arbitrary dynamic regime is ensured to be non-zero due to the \textit{ positivity} assumption. However, if the optimal regime assignment rate is small among the patient population, it is highly possible that none of the sampled patients may fully follow the optimal treatments at all stages, and thus the optimal regime is infeasible in practice. Clearly, this strict full-matching requirement may lead to difficulties on both optimization and estimation. We call this phenomenon the \textit{curse of full-matching}. 

\subsection{Curse of Full-Matching and Related Works}
\label{sec:curse-of-full-matching}

\begin{figure}[H]
    \centering
    \includegraphics[width=0.8\linewidth]{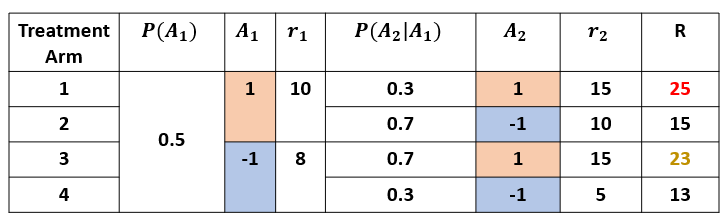}
    \caption{An example to illustrate the curse of full-matching. We consider a sequential randomized trial with a static treatment regime and constant rewards. The assignment rates $P(A_1)$ and $P(A_2|A_1)$ specify the sequential probability of allocating treatments at corresponding decision stages. The total reward $R$, which is the sum of stage immediate rewards $r_1$ and $r_2$ (i.e., $R=r_1+r_2$), evaluates the performance of each treatment arm.}   
\label{fig:treatment_arm}
\end{figure}

To fully understand the dilemma of full-matching, we frame this challenge within the context of a randomized trial and provide a concrete illustrative example presented in Table \ref{fig:treatment_arm}. Similar to observational studies, where the behavior policy can rarely keep making optimal decisions at all stages, this example assigns the optimal regime, Treatment Arm 1, at a small rate of 0.15 (i.e., $P(A_1) \cdot P(A_2|A_1) = 0.5 \cdot 0.3 = 0.15)$. Due to the requirement of \textit{full-matching}, IPWEs only focus on identifying these 15\% of subjects assigned to Treatment Arm 1 during empirical estimations. As a result, it might require many more samples to be collected before IPWE-based approaches could estimate the optimal regime and achieve near-asymptotic properties. Furthermore, note that the above example consists of only two decision stages. In practice, a growing number of stages tends to make estimation even harder since the target regime assignment rate decreases exponentially. 

\vspace{-0.25em}

\renewcommand{\figurename}{Table}
\renewcommand{\figurename}{Figure}

\setcounter{figure}{0}
\setcounter{table}{1}

\vspace{-0.25em}

Among existing literature, methods are developed to stabilize IPWEs from this strict full-matching condition. For instance, the robust estimators \citep{zhang2012estimating, zhang2012robust,zhang2013robust, zhao2019efficient, schulz2021doubly} augment IPWEs with an unbiased term  to achieve double-robustness. This provides a safeguard when estimating the propensity of the optimal regime becomes unstable due to the \textit{curse of full-matching}. RWL \citep{zhou2017residual}
estimates outcome residuals to stabilize IPWEs against outcome shifts. Q-learning based policy searches, such as C-learning \citep{zhang2018c}, leverage the advantages of regression-based frameworks, which do not rely on probability weighting and circumvent the \textit{curse of full-matching}. However, these methods all require accurate outcome models for their additional augmentation terms or regression components. Given the high heterogeneity among rewards and an increasing number of decision stages, correctly specifying the outcome model poses significant challenges and increases computational complexity. 

Our work seeks to address the challenges posed by the \textit{curse of full matching} while maintaining the simplicity of IPWEs and avoiding the specification issues commonly associated with outcome models. Specifically, referring back to the example in Table \ref{fig:treatment_arm}, we aim to leverage the substantial information available in Treatment Arms 2 and 3, which differ from the optimal regime by only one treatment but are more prevalent in the trial with a combined probability of 70\% (i.e., $P(\text{arm}2 \text{ or } \text{arm}3) = 0.5 \cdot (0.7 + 0.7) = 0.7$). Notably, the total rewards from these non-optimal treatment arms, especially Treatment Arm 3, are not significantly different from the optimal arm. Inspired by this observation, we propose a novel method which can incorporate these non-optimal treatments in the estimation of the optimal regime and break the \textit{curse of cull-matching} to improve sample efficiency.


\section{Methodology} \label{sec:methodology}

Our work is motivated by solving the \textit{curse of full-matching} challenges. In this section, we study a $k$-partially matching estimator and propose a new DTR learning method, namely Stage-Aware Learning (SAL), which allows treatment mismatches with the target regime as a resolution to improve sample efficiency and estimation stability. Furthermore, to additionally account for the heterogeneity of treatment effects at different decision stages, we introduce the Stage Weighted Learning (SWL) method as a variant of the SAL method.

\subsection{The $k$-partially Matching Estimator: $k$-IPWE}
\label{sec:K-IPWE}

The performance of IPWE-based approaches is largely restricted by the full-matching assignment rate of the optimal regime among the patient populations. Though we might not control how treatments are administrated according to the optimal regime at every decision stage, in this subsection, we propose to relax the strict full-matching requirement by allowing decision discrepancies between the assigned treatments and the target regime $\mathscr{D}$ at $T-k$ number of stages, where $0 \le k \le T$. In other words, the optimal regime is allowed to partially match the treatment sequence at exactly $k$ number of arbitrary decision stages. The rationale is that, while it is unrealistic to expect all decisions to be optimal, encountering one or more non-optimal decisions at some stages is more likely in practice. As a result, the probability of patients receiving optimal decisions at $k$ number of stages could be much larger than the probability of patients receiving optimal decisions at all stages. We call this relaxed version of the full-matching requirement the \textit{k-partially matching requirement}.

To formalize the notation, we let random variable $K$ denote the number of correct alignments between treatments $\mathbf{A} = \{A_j\}_{j=1}^T$ and decisions from an arbitrary target regime,
\begin{align}
    K \doteq \lvert\mathbf{A} \cap \mathscr{D}\rvert &= \sum_{j=1}^T \mathbb{I}(A_j = D_j(H_j)).
\label{eq:k-def} 
\end{align}
At each stage, we examine whether the assigned treatment aligns with the recommended treatment. Importantly, the recommended treatment is determined based on the patient's historical information, which makes treatments nested and their effects carried over the decision stages. Additionally, if we specify $K$ to be a realized value $k$ within the range $\{0,\ldots, T\}$, we are constraining our optimal regime search to the patient population who are $k$-partially matched with the optimal regime. A more extreme $k$ value (e.g., $k=0$ or $k=T$) corresponds to a more restricted alignment requirement. Under the IPWE framework, only patients with $K=T$ are included in the estimation procedure.

When $K=k$, a new restricted unknown distribution is induced, i.e., $(X_1, A_1 = \Tilde{d}_1, ..., X_T, A_T = \Tilde{d}_T, R) \sim P^{\mathscr{D}_{(k)}}$, where $\Tilde{d}_j = (-1)^{\mathbb{I}(j \in \mathcal{K}) + 1} \cdot d_j$ indicates that at any stage $0 \le j \le T$, the decision $\Tilde{d}_j$ is the same as the target regime $d_j$ only if $j$ is among the indexes of $k$ arbitrary matching stages $\mathcal{K}$ (i.e., $\Tilde{d}_j = d_j\;\text{if}\; j \in \mathcal{K}$). Subsequently, $P^{\mathscr{D}_{(k)}}$ is a distribution of an observation sequence where its $k$ out of T assignments $\{A_j\}_{j \in  \mathcal{K}}$ are followed by the regime $\mathscr{D}$. Based on the derivation of $k$-matching potential outcomes provided in Appendix \ref{S:K-est}, we can similarly adopt the density ratio correction and obtain an IPWE for the expected rewards evaluated under the new measure $P^{\mathscr{D}_{(k)}}$. We denote the new estimator $\mathbb{E}^{\mathscr{D}_{(k)}}[R]$ as $k$-IPWE and present the results in Proposition \ref{eq:k-IPWE}:

\begin{proposition}
    Under Assumptions \ref{A-SUTVA}-\ref{A-pos}, the expected total reward under the target regime $\mathscr{D}$ with k number of matching stages equals
    \begin{equation}
\mathbb{E}^{\mathscr{D}_{(k)}}[R] =  \mathbb{E}\left\{ \frac{R \cdot \mathbb{I}(\lvert\mathbf{A} \cap \mathscr{D}\rvert = k)}{\prod_{j=1}^T \pi_j(A_j | H_j)}\right\},
    \end{equation}
    and the corresponding maximizing regime $\Tilde{\mathscr{D}}_{(k)}$ is defined as 
    \begin{equation}
    \Tilde{\mathscr{D}}_{(k)} = \argmax_{\mathscr{D}} \mathbb{E}^{\mathscr{D}_{(k)}}\{R\}.
    \label{eq:k-VF}
    \end{equation}
\label{eq:k-IPWE}
\end{proposition}

The regime $\Tilde{\mathscr{D}}_{(k)}$ maximizing the $k$-IPWE would yield the largest expected reward if patients were treated by $\mathscr{D}$ at $k$ number of stages. In addition, note that we do not require the $k$  matching stages to be the same for each individual. As long as there is an exact $k$ number of treatment matchings between the assignments and target regime $\mathscr{D}$, those patients' rewards are involved in the maximization process. As a result, the $k$-IPWE provides a superior level of flexibility. 

However, the performance of the proposed $k$-IPWEs still depends on the pre-selected $k$ value. The purpose of designing the $K$ treatment matching number is to increase the conditional probability of the constrained population receiving optimal treatments. When the value $k$ is poorly selected, the probability of patients being $k$-partially matched could be small, and we could encounter a similar aforementioned empirical dilemma. For instance, in a population with a 99\% full-matching assignment rate, specifying the random variable $K$ with values other than $T$ leads to a small $k$-partially matching rate. In other words, the \textit{curse of full-matching }can be effectively minimized only if the pre-specified $k$ has the highest $k$-partially matching probability, i.e., $k = \argmax_{k \in \{0,\ldots,T\}} P(K=k)$. Finding such a $k$ value is possible but computationally cumbersome, and yet not every patient will participate in the optimization process due to the population conditional constraints. To reduce the uncertainty of selecting $k$ values and include all individuals from the sample, we further construct a learning method based on $k$-IPWE which can incorporate all scenarios of $k$-partially matching through applying a weighting scale on the matching number $K$.

\subsection{Stage-Aware Learning Method (SAL)}
\label{sec:SAL}
In this subsection, we formally introduce a novel learning method to combine all levels of $k$-IPWEs into one single estimation task. This addresses the dependencies of $k$-IPWEs at the pre-selected $k$-values. Since the new estimator accounts for treatment and regime matching status at any number of stages, we name the new learning method Stage-Aware Learning (SAL).

To start with, we re-weight each $k$-IPWE by $k/T$, proportional to the number of matching stages $k$, and estimate the optimal regime simultaneously by maximizing the following SAL value function derived in Appendix \ref{S:SAL},
\begin{equation}
     V^{SA}(\mathscr{D})=  \sum_{k=0}^T   \frac{k}{T} \cdot \mathbb{E}^{\mathscr{D}_{(k)}} \{R\} = \mathbb{E} \left\{ \frac{R \cdot \frac{1}{T} \sum_{j=1}^T \mathbb{I}(A_j = D_j(H_j))}{\prod_{j=1}^T \pi_j(A_j | H_j)}\right\}.
\label{eq:SAL-linear}
\end{equation}
We denote the estimated maximizing regime $\Tilde{\mathscr{D}} = \argmax_{\mathscr{D}} V^{SA}(\mathscr{D})$. In our choice, the applied weights increase with the $k$ value, and imply that we prioritize the regime which has closer alignment with the optimal decisions while achieving higher expected rewards during estimation. Structurally, the weighting component of the final SAL value function resembles the formulation of the Hamming loss \citep{tsoumakas2007multi}, which has the following three unique advantages.

First of all, compared to the IPWE value function \eqref{eq:VF-IPWE}, the SAL value function replaces the product of indicator functions with the correct treatment alignment percentage. Therefore, instead of excluding patients completely if one of their observed treatments is not aligned with the optimal decision, SAL still considers those patients but discounts their rewards based on the degree of alignment between observed and optimal treatments across all of the decision stages. That is, even if the decision sequence is long, the new learning process is able to utilize all available patients' outcomes and maximize the alignment percentages over those with high rewards. As a result, all patients and their treatment strategies are included in the optimal regime searching procedure. Second, the SAL learning scheme is analogous to a multi-label classification framework. Instead of employing a surrogate function involving all decision stages simultaneously in OWL \citep{SOWL}, our proposed regime can be optimized at each individual stage to match the optimal decisions. This leads to computational convenience as the optimization of the Hamming loss function has been well established and the multi-stage learning task can be segmented into single-stage sub-tasks.

In fact, SAL suggests a more general DTR learning framework by allowing a probability distribution on the matching number $K$. Particularly, the IPWE-based framework is a special case of ours under the general framework indicated in Remark \ref{remark:example}. Suppose the density function of $K$ is proportional to a scale function $\phi(\cdot)$.  After taking the iterated expectation of $k$-IPWEs under all possible choices of $k$ values, we obtain a new estimator of the expected rewards aimed to maximized for any arbitrary target regime $\mathscr{D}$,
\begin{align}
    \mathbb{E}_{K} \left \{\mathbb{E}^{\mathscr{D}_{(k)}} \{R\} \right\} &= \sum_{k=0}^T \phi(k) \cdot \mathbb{E}^{\mathscr{D}_{(k)}} \{R\}  = \mathbb{E} \left\{ \frac{R \cdot \phi\left( \lvert\mathbf{A} \cap \mathscr{D}\rvert \right)}{\prod_{j=1}^T \pi_j(A_j | H_j)}\right\}.
    \label{eq:EE}
\end{align} 

The new estimator results in a generic form where the total rewards are weighted by the density of $K$. Correspondingly, the maximization procedure not only finds the regime that yields the largest expected total rewards, but also identifies the one which matches with the observed treatment sequence closest to the underlying distribution of the treatment matching number. In addition, compared to the previous regime estimator $\Tilde{\mathscr{D}}_{(k)}$ based on the $k$-IPWE in equation \eqref{eq:k-VF}, the general estimator no longer constrains the rewards under a sub-population of patients. Instead, it takes every patient's reward into account as long as the density is positive for all matching number $K$, i.e., $\phi(k) > 0$ $\forall k \in \{1,..,T\}$.

\begin{remark}
The induced general framework allows a flexible specification of the density scale function $\phi(\cdot)$, which can be used to summarize our prior knowledge of how well the observed treatments can match the optimal regime given the $k$ matching number. For instance, the proposed SAL method adopts a linear scale function (i.e., $\phi(k) = k$) under our assumption that patients are more likely to receive a larger number of treatment assignments following the optimal regime. We include more discussions on this assumption in Appendix \ref{S:KIPWL}.
\end{remark}

\begin{remark}
The IPWE-based approaches can be recovered when a degenerated density function $\left(\phi(k) = \mathbb{I}(k=T)\right)$ is specified. It assumes that the probability of the assigned treatments fully aligning with the target regime at all decision stages is equal to one. Consequently, only patients meeting the full-matching requirement can be counted in the regime estimation process, and the resulting maximizing regime $\Tilde{\mathscr{D}}$ is equivalent to the optimal treatment regime $\mathscr{D}^*$, which enjoys all of the pre-established theoretical results \citep{qian2011performance, SOWL}. Similarly, when $K$ follows a degenerated distribution $\phi(k) = \mathbb{I}(k=j)$ at other stages j ($0 \le j < T$), we can solve the general framework under the $k$-IPWE and obtain the maximizing regime $\Tilde{\mathscr{D}} = \Tilde{\mathscr{D}}_{(j)}$.
\label{remark:example}
\end{remark}

In summary, we propose a novel SAL method under a more general DTR estimation framework, to improve data efficiency and empirical adaptivity of IPWE-based methods. By combining each level of $k$-IPWEs, we include all matching scenarios and relax the selection of the $K$ value. Through imposing higher weights to $k$-IPWEs with larger $k$ value, SAL possesses the interpretation of searching samples with high total rewards and large optimal treatment matching percentages simultaneously without the need for stage immediate rewards. However, the absence of immediate rewards may complicate the learning process of treatment effects at each stage. It requires larger sample sizes for SAL to attribute variations in total rewards to a single stage, especially when dealing with individuals having similar matching percentages. Next, we further propose a weighted learning scheme based on SAL to incorporate stage heterogeneity and facilitate the DTR learning process.

\subsection{Stage Weighted Learning Method (SWL)}

In this subsection, we propose a non-trivial variant of SAL, namely the Stage Weighted Learning (SWL) method, based on a weighted multi-label framework to enhance stage heterogeneity and facilitate distributing stage-wise treatment effects from total rewards into individual stages,  We formulate the SWL value function as follows,
\begin{equation}
    V^{SW}(\mathscr{D}) \doteq \mathbb{E} \left\{ \frac{R \cdot \sum_{j=1}^T \omega_j \cdot \mathbb{I}(A_j = D_j(H_j))}{\prod_{j=1}^T \pi_j (A_j | H_j)} \right\},
\label{eq:SWL-VF}
\end{equation}
where $\{\omega_j\}_{j=1}^T$ are the real-valued weights satisfying $\omega_j \in [0,1]$ and $\sum_{j=1}^T \omega_j = 1$ for any $j$ in $\{1,..,T\}$. Intuitively, a larger weight is imposed on a stage with more substantial treatment effect contributing to the total rewards. As the weights are designed to quantify the relative importance of treatment effects among decision stages, we denote the imposed weights $\{\omega_j\}_{j=1}^T$ as stage importance scores.

With the incorporation of stage importance scores, a notable advantage of SWL is the rescaling of the reward by a weighted average of the treatment alignments. Compared to the SAL's uniform stage weights ($1/T$) applied to each treatment-matching stage outlined in equation \eqref{eq:SAL-linear}, the nonidentical stage importance scores introduce stage heterogeneity to the matching percentage component of the value function. Specifically, individuals with the same number of alignments between assigned treatments and optimal decisions no longer have identical treatment-matching percentages as seen in SAL. Instead, their matching percentages vary based on the importance scores assigned to the matched stages, i.e., a higher matching percentage is achieved when more stages with larger importance scores are matched. Consequently, in addition to the total rewards which capture the combined treatment effects across all decision stages, the matching percentages augment the stage-wise treatment effect through the importance scores, and thus further account for the stage heterogeneity within the SWL learning scheme.

The proposed stage importance scores also enhance the learning process of treatment regimes. As the total rewards are scaled by the weighted matching percentages in equation \eqref{eq:SWL-VF}, treatment mismatch at stages with large importance scores would lead to a substantial loss in total expected rewards; whereas the expected rewards only have minor fluctuations at stages with negligible importance scores, regardless of treatment assignments. Under the main objective of maximizing the expected total rewards, SWL consequently prioritizes improving treatment alignment accuracy at important stages, and the importance scores at the individual stage level can effectively direct SWL's attention towards those stages with substantial treatment effects.

\begin{figure}[hbt!]
\centering
\includegraphics[width=0.75\textwidth]{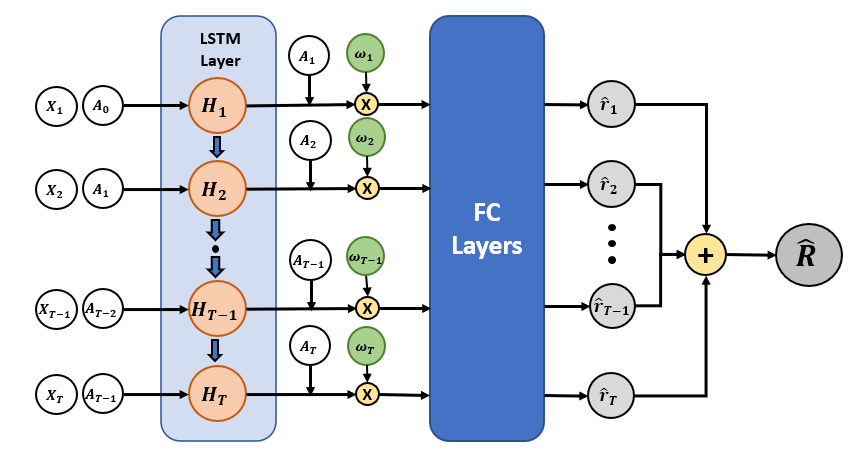}
\caption{Architecture of stage importance scores searching network. The stage importance scores are treated as the attention weights applied on the patients' historical information by the LSTM layer \citep{hochreiter1997long}, and are later estimated by minimizing the MSE between the observed and surrogate total rewards after the fully-connected (FC) layers transformation.} 
\label{fig:lstm_frameworkl}
\end{figure}

Noticeably, one of the key components in SWL is the stage importance score. However, estimating these scores is non-trivial, mainly because the immediate rewards used to evaluate the treatment effects at each stage in some cases are unobservable. To solve this challenge, we utilize the attention-based mechanism \citep{bahdanau2014neural}. Suppose there exists an underlying immediate reward function structure $\{\mathbf{\Tilde{r}}_j\}_{j=1}^T$ which takes the stage importance score as an additional parameter. Formally, it affects individual's immediate rewards at the $j^{th}$ stage as,
\begin{equation}
    \mathbf{r}_j(H_{ij}, A_{ij}) = \mathbf{\Tilde{r}}_j(\omega_j, H_{ij}, A_{ij}),
    \label{eq:homogeneous-reward}
\end{equation}
where $\mathbf{r}_j: \mathcal{H} \times \mathcal{A} \mapsto \mathbb{R}^+$ is the conventional definition of the immediate reward function. Under our assumption, the importance scores can be disentangled from the original rewards and are invariant to individual patients, representing stage-wise heterogeneity. Then, due to the fact that the expected total reward is the summation of expected immediate rewards, we estimate the importance scores scalars by minimizing the empirical squared loss between total rewards and constructed surrogate rewards from a semi-parametric point of view, i.e., 
\begin{equation}
    \{\hat{\omega}_1,...,\hat{\omega}_T\} = \argmin_{(\omega_1,...,\omega_T) \in \mathbb{R}_{[0,1]}^{|T|}} \minsubsub_{\vphantom{\mathbb{R}_{[0,1]}^{|T|}} (\mathbf{\Tilde{r}}_1, ..., \mathbf{\Tilde{r}}_T) \in \mathcal{R}^{|T|}} \frac{1}{n} \sum_{i=1}^n \left(R_i - \sum_{j=1}^T \mathbf{\Tilde{r}}_{j} \left(\omega_j, H_{ij}, A_{ij}\right) \right)^2.
    \label{eq:stage-importance-score-est}
\end{equation}
Note that we do not limit the parametric form of the reward function space $\mathcal{R}$. For illustration purposes, we represent each reward function $\Tilde{\mathbf{r}}_j$ with a fully-connected (FC) network due to its flexible capability of function approximation \citep{devore2021neural}, and adopt a long-short term memory (LSTM) network \citep{hochreiter1997long} to capture the unobserved patients' historical information $H_{ij}$ using up-to-date patients' covariate information $\{X_{it}\}_{t=1}^j$ and past treatments $\{A_{it}\}_{t=1}^{j-1}$. Finally, we leverage the attention mechanism and propose an attention-based recurrent neural network architecture in Figure \ref{fig:lstm_frameworkl} to estimate the stage importance scores.

The idea behind the attention mechanism is to scale an input sequence by relevance to the predicting outcomes, with a more relevant part being assigned a higher weight. These weights focus the attention of the prediction model on the most relevant part of the input sequence to improve model performance. In our scenario, we view the stage importance scores as the attention weights so that stages with higher contributions and more relevance to the final total rewards possess larger attention weights. Therefore, the weights not only explicitly impose stage heterogeneity on the proposed neural network, but also direct the network to pay more attention to the stages with large treatment effects when predicting the total rewards. Once we compute the surrogate total reward from $\Tilde{R} = \sum_{j=1}^T \Tilde{r}_j$, the final importance scores can be optimized by minimizing the loss function \eqref{eq:stage-importance-score-est}.

With estimated importance scores, we can search for the optimal treatment regime under the SWL value function \eqref{eq:SWL-VF}, via maximizing the objective function with a smooth convex surrogate function $\psi$ \citep{bartlett2006convexity}:
\begin{equation}
\hat{\mathscr{D}}^{SW}_\psi = (\hat{f}_{\psi 1},...,\hat{f}_{\psi T})^{SW} = \argmax_{\{f_1,...,f_T\} \in \mathcal{F}^{|T|}} \frac{1}{n} \sum_{i=1}^n \frac{R_i\sum_{j=1}^T \hat{\omega}_j \; \psi(a_{ij} \cdot  f_j(H_{ij}))}{\prod_{j=1}^T \pi_j(a_{ij} | H_{ij})}.
    \label{eq:SWL_obj_max_sur}
\end{equation}
In particular, we can employ the logistic function as a surrogate to the indicators, i.e., $\psi(x;\lambda)=(e^{-\lambda x}+1)^{-1}$. The hyper-parameter $\lambda$ controls the growth rate where a larger $\lambda$ value makes the surrogate converge to the 0-1 indicator function faster.

To conclude, the proposed SWL method inherits the property of treatment mismatching from SAL and further enhance stage heterogeneity via the empirical stage importance scores estimated from the attention mechanism. The resulting SWL scheme is able to estimate the sequential DTR under the weighted multi-label classification framework, where optimal regime learning at stages with substantial treatment effects are prioritized. In the next section, we will present the theoretical results of the proposed SWL method.

\section{Theoretical Results} \label{sec:theory}

In this section, we study the theoretical properties of the proposed SWL method. In Theorem \ref{thm:fisher-1}, we show the Fisher consistency of the SWL surrogate estimator  $\Tilde{\mathscr{D}}^{SW}_\psi$ compared to the SWL optimal regime $\Tilde{\mathscr{D}}^{SW}$. Next, we demonstrate the SWL Fisher consistency of the optimal regime $\mathscr{D}^*$ in Theorem \ref{thm: optimal-fisher}, and establish the finite-sample performance error bound with a flexible metric entropy in Theorem \ref{theorem:performance-bound}. To the best of our knowledge, Theorem \ref{thm: optimal-fisher} is the first theoretical result to fully discuss the gap in Fisher consistency between multi-stage DTR methods and the multi-label classification framework. The proof of the theorems and additional Lemmas are deferred to Appendix.

In the following, we first introduce our notation for the technical developments. We let $\mathcal{G}^{|T|}$ be a generic product function space; and denote $\mathbf{g}^* = (g^*_1,...,g^*_T) \in \mathcal{G}^{|T|}$ and  $\mathbf{g}^*_{\psi} = (g^*_{\psi 1},...,g^*_{\psi T}) \in \mathcal{G}^{|T|}$ as the optimal treatment regimes with respect to the SWL value function $V^{SW}$ \eqref{eq:SWL-VF} and its surrogate counterpart $V^{SW}_{\psi}$, respectively. Moreover, we define a parametric product function space $\mathcal{F}^{|T|}$, where we search the maximizer  $\mathbf{f}^* = (f^*_1,...,f^*_T) \in \mathcal{F}^{|T|}$ and  $\mathbf{f}^*_{\psi} = (f^*_{\psi 1},...,f^*_{\psi T}) \in \mathcal{F}^{|T|}$ for the function approximations for $\mathbf{g}^*$ and $\mathbf{g}^*_{\psi}$, respectively. Given the observed data, we further define $\hat{\mathbf{f}}_n = (\hat{f}_{\psi 1},...,\hat{f}_{\psi T}) \in \mathcal{F}^{|T|}$ as the empirical maximizer of the SWL objective function $\hat{V}^{SW}_\psi$ \eqref{eq:SWL_obj_max_sur}. Importantly, we require the stage importance scores have been estimated and remain fixed throughout the estimation of SWL. Lastly, before establishing the theoretical results, we present the following necessary regularity conditions in addition to the assumptions introduced in Section \ref{sec:background}.

\begin{assumption}
(Finite Reward) The total reward is positive and upper-bounded by a finite constant M, i.e., $0 \le \|R\|_{\infty} \le M < \infty$.
\label{A-reward}
\end{assumption}

\begin{assumption}
(Strong Positivity) The propensity score $\pi_j(A_j|H_j)$ is pre-defined and lower-bounded by a positive real number, $c_0$, s.t. $0 < c_0 \le \pi_j < 1$.
\label{A-propensity}
\end{assumption}

\begin{assumption}
(No Approximation Error) Suppose for any parameterized functional space $\mathcal{F}^{|T|}$, the approximation error $\epsilon_{app}$ satisfies the following:
\begin{equation}
    \epsilon_{app} := \sup_{\mathbf{g} \in \mathcal{G}^{|T|}} \newinf_{
\mathbf{f} \in \mathcal{F}^{|T|}} \|\mathbf{g} - \mathbf{f}\|_\infty = 0. \nonumber
\end{equation}
\label{A-approx}
\end{assumption}

Assumption \ref{A-reward} is a standard assumption, which requires the total reward to be positive and bounded. Assumption \ref{A-propensity} indicates that the probability of assigning any treatment to arbitrary stages is positive and lower bounded. In addition, we assume the propensity scores are known and pre-defined in a trial design to facilitate the development of theoretical results.  Assumption \ref{A-approx} defines an approximation error $\epsilon_{app}$ due to the difference between the parameterized space $\mathcal{F}^{|T|}$ and the generic space $\mathcal{G}^{|T|}$. By setting $\epsilon_{app}$ to zero, the assumption states that for any function sequence $\mathbf{g}$ in the generic function space $\mathcal{G}^{|T|}$, we can find a function sequence $\mathbf{f}$ in the parameterized space $\mathcal{F}^{|T|}$ such that $\mathbf{f} = \mathbf{g}$. Equivalently, it can be shown that the optimal regime $\mathbf{g}^*$ belongs to $\mathcal{F}^{|T|}$ and $\mathbf{g}^* = \mathbf{f}^*$.

\subsection{Surrogate Fisher consistency}
\label{sec:fisher-1}
The adoption of surrogate functions eases the optimization procedure. In this subsection, we establish the Fisher consistency between the value function $V^{SW}$ and its surrogate form $V^{SW}_\psi$. Specifically, we show that the optimal surrogate treatment decision $\text{sign}(f^*_{\psi j})$ is aligned with the optimal decision $\text{sign}(f^*_{j})$ at each stage. The obtained result is presented in Theorem \ref{thm:fisher-1}.

\begin{theorem}
Let $\psi(a, f; \lambda): \mathcal{A} \times \mathcal{F} \times \Lambda \mapsto \mathbb{R}$ be a surrogate function with tuning parameters $\lambda$ that satisfies $\psi(a, f; \lambda) = \psi(-a, -f; \lambda)$ and $sign(\psi(1, f; \lambda) - \psi(-1, f; \lambda)) = sign(f)$. Then, for all $t=1,...,T$ and $H_t \in \mathscr{H}_t$,
\begin{equation}
    \text{sign}(f^*_{\psi t}(H_{t})) = \text{sign}(f^*_t(H_{t})) = \argmax_{a_t \in \{-1,1\}} \mathbb{E}\left\{r_t + \sum_{j=t+1}^T r_j \;\middle|\; A_t=a_t, H_t \right\}.
\label{eq:fisher-1}
\end{equation}
\label{thm:fisher-1}
\end{theorem}

Theorem \ref{thm:fisher-1} guarantees the same treatment decisions could be obtained from the surrogate estimators and the maximizing regime under the SWL scheme. This validates the usage of smooth surrogate functions to approximate the indicator functions at each decision stage which cannot be solved under the IPWE framework. In addition, according to Lemma \ref{lemma:fisher} in Appendix \ref{S:Thm-1}, the surrogate is only required to be an even function and produce the same sign as the treatment effect. In fact, a wide class of surrogate functions, such as indicators, logistic functions, and binary cross-entropy, satisfies such requirements and increases the optimization flexibility of the proposed SWL method.

In the following subsection, we also demonstrate the Fisher consistency between SWL and the optimal DTR $\mathscr{D}^* = \{\mathcal{D}_j^*\}_{j=1}^T$. The optimal decision $d^*_t$ on arbitrary stage $t$ from equation \eqref{optimalDTR} has the form of 
\begin{equation}
    d^*_t = \mathcal{D}^*_t(H_t) = \argmax_{a_t \in \{-1, 1\}} \mathbb{E} \left\{ r_t + \sum_{j=t+1}^T r_j \;\middle|\; A_t=a_t, H_t, A_{(t+1):T} = d^*_{(t+1):T}\right\}.
\label{eq:optimal-fisher}
\end{equation}
Compared to SWL maximizing regime $\mathbf{f}^*$ in equation \eqref{eq:fisher-1}, the optimal DTR $\mathscr{D}^*$ also aims to maximize expected future reward, but further assumes every future treatment step matches with the optimal decision. Thus, depending on the underlying behavioral distributions of rewards and treatment assignments, SWL could recover $\mathscr{D}^*$ asymptotically only under much more stringent conditions.

\subsection{Fisher consistency: optimal treatment dominance}
\label{sec:FC-2}
To fill in the gap between SWL and optimal DTR, we investigate the boundary condition where the SWL estimators might produce different treatment decisions from the optimal regime. This condition can be quantified via the relationships between the expected optimal reward and potential loss due to sub-optimal treatments, defined as follows:
\begin{condition}
(Optimal Treatment Dominance) Any decision Stage $t$, where $1 \le t \le T$, is said to be dominated by the optimal treatment if the optimal treatment yields the largest net payoffs, i.e., $d_t^* = \argmax_{a_t \in \{-1,1\}} \textup{NetPayoffs}(A_t = a_t, H_t)$, where
\begin{equation}
\mathbf{NetPayoffs}(A_t, H_t) = \underbrace{R^*(A_t, H_t)}_{\text{Gain from optimal}} - \underbrace{(R^*(A_t, H_t) - R^{\dagger}(A_t, H_t)) \cdot P^{\dagger}(A_t, H_t).}_{\text{Cost: expected loss from sub-optimal future treatments}}
\end{equation}
\label{def:optimal-dominance}
\end{condition}
\noindent Specifically, $R^*(A_t,H_t)=\mathbb{E}\{R \;|\; A_t, H_t, A_{(t+1):T} = d^*_{(t+1):T}\}$ is the expected total reward if all future treatments are optimal, $R^{\dagger}(A_t,H_t)= \{\mathbb{E}\{R \;|\; A_t, H_t, A_{(t+1):T} = d^{\dagger}_{(t+1):T} \}$  represents the expected total reward if some future treatments failed to be assigned optimally, i.e., $d^{\dagger}_{(t+1):T}$ is a collection of all possible sub-optimal regimes where $d^{\dagger}_j \neq d^*_j$ at some stages $t < j \le T$, and $P^{\dagger}(A_t, H_t)=P(A_{(t+1):T} = d^{\dagger}_{(t+1):T} \;|\; A_t, H_t)$ denotes the probability of assigning any sub-optimal treatments in the future. Overall, Condition \ref{def:optimal-dominance} indicates optimal dominance if the underlying optimal treatment not only maximizes expected reward but also minimizes potential cost in cases where future treatments may deviate from the optimal treatments. Then, if Condition \ref{def:optimal-dominance} is satisfied at every decision stage, Fisher consistency between SWL and optimal DTR can be reached.

\begin{theorem}
For all stages $t=1,...,T$ and $H_t \in \mathscr{H}_t$, $\text{sign}(f_{\psi t}^*) = \text{sign}(f_{t}^*) = d_t^*$ if and only if Stage $t$ is dominated by the optimal treatment.
\label{thm: optimal-fisher}
\end{theorem}

As Theorem \ref{thm: optimal-fisher} shows, Fisher consistency depends on the dominance of the optimal treatments. For a better understanding of Condition \ref{def:optimal-dominance}, we begin with an extreme scenario where every individual receives the optimal decisions. Obviously, optimal decisions in this case are dominant as there is no other regime assigned, and Condition \ref{def:optimal-dominance} is satisfied at every stage, i.e., $P^{\dagger}(A_t, H_t) = 0$ and the optimal decision $d^*_t$ maximize the $R^*(A_t=d^*_t, H_t)$.  Furthermore, since the expected reward $\mathbb{E} \left\{R \;\middle|\; A_t, H_t \right\}$ is equal to the expected optimal reward $R^*(A_t, H_t)$, it is straightforward to show that the SWL maximizing regime $\mathbf{f}^*$ is the same as the optimal DTR $\mathbf{d}^*$ according to Equations \eqref{eq:fisher-1} and \eqref{eq:optimal-fisher}.

In a more general sense, Condition \ref{def:optimal-dominance} balances the optimal reward gain and the risk of future sub-optimal decisions when making a current-stage decision. For example, though the optimal regime could be hardly assigned to some patients, i.e., $P^{\dagger}(d^*_t, H_t)$ is large, the optimal decision $d^*_t$ is still preferred if its expected optimal reward gain, $R^*(d^*_t, H_t)$, is much higher and therefore dominates other regimes. On the other hand, if different decisions yield similar optimal reward gain, i.e., $R^*(d^*_t, H_t) - R^*(d^{\dagger}_t, H_t)$ is small, and there is a high chance of significant losses from future sub-optimal treatments after assigning the optimal treatment at the current stage, i.e., $(R^*(d^*_t, H_t) - R^{\dagger}(d^*_t, H_t)) \cdot P^\dagger(d^*_t,H_t)$ is large, the SWL estimators will choose an alternative decision which provides a reasonable future reward while minimizing the reward losses from future sub-optimal treatments. Accordingly, due to Condition \ref{def:optimal-dominance}, our proposed SWL is able to recover the optimal regime and adaptively make treatment changes based on the expected rewards, regime assignment rates, and possibilities of making sub-optimal regime from the collected empirical data.

\subsection{Finite-sample performance error bound}
\label{sec:thm-3}

Theorems \ref{thm:fisher-1} and \ref{thm: optimal-fisher} establish the consistency properties of the proposed SWL method. To investigate finite sample performance of the proposed approach,  we also establish the performance error bound and investigate the convergence rate. In the following, we require measuring the function space complexity for the parameterized functional space $\mathcal{F}$. 

\begin{assumption}
(Capacity of Function Space) Let $\mathcal{F} = \{f \in \mathcal{F}: \|f\| \le 1\}$ and $H_1,...,H_n \in \mathcal{H}$. There exist constants $C > 0$ and $0 < \alpha < 1$ such that for any $u > 0$, the following condition on metric entropy is satisfied: 
\begin{equation}
    \log \mathcal{N}_2(u, \mathcal{F}, H_{1:n}) \le C \left( \frac{1}{u}\right)^{2 \alpha}.
\end{equation}
\label{assumption:Metric-Entropy}
\end{assumption}
Assumption \ref{assumption:Metric-Entropy} characterizes the functional space complexity with the logarithmic minimum number of balls with radius $u$ required to cover a unit ball in $\mathcal{F}$, and is satisfied under various functional spaces such as the reproducing kernel Hilbert space (RKHS) and Sobolev space \citep{van2000empirical,steinwart2008support}. Consequently, the performance bound between $V^{SW}(\mathbf{f}^*)$ and $\hat{V}^{SW}_\psi(\hat{\mathbf{f}}_n)$ is provided in the following Theorem \ref{theorem:performance-bound}.

\begin{theorem}
Under Assumptions \ref{A-reward}-\ref{assumption:Metric-Entropy}, there exist constants $C_1 > 0$ and $0 < \alpha < 1$ such that for any $\delta \in (0, 1)$, w.p. at least $1-\delta$, the performance error is upper-bounded by:
\begin{equation}
\left|V^{SW}(\mathbf{f}^*) - \hat{V}^{SW}_\psi(\hat{\mathbf{f}}_n) \right| \le  \underbrace{\frac{M}{c_0^T} \sum_{j=1}^T \omega_j \epsilon_{n,j}}_{\text{Surrogate error}}  + \underbrace{\frac{6(\alpha+1)}{\alpha} \left[\alpha C_1 \sqrt{\frac{T}{n}} \left( \frac{\lambda M}{4c_0^T}\right)^\alpha \right]^{\frac{1}{\alpha + 1}} + \frac{9M}{c_0^T} \sqrt{\frac{\log 2 / \delta}{2n}}}_{\text{Empirical estimation error}},
\end{equation}
where $\epsilon_{n,j} = \sup_{A_j, H_j}\left| \mathbb{I}(A_jf_j(H_j) > 0) - \psi(A_jf_j(H_j); \lambda_n) \right|$.
\label{theorem:performance-bound}
\end{theorem}

In Theorem \ref{theorem:performance-bound}, the finite-sample performance bound can be broken down into two separate bounds: the surrogate error bound between $V^{SW}$ and $V^{SW}_\psi$ and the empirical estimation error bound of $V^{SW}_\psi$. As a result, the SWL performance error convergence rate could be obtained at $\mathcal{O}(\epsilon_{n, j} + n^{-1/(2\alpha+2)})$. In particular, the first term depends on the choice of the surrogate function. When the surrogate is well-selected as a logistic function, e.g., $\psi(x;\lambda_n) = (e^{-\lambda_n x} + 1)^{-1}$ with a rate hyper-parameter $\lambda_n$, the surrogate error $\epsilon_{n, j}$ vanishes to zero at the rate of $\mathcal{O}(e^{-n})$, which is much faster than the second term; and therefore the performance error bound could be reduced to $\mathcal{O}(n^{-1/(2\alpha+2)})$.

Furthermore, based on the $\mathcal{O}(n^{-1/(2\alpha+2)})$ convergence rate, Theorem \ref{theorem:performance-bound} provides the finite-sample upper bound which validates the estimation risk and demonstrates how SWL converges under different parametric space settings. For instance, when the historical information $\mathcal{H}$ is an open Euclidean ball in $\mathbf{R}^d$ and the functional space is specified as the Sobolev space $\mathbb{W}^{k}(\mathcal{H})$ where $k > d/2$, one can choose $\alpha=d/2k$ to obtain an error upper bound of rate at $\mathcal{O}(n^{-d/2(d+2k)})$. In addition, when the functional space is finite, the upper error bound could reach the best rate at $\mathcal{O}(n^{-1/2})$ as $\alpha \rightarrow 0$, which achieves the optimal rate provided in the literature  \citep{SOWL, zhao2019efficient}.

To summarize, our finite-sample performance error bound recovers the best-performing convergence rate found in the existing literature and meanwhile provides a non-asymptotic explanation of the proposed multi-stage DTR method in empirical settings. With different choices of metric entropy, the performance error bound can be flexibly adapted to various functional spaces and is not limited to the RKHS discussed in \cite{SOWL}. In addition, our metric condition from Assumption \ref{assumption:Metric-Entropy} only needs to be satisfied under a more relaxed empirical $L_2$-norm on the collected samples $H_{1:n}$, compared to the supremum norm assumption in \cite{zhao2019efficient}.

\section{Implementation and Algorithm} \label{sec:algo}

In this section, we provide our main algorithm for stage importance scores searching and the optimal SWL regime estimation. The goal is to find a set of optimal parameters that minimizes the MSE of rewards \eqref{eq:stage-importance-score-est} and maximizes the objective function of SWL \eqref{eq:SWL_obj_max_sur}.

The algorithm starts with finding the stage importance weights by constructing the neural network as specified in Figure \ref{fig:lstm_frameworkl}, which could be summarised into two major steps. First, we model the deterministic summary function $\mathbf{S}$ via LSTMs and estimate patients' historical information $H_j$ at each stage. Second, we use the estimators of stage importance scores to scale $H_j$ and apply fully-connected (FC) layers on the weighted historical information to estimate the total rewards according to the attention mechanism. Once the surrogate total rewards are estimated, the MSE loss between the observed and surrogate total rewards can be computed, and the parameters in the neural networks are updated from the back-propagation process with stochastic gradient descent (SGD)-based optimizers \citep{robbins1951stochastic}.

For estimating the optimal regime,  there are still two missing pieces need to be filled in according to the SWL objective function \eqref{eq:SWL_obj_max_sur}: the function representations of the target regime and the propensity scores of each observed treatment. In this proposed algorithm, we model the treatment rules $\{f_j\}_{j=1}^T$ with a FC-network. Since the treatment rule could be linear or non-linear, we adjust the activation functions applied on each layer within the network accordingly. To estimate the propensity sores, we apply the logistic regression for each individual at each stage $j$, i.e., $\{\hat{\pi}_{ij}\}_{i=1}^n$. Finally, we combine every component and managed to present the entire workflow in Algorithm \ref{alg:SWL}.

\begin{figure}[hbt!]
\begin{algorithm}[H]
\caption{Stage Weighted Learning} \label{alg:SWL}
\small 
\begin{algorithmic}[1]
    \State \textbf{Initialize} stage weights $\{\omega_j\}_{j=1}^T$; the LSTMs parameterized by $\theta_L$; the stage-weight FC-network parameterized by $\theta_s$; the treatment FC-network parameterized by $\theta_f$; learning rate $\lambda$; maximum iterations $T_{max}$; and a stopping error criterion $\epsilon_s$
    \State \textbf{Input} all observed sequence $\{(X_{i1}, A_{i1}, X_{i2}, A_{i2},..., X_{iT}, A_{iT}, R_i)\}_{i=1}^n$
    \For{$k \gets 1$ to $T_{max}$}     
        \State Compute gradient w.r.t. $\theta_l$, $\theta_s$ and $\{\omega_j\}_{j=1}^T$ as
        \State $\mathcal{L}_1^{k} = \frac{1}{n}\sum_{i=1}^n \left[R_i - \sum_{j=1}^T \text{FC}_{\theta_s}^k \left((\omega_j \cdot \text{LSTM}_{\theta_l}^k(X_{ij}, A_{i,j-1})\right)\right]^2$
        \State Update parameters of interests $(\{\omega_j\}_{j=1}^T, \theta_l, \theta_s)^{k+1} \gets (\{\omega_j\}_{j=1}^T, \theta_l, \theta_s)^{k} - \lambda \cdot \nabla \mathcal{L}_1^k$
        \State Stop if $|\mathcal{L}_1^{k+1} - \mathcal{L}_1^{k}| \le \epsilon$
    \EndFor
    \State \textbf{Normalize} $\hat{\omega}_j = \exp({|\hat{\omega}_j|}) / \sum_{j=1}^T \exp({|\hat{\omega}_j|})$ and finalize $\hat{H}_{ij} = \text{LSTM}_{\theta_l^k} (X_{ij}, A_{i,j-1})$
    \State \textbf{Estimate} $\{(\hat{\pi}_{ij})_{j=1}^T\}_{i=1}^n$ via logistic regressions on $A_{ij} \sim 1 + \hat{H}_{ij}$
    \For{$k \gets 1$ to $T_{max}$}
        \State Compute gradient w.r.t. $\theta_f$ as
        \State $\mathcal{L}_2^k =  -\frac{1}{n} \sum_{i=1}^n \left[\frac{R_i}{\prod_{j=1}^T \hat{\pi}_{ij}} \cdot \sum_{j=1}^T  \hat{\omega}_j \cdot \left(\exp({- A_{ij} \cdot \text{FC}_{\theta_f^k}(\hat{H}_{ij}))}) + 1 \right)^{-1} \right]$ 
        \State Update parameters of interests $\theta_f^{k+1} \gets \theta_f^{k} - \lambda \cdot \nabla \mathcal{L}_2^k$
        \State Stop if $|\mathcal{L}_2^{k+1} - \mathcal{L}_2^{k}| \le \epsilon$
    \EndFor
    \State \textbf{Return} estimated treatment regime network $\hat{\theta}_f = \theta_f^k$
\end{algorithmic}
\end{algorithm}
\end{figure}

Algorithm \ref{alg:SWL} demonstrates the end-to-end procedure of the SWL optimal regime estimation. For illustration purposes, we adopt neural networks and optimize via the standard SGD method. But one can also choose to first parameterize the functions and estimate the function parameters through the more conventional Broyden–Fletcher–Goldfarb–Shanno (BFGS) method \citep{head1985broyden}, or the Nelder–Mead method \citep{olsson1975nelder} on the computed loss. In the following, we further introduce techniques that could be considered to improve neural network convergence and estimation results. For instance, the Adam optimizer \citep{Adam} could be a suitable alternative to improve the convergence performance of SGD on highly-complex and non-convex objective functions. In addition, instead of setting learning rates to be constant as presented in the algorithm, utilizing the cosine annealing warm-restart schedule \citep{cosAnnealing} and different initialization seeds \citep{initialseed} could improve the optimization to achieve better local convergence. Furthermore, we can also tune the hyper-parameters, such as the learning rate and the number of network hidden layers, by conducting a $d$-fold cross-validation on the dataset. The detailed cross-validation procedure is described as follows. The dataset is first randomly partitioned into $d$ evenly-sized subsets, and then the neural network is trained on each of the $(d-1)$ subsets and tested on one remaining subset. After averaging the $d$ testing loss, the set of hyper-parameters with minimal testing loss is selected as optimal based on the empirical dataset. Once the two-step algorithm is converged and the maximal value of the objective function is reached, we obtain the optimal empirical SWL regime.

\section{Simulation} \label{sec:simulation}

In this section, we present simulation studies to showcase the empirical advantages of our proposed methods over popular multi-stage DTR frameworks; e.g., Q-learning \citep{zhao2009reinforcement}, BOWL \citep{SOWL}, RWL\citep{zhou2017residual}, augmented-IPW estimator (AIPWE) \citep{zhang2013robust}, and C-learning \citep{zhang2018c}, where the later two methods utilize robust estimators. Specifically, we investigate the effects of the sample size, number of stages, optimal regime assignment rate and regime function complexity on the model performance. Furthermore, we show the advantages of incorporating stage importance scores when stage heterogeneity exists in the decision stages.

The general simulation setting is described as follows. First of all, a total number of 20 features $\{X_{i1k}\}_{k=1}^{20}$ are independently generated from a standard normal distribution $N(0, 1)$ at baseline ($t=1$), and progress according to the treatment assigned at the previous decision stage via two progression functions $f_t: \mathcal{X}_t \mapsto \mathcal{X}_{t+1}$ and $g_t: \mathcal{X}_t \mapsto \mathcal{X}_{t+1}$, i.e., 
\begin{equation}
     X_{i, t+1} = \mathbb{I}(A_{it}=1) \cdot f_t(X_{it}) + \mathbb{I}(A_{it}=-1) \cdot g_t(X_{it}).
\end{equation}
Here, we choose $f_t(X_{it})=0.8 \cdot X_{it} + 0.6 \cdot \epsilon$ and $g_t(X_{it})=0.6 \cdot X_{it} + 0.8 \cdot \epsilon$, where $\epsilon \sim \mathcal{N}(0, I)$. Notably, since past treatment assignments affect the health variables, which in turn influence future treatment assignments, we establish temporal dependence and interactions among treatments at different stages. 

Next, we design the optimal treatment regime function at each decision stage under linear and non-linear settings. Under the linear setting, the optimal regime function is a linear combination of the covariates without interactions; whereas under the nonlinear setting, we select functions $g_t$ from a basis of functions $\{X, X^2, X^3, \arctan X, \text{sign}(X)\}$, and interaction terms are included among the transformed covariates to increase the function complexity. The optimal treatment regime generation procedure is formalized as follows,
\begin{equation}
    f_t^*(X_{i,t,1:K}) = \begin{cases}
      \sum_{j \in \mathcal{J}} \beta_{tj} \cdot X_{i,t,j} & \text{Linear setting} \\
      \sum_{j \in \mathcal{J}} \beta_{tj} \cdot \prod_{s \in \mathcal{S}_j} g_{ts}\left(X_{i,t,s}\right) & \text{Non-linear setting},
    \end{cases}
\end{equation}
where $\beta_{tj} \sim N(0,1)$, $\mathcal{J}$ is the randomly selected covariate index with cardinality $|\mathcal{J}| \sim \text{Unif}(5, 20)$, $\mathcal{S}_j$ is a random index set with cardinality $|\mathcal{S}_j| \sim \text{Unif}(1,3)$ specifying the interaction terms for the $j^{th}$ transformed covariate, and  each $g_{ts}$ is a nonlinear function randomly sampled from the pre-specified functional basis. Consequently, the optimal decision functions and the number of covariates contributing to the optimal treatment rule vary at each decision time, and can be expanded to long decision sequences. 

Lastly, we define a linear immediate reward function after obtaining the optimal decision $d_t^*$ from the optimal regime at each step as
\begin{equation}
    r_{it} = \mathbf{\Tilde{r}}_t(\omega_t, X_{i,t,1:K}, A_{it}) = \omega_t \left[\left(\sum_{j \in \mathcal{J}_r} \beta_{tj}^r \cdot X_{i,t,j} \right) + A_{it} \cdot d_{it}^* \right] + \epsilon_r,
\end{equation}
where $|\mathcal{J}_r| \sim \text{Unif}(5, 20)$, $\beta_{tj}^r \sim N(0,1)$ and $\epsilon_r \sim N(0,1)$. Notice that the reward function consists of three main components: the base reward from patients' covariates, the treatment effect, and the stage importance scores. To specify the values of the importance scores $\{\omega_t\}_{t=1}^T$, we  sample weights from $\text{Unif}(0,0.2)$ for non-important stages and from $\text{Unif}(0.8,1.0)$ for important stages. The importance scores are then normalized such that $\omega_t =\alpha_t / \sum_{t=1}^T \alpha_t$. Correspondingly, important stages have larger importance scores and thus more substantial treatment effects. The total reward for each patient $i$ is computed as $R_i = \sum_{t=1}^{T} r_{it}$, and the performance of the assigned regime is evaluated by the value function $V(\mathbf{X}, \mathbf{A}) = \frac{1}{n} \sum_{i=1}^n R_i$.

In general, for each specification of listed parameters under the general setting, we repeat experiments $50$ times for data generation. All methods are trained using $80\%$ of the simulated training data, and evaluated on the 20\% testing set via value functions and the matching accuracy between the estimated and optimal treatment regimes. Additionally, since the true data-generating process is known in the simulation, we progress the patient's health variables according to the treatments assigned by the regime and calculate their total rewards accordingly. Comprehensive experiment results can be found in Appendix \ref{S:simulation}.

\subsection{Effects of sample size and number of decision stages} \label{sim:6-1}

Sample sizes and the number of decision stages are the two important factors that affect the \textit{curse of full matching} as introduced in section \ref{sec:curse-of-full-matching}. To fully examine the effects of these two factors on our proposed algorithm, we first conduct simulations on sample sizes $n=500, 1000, 5000$, and the number of decision stages $T=5,8,10$ where treatments $A_t$ are randomly matched with the nonlinear optimal decisions at $50\%$ chance and the number of important stages is set to $0$. The results of model performance are summarized in Table \ref{table: number-of-samples}.

\begin{table}[hbt!]
    \centering
    \bgroup
    \def\arraystretch{1.5}%
    \resizebox{0.98\textwidth}{!}{
\begin{tabular}{|c|c|ccccc|cc|cc|c|}
\hline
T & n &       Q-learning & BOWL & AIPWE & RWL & C-learning & \textbf{SAL} & \textbf{SWL} & Observed & Oracle & \begin{tabular}{@{}c@{}} \textbf{Imp-rate} \\ (to Best)\end{tabular} \\
\hline
5 & 5000 & 0.157 (0.063) & 0.580 (0.030) & 0.380 (0.040) & 0.611 (0.024) & 0.604 (0.024) & \textbf{0.673}  (0.025) & 0.671 (0.026) & -0.000 (0.017) & 0.999 (0.007) & 10.081\% \\
 & 1000 & 0.145 (0.064) & 0.418 (0.050) & 0.215 (0.046) & 0.463 (0.044) & 0.421 (0.049) & \textbf{0.533}  (0.045) & \textbf{0.533}  (0.045) & -0.008 (0.035) & 0.995 (0.015) & 14.979\% \\
 & 500 & 0.162 (0.061) & 0.316 (0.068) & 0.148 (0.067) & 0.363 (0.054) & 0.336 (0.075) & 0.464 (0.063) & \textbf{0.465}  (0.063) & 0.008 (0.054) & 0.995 (0.021) & 27.924\% \\
\hline
8 & 5000 & 0.131 (0.069) & 0.420 (0.037) & 0.266 (0.028) & 0.466 (0.026) & 0.550 (0.025) & \textbf{0.647}  (0.019) & 0.647 (0.019) & -0.000 (0.012) & 1.000 (0.008) & 17.551\% \\
 & 1000 & 0.112 (0.070) & 0.264 (0.050) & 0.156 (0.038) & 0.314 (0.047) & 0.341 (0.053) & \textbf{0.459}  (0.037) & \textbf{0.459}  (0.037) & -0.009 (0.032) & 1.001 (0.013) & 34.660\% \\
 & 500 & 0.130 (0.057) & 0.209 (0.053) & 0.117 (0.049) & 0.247 (0.057) & 0.262 (0.055) & \textbf{0.410}  (0.062) & 0.409 (0.062) & 0.011 (0.046) & 1.001 (0.020) & 56.281\% \\
\hline
10 & 5000 & 0.115 (0.072) & 0.347 (0.026) & 0.213 (0.027) & 0.382 (0.034) & 0.518 (0.026) & \textbf{0.624}  (0.021) & 0.624 (0.021) & -0.001 (0.012) & 0.999 (0.009) & 20.541\% \\
 & 1000 & 0.072 (0.051) & 0.214 (0.038) & 0.124 (0.037) & 0.247 (0.041) & 0.308 (0.046) & 0.438 (0.041) & \textbf{0.440}  (0.042) & -0.012 (0.028) & 1.000 (0.016) & 42.969\% \\
 & 500 & 0.100 (0.068) & 0.157 (0.062) & 0.089 (0.048) & 0.199 (0.052) & 0.224 (0.067) & \textbf{0.378}  (0.063) & 0.376 (0.064) & 0.013 (0.042) & 1.000 (0.027) & 68.250\% \\
\hline
\end{tabular}
}
\egroup
\caption{Estimated total rewards when the optimal regime is nonlinear, assigned treatment $A_t \sim \text{Bernouli(0.5)} \cdot d^*_t$ and no important stage. Standard errors are listed next to the estimated means. The Oracle stands for the best estimated total rewards if all treatments are assigned optimaly. The improvement rate compares SAL/SWL against the best performer of competing methods.}
\label{table: number-of-samples}
\end{table}

According to Table \ref{table: number-of-samples}, we notice that SAL/SWL outperform all other competing methods with respect to the estimated total rewards. Given a fixed number of stages, every single model deteriorates as expected when the sample size decreases as indicated by the smaller values of estimated total rewards, but the improvement margin of the proposed methods compared to the best-performing competing method increases. In particular, when $T=5$, the SAL/SWL improves the estimated total rewards nearly three times as much, from $10.08\%$ to $27.92\%$, as the sample size decreases from $5000$ to $500$. In addition, the difference between our model performance and other competing methods enlarges with an increasing number of decision stages. For instance, when $n=5000$, the improvement rates increase from $10.08\%$ to $20.54\%$ as $T$ grows from $5$ to $10$. This implies that the proposed method has a more efficient utilization of the observed information and more advantages when the sample size is small and the number of treatment stages is large. 

\subsection{Full-matching rates between assigned and optimal treatments} \label{sim::6-2}

We illustrate the \textit{curse of full matching} under various sample sizes and numbers of stages. However, the empirical dilemma could compromise the convergence of the DTR methods at the same time. To describe a more straightforward association between full-matching rates and model performance while minimizing the effects from non-convergent results, we set $n=5000$, $T=10$, and directly adjust the matching probabilities between the assigned and optimal treatments. Specifically, at each stage, we consider the assigned treatment with a 50\%, 60\%, 70\%, and 80\% probability of matching the optimal treatment. As a result, the number of stage receiving optimal decisions follows a binomial distribution with these respective probabilities, and the full-matching rates are $0.5^{10} \approx 0.001$, $0.6^{10} \approx 0.006$, $0.7^{10} \approx 0.03$, and $0.8^{10} \approx 0.1$.

\begin{table}[hbt!]
\centering
\bgroup
\def\arraystretch{1.5}%
\resizebox{\textwidth}{!}{
    \begin{tabular}{|c|ccccc|cc|cc|c|}
    \hline
    \begin{tabular}{@{}c@{}} \textbf{Full-Matching Probability} \\ $P(\sum_{t=1}^{10} \mathbb{I}(A_t = d^*_t) = 10)$ \end{tabular}
      & BOWL &  Q-learning & AIPWE & RWL & C-learning & \textbf{SAL} & \textbf{SWL} &   Observed &      Oracle & \begin{tabular}{@{}c@{}} \textbf{Imp-rate} \\ (to BOWL)\end{tabular} \\
    \hline
    Scenario 1 (0.100) & 0.659 (0.014) & 0.174 (0.068) & 0.214 (0.030) & 0.639 (0.019) & 0.511 (0.026) & \textbf{0.686}  (0.016) & \textbf{0.686}  (0.016) & 0.598 (0.011) & 1.000 (0.009) & 4.058\% \\
    Scenario 2 (0.030) & 0.578 (0.022) & 0.165 (0.070) & 0.222 (0.027) & 0.547 (0.024) & 0.524 (0.027) & 0.657 (0.020) & \textbf{0.658}  (0.021) & 0.400 (0.011) & 0.999 (0.009) & 13.780\% \\
    Scenario 3 (0.006) & 0.455 (0.034) & 0.146 (0.075) & 0.218 (0.028) & 0.428 (0.028) & 0.530 (0.023) & \textbf{0.635}  (0.022) & 0.633 (0.023) & 0.200 (0.010) & 0.999 (0.009) & 39.603\% \\
    Scenario 4 (0.001) & 0.347 (0.026) & 0.115 (0.072) & 0.213 (0.027) & 0.382 (0.034) & 0.518 (0.026) & \textbf{0.624}  (0.021) & 0.624 (0.021) & -0.001 (0.012) & 0.999 (0.009) & 79.694\% \\

\hline
    \end{tabular}
}
\egroup
\caption{Estimated total rewards of listed models when $n=5000$, $T=10$, and the treatments are matched with the nonlinear optimal decisions based on the pre-specified matching probabilities. Standard errors are provided in parentheses. Improvement rates compare SAL/SWL against BOWL.}
\label{table:matching-rates}
\end{table}

Based on the results presented in Table \ref{table:matching-rates}, we observe that SAL/SWL outperform the rest of the competing methods, with greater improvement over BOWL as the full-matching probability decreases. This is expected since BOWL, as an IPWE-based method, depends heavily on full-matching treatments for convergence, while our methods incorporate sub-optimal treatments and thus are more robust across varying treatment matching scenarios. Among competing methods, the Q-learning performs poorly due to its dependence on a correctly specified outcome model, which is challenging with high heterogeneity and a large number of stages. AIPWE and RWL face similar challenges with their augmented unbiased and residual terms, and AIPWE further suffers from its complex estimator format. The C-learning, which combines regression-based methods with AIPWE policy search techniques, shows the highest performance among the competing methods when the full-matching rates are small. Nonetheless, both the regression and robust policy-search components increase computational complexity and rely on outcome models, parametric or non-parametric, to stabilize the IPWEs. In contrast, the proposed methods provide a fundamental solution to break the \textit{curse of full-matching}, effectively enhance sample efficiency, and achieve superior model performance.

\subsection{Optimal regime function complexities}
\begin{figure}[H]
    \centering
    \includegraphics[width=0.95\textwidth]{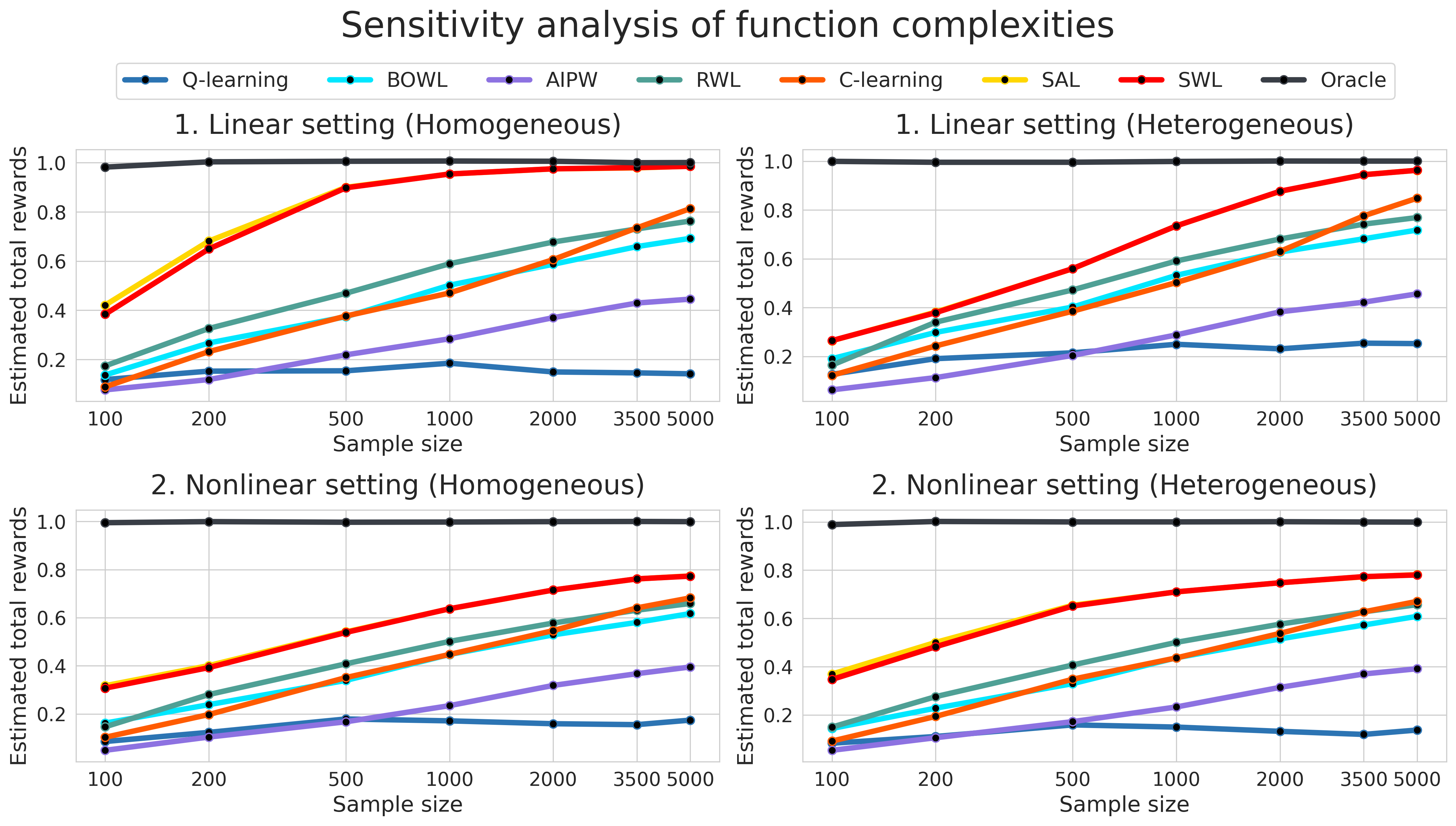}
    \caption{Sensitivity plots of estimated total rewards under four function settings against sample sizes. The number of decision stages is set to 5 and there are no important stages for this example.}
    \label{fig:simulation-sensitivity}
\end{figure}

In this numerical study, we are interested in analyzing the sensitivity of our proposed method to the functional complexity of the underlying optimal regime. Specifically, we also consider the linear treatment regimes and include the  homogeneous decision rule setting where the optimal rules are the same at all time points, i.e. $f_j^* = f_1^*$ for all $2 \le j \le T$. Note that the homogeneous rules can be oftentimes encountered in a high-frequency treatment session as the optimal regime is unlikely to update in a short period of time. We combine the results of four settings when $T=5$ in Figure \ref{fig:simulation-sensitivity}. 

As algorithms converge, a decrease in the performance improvement rate/slope is expected. According to this criterion, we observe that the proposed SAL/SWL methods have similar convergence rates and converge to Oracle faster than all other methods. For instance, under the linear homogeneous treatment rule setting, SWL achieves an averaged $97.36\%$ matching accuracy, compared to $74.71\%$ for BOWL when the sample size reaches $1000$. Though the presented empirical results can be affected by the implementation and choices of hyper-parameters, based on the similar performance increasing rate between SWL and BOWL when the sample size is larger than $2000$, we can confirm with our theoretical results that our SWL reaches the same state-of-the-art asymptotic convergence rate as BOWL. In addition, from the deteriorated performance in the nonlinear heterogeneous setting compared to the linear homogeneous setting, we verify that the increasing level of functional complexity raises the optimization difficulties and hinders the model convergence rate with limited empirical examples. Nevertheless, our proposed method outperforms all competing methods by a considerable margin, especially when the sample size is small.

\subsection{Stage heterogeneity: number of important stages} \label{sim:6-4}

\begin{figure}[hbt!]
    \centering
    \includegraphics[width=0.9\textwidth]{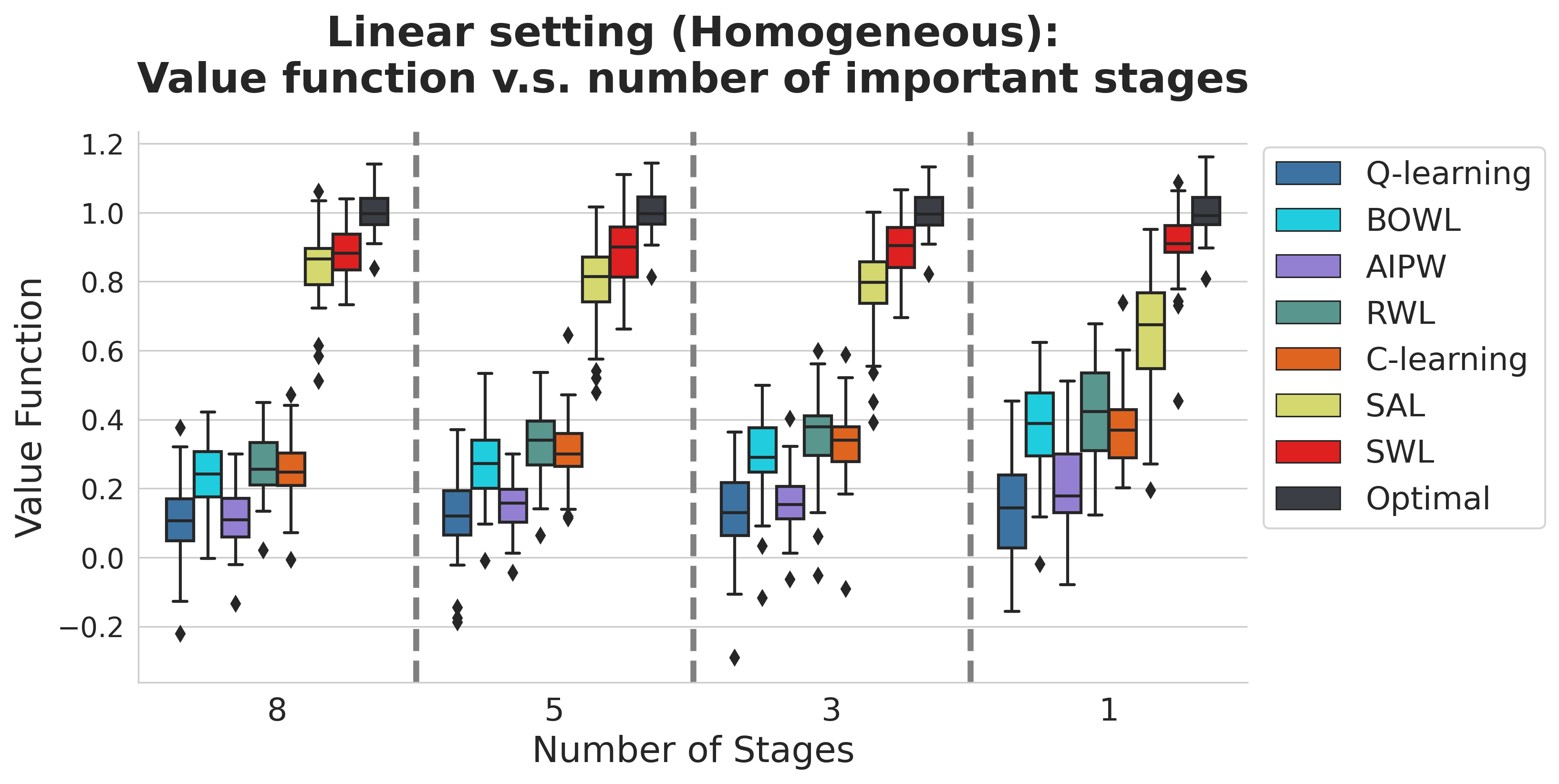}
    \caption{Boxplots of the estimated total rewards of listed methods versus the number of important stages when $n=500$, $T=10$, and the optimal treatment rule is linear and homogeneous.}
    \label{fig:simulation-stage-heteor}
\end{figure}

In this subsection, we illustrate the advantages of incorporating stage heterogeneity with the stage importance scores. We adjust the level of heterogeneity by changing the number of important stages, where fewer important stages induce stronger heterogeneity. In addition, to maximize the heterogeneity among stages, we consider the linear homogeneous decision rule setting as illustrated in the previous simulation subsection. Figure \ref{fig:simulation-stage-heteor} provides the obtained results when $n=500$ and $T=10$.

As shown, while the proposed SAL still outperforms the rest of the competing methods under the first three scenarios, the importance scores can further improve the performance of SAL with a greater margin when the heterogeneity among the stages gets stronger. Moreover, the stability of the SAL can be improved with the stage weights when strong stage heterogeneity exists. We conclude that the proposed stage importance scores are able to explicitly incorporate stage heterogeneity into the SAL estimator and can be used for regime searching on stages which contribute to improving treatment effects.

\section{Data Analysis} \label{sec:data}

In this section, we apply the proposed method to UC COVID Research Data Sets (UC CORDS) \citep{uccords}, which combines timely COVID-related testing and hospitalization healthcare data from six University of California schools and systems. As of December 2022, UC CORDS include a total number of 108,914 COVID patients, where 31,520 of them had been hospitalized and 2,333 of them had been admitted to the ICU. Aiming to facilitate hospital management by reducing inpatients' length of stay at hospitals and further prevent them from developing more severe symptoms, we are interested in selecting effective treatments tailored to individual patients.

One of the first few FDA-approved drugs that have been found effective against COVID was Dexamethasone \citep{ahmed2020dexamethasone}. However, the precise treatment plan using Dexamethasone still remains unclear. As suggested by \cite{waterer2020steroids}, clinicians need to consider individual risks especially among elderly patients with age over 65 years old and patients with comorbidities, such as diabetes and cardiovascular diseases. In fact, according to the UC CORDS medical records, elderly patients spend 3 more days on average in both hospitals and ICUs compared to younger patients. Thus, our goal is to apply DTR methods to provide an optimal individualized treatment decision for Dexamethasone (i.e., whether the patient should take the drug or not) with the incorporation of heterogeneity among patients at a decision stage.

In this application, we list two emerging technical challenges. First, the number of decision stages involved can be large during the average 7 days of inpatient stay, and unlike the infinite horizon DTR method \citep{ertefaie2018constructing, zhou2022estimating}, a finite number of treatment stages is considered in this application. As a result, an efficient DTR estimation method should be robust against the \textit{curse of full-matching}, e.g., long decision sequences and few patients receiving optimal treatments. Second, according to \cite{lee2021effect} who found an early administration of Dexamethasone can reduce hospital stay, we speculate that stage heterogeneity exists in evaluating the treatment effect, and therefore examine methods which can incorporate timing effects on the estimation procedure. Taking the above two challenges into account, the proposed SAL and SWL are able to fulfill these needs for this real-data application. 

\begin{figure}[hbt!]
    \centering
    \includegraphics[width=0.85\textwidth]{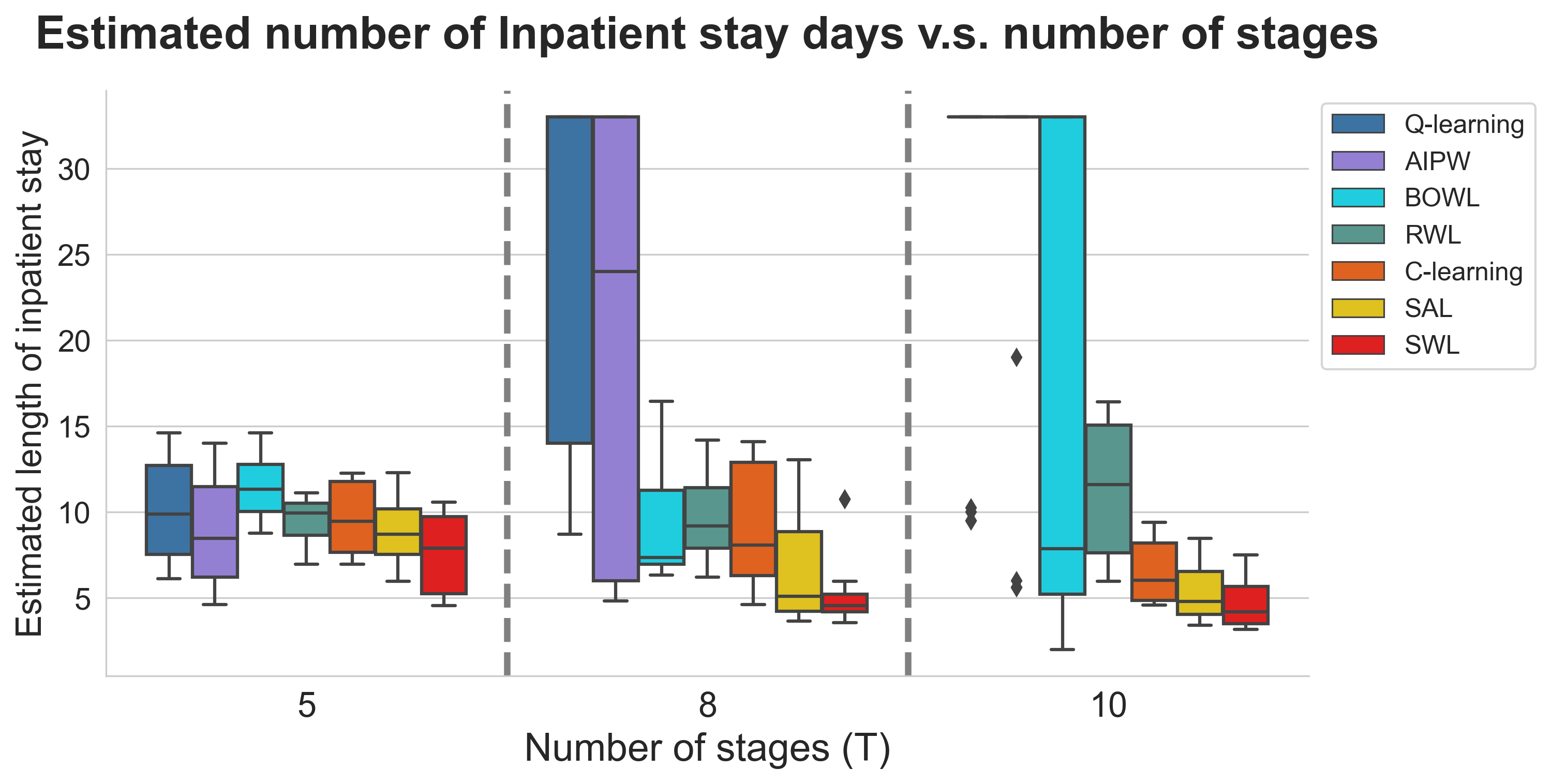}
    \caption{Boxplots of the estimated number of inpatient days by the number of decision stages.}
    \label{fig:real-data}
\end{figure}

We first pre-process UC CORDS data following the procedure elaborated in Appendix \ref{S:real-data}. Then we fit the proposed models and competing methods to search for the optimal DTR of Dexamethasone which can reduce the number of inpatient or ICU day stays for admitted patients. In this application, we include patients who received a total number of 5, 8, and 10 treatment decisions during their stay in the hospital and were later successfully transitioned to outpatient care after recovery. In addition, we randomly select $80\%$ of the data as a training set and repeat the process 20 times to obtain a Monte-Carlo sample of the model performance scores. All methods are evaluated under the empirical value outcome according to \cite{SOWL}, i.e.,
\begin{equation}
    \hat{V}^d = \frac{ \mathbb{E}_n\left\{ R \cdot \prod_{j=1}^T \mathbb{I}(a_j=d_j) / \prod_{j=1}^T \hat{\pi}_j(a_j | H_j) \right\} }{\mathbb{E}_n \left\{ \prod_{j=1}^T \mathbb{I}(a_j=d_j) / \prod_{j=1}^T \hat{\pi}_j(a_j | H_j)\right\}}.
\end{equation}
To show the model performance, we list the estimated outcomes in terms of hospitalization days in Table \ref{table: real-data} and Figure \ref{fig:real-data}, where a smaller value indicates better model performance.

\begin{table}[hbt!]
    \centering
    \bgroup
    \def\arraystretch{1.5}%
    \resizebox{0.95\textwidth}{!}{
        \begin{tabular}{|c|c|c|ccccc|cc|}
        \hline
        Stay Type & Number of Stage  &  Observed &  Q-learning &   BOWL & AIPWE   &      RWL &     C-learning &            \textbf{SAL} &  \textbf{SWL} \\
        \hline
        ICU & 5 (n=623)   & 14.904   & 18.217 (17.920)  &  11.720 (9.412) & 15.251 (11.720) &  10.455 (10.240)  &  9.518 (10.252) & 7.175 (7.832) &  \textbf{5.855} (4.666) \\
        & 8 (n=345)       & 14.345   &  44.200 (16.167) &   10.340 (8.295) &  41.300 (17.254) & 7.173 (6.237)  &  7.680 (7.124) & \textbf{4.913} (3.375) &  6.749 (7.636) \\ \hline
        Inpatient & 5 (n=3256)    & 9.077   &  10.108 (3.291) & 11.525 (2.212) & 10.003 (6.660) & 9.493 (1.507)   & 9.784 (2.667) &  9.019 (3.051) & \textbf{7.790} (2.861) \\
        & 8 (n=1876)      & 8.548           & 24.950 (11.200) &  9.543 (4.824) & 19.815 (13.451) & 9.589 (2.790)  &  9.490 (4.690) & 6.704 (3.541) & \textbf{5.664} (3.263)   \\
        & 10 (n=1419)   & 8.250        &   27.712 (10.713) & 15.129 (12.910) & 27.800 (10.884) & 11.412 (4.031)  &  6.537 (2.065) & 5.582 (2.318) & \textbf{4.817} (1.958) \\
        \hline
        \end{tabular}
    }
    \egroup
\caption{Estimated number of hospitalization days obtained from DTR methods under two different stay types: Inpatient and ICU. The sample size is listed next to the number of stages.}
\label{table: real-data}
\end{table}

Both Table \ref{table: real-data} and Figure \ref{fig:real-data} show that our SAL/SWL methods achieve the overall best performance in terms of reducing the length of hospital stay for patients following the suggested DTR. Specifically, the proposed methods reduce almost $40\%$ of the ICU duration from 10 days to 6 days, compared to the C-learning when a total of 5 decision stages is involved. Apart from the averaged performance, our methods attain leading model stability compared to other competing methods. In particular, the SWL can further improve the model stability of SAL over a larger number of decision stages, which involve increased level of heterogeneity and can be more effectively captured by the stage importance scores via the attention-based neural network.

Based on our analyses, we can summarize that Q-learning has difficulties to estimate the outcome when a large number of decision stages are involved and the underlying reward mechanism is complicated, especially in the case of COVID where the association of recovery time and Dexamethasone treatment still remains unclear. Meanwhile, BOWL is based on the IPWE framework and requires a sufficient number of patients receiving optimal treatments at all stages. Thus, as the number of stages increases and the number of involved patients decreases, the BOWL shows deteriorating performance and larger variance. Among methods aimed at stabilizing IPWEs, AIPWE utilizes robust estimators, but due to high heterogeneity among COVID patients, accurately modeling either propensity or outcome to achieve double robustness is challenging in practice. On the other hand, RWL, which mitigates outcome shifts via residuals, and C-learning, which leverages regression-based models, have both shown effectiveness in stabilizing IPWEs. However, both methods still require estimating outcomes, leading to increased computational complexities and vulnerability to model misspecification as the number of decision stages grows. In comparison, our methods are able to combine the efficiency of IPWE-based methods and meanwhile improve model stability by taking into account heterogeneity-matching schemes between the observed and underlying optimal regimes. Our real-world application to the UC CORDS dataset demonstrates the superior empirical performance of the proposed SAL/SWL methods.

\section{Discussion} \label{sec:discussion}

In this paper, we introduce a novel individualized learning method for estimating the optimal dynamic treatment regime. The proposed SAL/SWL utilizes the matching status between the observed and underlying optimal regime at any stage and substantially improves the sample efficiency of the IPWE-based approaches. With the stage importance scores, SWL enhances stage heterogeneity and therefore more accurately captures the differences in treatment effectiveness at various stages. In theory, we establish the Fisher consistency and a finite-sample performance error bound, which achieves the best convergence rate in the literature and provides a non-asymptotic explanation.

There are future improvements and extensions for our work. For example, to estimate the stage importance scores, we construct an attention-based neural network in the current work. As a result, the estimated importance scores remain fixed at all stages and provide a general interpretation of the importance of a stage among the patient population. However, the treatment stage heterogeneity could vary among the patients. As for future exploration, we can incorporate the multi-head attention architecture \citep{vaswani2017attention} which provides individualized stage importance scores at the patient level. Additionally, our current framework accommodates only binary treatment options at each stage, which could be explored for potential expansion to multiple treatment choices in the future. Furthermore, we can investigate extending our work to an infinite-horizon or shared-parameter setting, where decision rules become Markovian over a long series of decision stages. We can also develop a data-driven procedure to estimate the weighting scale used in the general framework introduced by SAL. These extensions are further discussed in Appendix \ref{S:KIPWL} and \ref{S:simulation}.

\vskip 0.2in
{
\small
\bibliography{reference}

\begin{thebibliography}{65}
\providecommand{\natexlab}[1]{#1}
\providecommand{\url}[1]{\texttt{#1}}
\expandafter\ifx\csname urlstyle\endcsname\relax
  \providecommand{\doi}[1]{doi: #1}\else
  \providecommand{\doi}{doi: \begingroup \urlstyle{rm}\Url}\fi

\bibitem[Ahmed and Hassan(2020)]{ahmed2020dexamethasone}
Mukhtar~H Ahmed and Arez Hassan.
\newblock Dexamethasone for the treatment of coronavirus disease (covid-19): A review.
\newblock \emph{SN Comprehensive Clinical Medicine}, 2\penalty0 (12):\penalty0 2637--2646, 2020.

\bibitem[Bahdanau et~al.(2014)Bahdanau, Cho, and Bengio]{bahdanau2014neural}
Dzmitry Bahdanau, Kyunghyun Cho, and Yoshua Bengio.
\newblock Neural machine translation by jointly learning to align and translate.
\newblock \emph{arXiv preprint arXiv:1409.0473}, 2014.

\bibitem[Balzanelli et~al.(2022)Balzanelli, Distratis, Lazzaro, D’Ettorre, Nico, Inchingolo, Dipalma, Tomassone, Serlenga, Dalagni, et~al.]{balzanelli2022new}
Mario~Giosu{\`e} Balzanelli, Pietro Distratis, Rita Lazzaro, Ernesto D’Ettorre, Andrea Nico, Francesco Inchingolo, Gianna Dipalma, Diego Tomassone, Emilio~Maria Serlenga, Giancarlo Dalagni, et~al.
\newblock New translational trends in personalized medicine: autologous peripheral blood stem cells and plasma for covid-19 patient.
\newblock \emph{Journal of Personalized Medicine}, 12\penalty0 (1):\penalty0 85, 2022.

\bibitem[Bartlett et~al.(2006)Bartlett, Jordan, and McAuliffe]{bartlett2006convexity}
Peter~L Bartlett, Michael~I Jordan, and Jon~D McAuliffe.
\newblock Convexity, classification, and risk bounds.
\newblock \emph{Journal of the American Statistical Association}, 101\penalty0 (473):\penalty0 138--156, 2006.

\bibitem[Blatt et~al.(2004)Blatt, Murphy, and Zhu]{AL-blatt}
Doron Blatt, Susan~A Murphy, and Ji~Zhu.
\newblock A-learning for approximate planning.
\newblock \emph{Techinical Report}, pages 04--63, 2004.

\bibitem[Chakraborty et~al.(2016)Chakraborty, Ghosh, Moodie, and Rush]{chakraborty2016estimating}
Bibhas Chakraborty, Palash Ghosh, Erica~EM Moodie, and A~John Rush.
\newblock Estimating optimal shared-parameter dynamic regimens with application to a multistage depression clinical trial.
\newblock \emph{Biometrics}, 72\penalty0 (3):\penalty0 865--876, 2016.

\bibitem[Dembczy{\'n}ski et~al.(2012)Dembczy{\'n}ski, Waegeman, Cheng, and H{\"u}llermeier]{dembczynski2012label}
Krzysztof Dembczy{\'n}ski, Willem Waegeman, Weiwei Cheng, and Eyke H{\"u}llermeier.
\newblock On label dependence and loss minimization in multi-label classification.
\newblock \emph{Machine Learning}, 88:\penalty0 5--45, 2012.

\bibitem[DeVore et~al.(2021)DeVore, Hanin, and Petrova]{devore2021neural}
Ronald DeVore, Boris Hanin, and Guergana Petrova.
\newblock Neural network approximation.
\newblock \emph{Acta Numerica}, 30:\penalty0 327--444, 2021.

\bibitem[Diamond et~al.(2016)Diamond, Takapoui, and Boyd]{initialseed}
Steven Diamond, Reza Takapoui, and Stephen Boyd.
\newblock A general system for heuristic solution of convex problems over nonconvex sets.
\newblock \emph{arXiv preprint arXiv:1601.07277}, 2016.

\bibitem[Ernst et~al.(2005)Ernst, Geurts, and Wehenkel]{NPQL-1}
Damien Ernst, Pierre Geurts, and Louis Wehenkel.
\newblock Tree-based batch mode reinforcement learning.
\newblock \emph{Journal of Machine Learning Research}, 6:\penalty0 503--556, 2005.

\bibitem[Ertefaie and Strawderman(2018)]{ertefaie2018constructing}
Ashkan Ertefaie and Robert~L Strawderman.
\newblock Constructing dynamic treatment regimes over indefinite time horizons.
\newblock \emph{Biometrika}, 105\penalty0 (4):\penalty0 963--977, 2018.

\bibitem[Geurts et~al.(2006)Geurts, Ernst, and Wehenkel]{NPQL-2}
Pierre Geurts, Damien Ernst, and Louis Wehenkel.
\newblock Extremely randomized trees.
\newblock \emph{Machine Learning}, 63\penalty0 (1):\penalty0 3--42, 2006.

\bibitem[Goetz and Schork(2018)]{goetz2018personalized}
Laura~H Goetz and Nicholas~J Schork.
\newblock Personalized medicine: motivation, challenges, and progress.
\newblock \emph{Fertility and sterility}, 109\penalty0 (6):\penalty0 952--963, 2018.

\bibitem[Head and Zerner(1985)]{head1985broyden}
John~D Head and Michael~C Zerner.
\newblock A \textsc{B}royden—\textsc{F}letcher—\textsc{G}oldfarb—\textsc{S}hanno optimization procedure for molecular geometries.
\newblock \emph{Chemical physics letters}, 122\penalty0 (3):\penalty0 264--270, 1985.

\bibitem[Hochreiter and Schmidhuber(1997)]{hochreiter1997long}
Sepp Hochreiter and J{\"u}rgen Schmidhuber.
\newblock Long short-term memory.
\newblock \emph{Neural Computation}, 9\penalty0 (8), 1997.

\bibitem[Horvitz and Thompson(1952)]{horvitz1952generalization}
Daniel~G Horvitz and Donovan~J Thompson.
\newblock A generalization of sampling without replacement from a finite universe.
\newblock \emph{Journal of the American Statistical Association}, 47\penalty0 (260):\penalty0 663--685, 1952.

\bibitem[Jin et~al.(2020)Jin, Bai, He, Wu, Liu, Han, Liu, and Yang]{jin2020gender}
Jian-Min Jin, Peng Bai, Wei He, Fei Wu, Xiao-Fang Liu, De-Min Han, Shi Liu, and Jin-Kui Yang.
\newblock Gender differences in patients with covid-19: focus on severity and mortality.
\newblock \emph{Frontiers in public health}, 8:\penalty0 545030, 2020.

\bibitem[Kao et~al.(2008)Kao, Tyson, Blakely, and Lally]{kao2008clinical}
Lillian~S Kao, Jon~E Tyson, Martin~L Blakely, and Kevin~P Lally.
\newblock Clinical research methodology i: introduction to randomized trials.
\newblock \emph{Journal of the American College of Surgeons}, 206\penalty0 (2):\penalty0 361--369, 2008.

\bibitem[Kingma and Ba(2014)]{Adam}
Diederik~P Kingma and Jimmy Ba.
\newblock Adam: A method for stochastic optimization.
\newblock \emph{arXiv preprint arXiv:1412.6980}, 2014.

\bibitem[Laber and Zhao(2015)]{laber2015tree}
Eric~B Laber and Ying-Qi Zhao.
\newblock Tree-based methods for individualized treatment regimes.
\newblock \emph{Biometrika}, 102\penalty0 (3):\penalty0 501--514, 2015.

\bibitem[Laha et~al.(2024)Laha, Sonabend-W, Mukherjee, and Cai]{laha2024finding}
Nilanjana Laha, Aaron Sonabend-W, Rajarshi Mukherjee, and Tianxi Cai.
\newblock Finding the optimal dynamic treatment regimes using smooth fisher consistent surrogate loss.
\newblock \emph{The Annals of Statistics}, 52\penalty0 (2):\penalty0 679--707, 2024.

\bibitem[Lee et~al.(2021)Lee, Park, Lee, Park, and Heo]{lee2021effect}
Hyun~Woo Lee, Jimyung Park, Jung-Kyu Lee, Tae~Yeon Park, and Eun~Young Heo.
\newblock The effect of the timing of dexamethasone administration in patients with covid-19 pneumonia.
\newblock \emph{Tuberculosis and \textsc{R}espiratory \textsc{D}iseases}, 84\penalty0 (3):\penalty0 217, 2021.

\bibitem[Loshchilov and Hutter(2016)]{cosAnnealing}
Ilya Loshchilov and Frank Hutter.
\newblock Sgdr: Stochastic gradient descent with warm restarts.
\newblock \emph{arXiv preprint arXiv:1608.03983}, 2016.

\bibitem[Mnih et~al.(2013)Mnih, Kavukcuoglu, Silver, Graves, Antonoglou, Wierstra, and Riedmiller]{mnih2013playing}
Volodymyr Mnih, Koray Kavukcuoglu, David Silver, Alex Graves, Ioannis Antonoglou, Daan Wierstra, and Martin Riedmiller.
\newblock Playing atari with deep reinforcement learning.
\newblock \emph{arXiv preprint arXiv:1312.5602}, 2013.

\bibitem[Mo et~al.(2021)Mo, Qi, and Liu]{mo2021learning}
Weibin Mo, Zhengling Qi, and Yufeng Liu.
\newblock Learning optimal distributionally robust individualized treatment rules.
\newblock \emph{Journal of the American Statistical Association}, 116\penalty0 (534):\penalty0 659--674, 2021.

\bibitem[Murphy(2003)]{AL-murphy}
Susan~A Murphy.
\newblock Optimal dynamic treatment regimes.
\newblock \emph{Journal of the Royal Statistical Society: Series B (Statistical Methodology)}, 65\penalty0 (2):\penalty0 331--355, 2003.

\bibitem[Nahum-Shani et~al.(2012)Nahum-Shani, Qian, Almirall, Pelham, Gnagy, Fabiano, Waxmonsky, Yu, and Murphy]{nahum2012q}
Inbal Nahum-Shani, Min Qian, Daniel Almirall, William~E Pelham, Beth Gnagy, Gregory~A Fabiano, James~G Waxmonsky, Jihnhee Yu, and Susan~A Murphy.
\newblock Q-learning: \textsc{A} data analysis method for constructing adaptive interventions.
\newblock \emph{Psychological Methods}, 17\penalty0 (4):\penalty0 478, 2012.

\bibitem[Olsson and Nelson(1975)]{olsson1975nelder}
Donald~M Olsson and Lloyd~S Nelson.
\newblock The \textsc{N}elder-\textsc{M}ead simplex procedure for function minimization.
\newblock \emph{Technometrics}, 17\penalty0 (1):\penalty0 45--51, 1975.

\bibitem[Pawlik et~al.(2005)Pawlik, Abdalla, Barnett, Ahmad, Cleary, Vauthey, Lee, Evans, and Pisters]{pawlik2005feasibility}
Timothy~M Pawlik, Eddie~K Abdalla, Carlton~C Barnett, Syed~A Ahmad, Karen~R Cleary, Jean-Nicolas Vauthey, Jeffrey~E Lee, Douglas~B Evans, and Peter~WT Pisters.
\newblock Feasibility of a randomized trial of extended lymphadenectomy for pancreatic cancer.
\newblock \emph{Archives of Surgery}, 140\penalty0 (6):\penalty0 584--591, 2005.

\bibitem[Qi et~al.(2020)Qi, Liu, Fu, and Liu]{qi2020multi}
Zhengling Qi, Dacheng Liu, Haoda Fu, and Yufeng Liu.
\newblock \emph{Journal of the American Statistical Association}, 115\penalty0 (530):\penalty0 678--691, 2020.

\bibitem[Qian and Murphy(2011)]{qian2011performance}
Min Qian and Susan~A Murphy.
\newblock Performance guarantees for individualized treatment rules.
\newblock \emph{Annals of Statistics}, 39\penalty0 (2):\penalty0 1180, 2011.

\bibitem[Robbins and Monro(1951)]{robbins1951stochastic}
Herbert Robbins and Sutton Monro.
\newblock A stochastic approximation method.
\newblock \emph{The Annals of Mathematical Statistics}, 22\penalty0 (3):\penalty0 400--407, 1951.

\bibitem[Robins(1986)]{robins1986new}
James Robins.
\newblock A new approach to causal inference in mortality studies with a sustained exposure period—application to control of the healthy worker survivor effect.
\newblock \emph{Mathematical Modelling}, 7\penalty0 (9-12):\penalty0 1393--1512, 1986.

\bibitem[Robins et~al.(1994)Robins, Rotnitzky, and Zhao]{robins1994estimation}
James~M Robins, Andrea Rotnitzky, and Lue~Ping Zhao.
\newblock Estimation of regression coefficients when some regressors are not always observed.
\newblock \emph{Journal of the American Statistical Association}, 89\penalty0 (427):\penalty0 846--866, 1994.

\bibitem[Rubin(1974)]{rubin1974estimating}
Donald~B Rubin.
\newblock Estimating causal effects of treatments in randomized and nonrandomized studies.
\newblock \emph{Journal of Educational Psychology}, 66\penalty0 (5):\penalty0 688, 1974.

\bibitem[Rubin(1980)]{rubin1980randomization}
Donald~B Rubin.
\newblock Randomization analysis of experimental data: The \textsc{F}isher randomization test comment.
\newblock \emph{Journal of the American Statistical Association}, 75\penalty0 (371):\penalty0 591--593, 1980.

\bibitem[Schulte et~al.(2014)Schulte, Tsiatis, Laber, and Davidian]{QALReview}
Phillip~J Schulte, Anastasios~A Tsiatis, Eric~B Laber, and Marie Davidian.
\newblock Q-and \textsc{A}-learning methods for estimating optimal dynamic treatment regimes.
\newblock \emph{Statistical Science: A review journal of the Institute of Mathematical Statistics}, 29\penalty0 (4):\penalty0 640, 2014.

\bibitem[Schulz and Moodie(2021)]{schulz2021doubly}
Juliana Schulz and Erica~EM Moodie.
\newblock Doubly robust estimation of optimal dosing strategies.
\newblock \emph{Journal of the American Statistical Association}, 116\penalty0 (533):\penalty0 256--268, 2021.

\bibitem[Shi et~al.(2018)Shi, Fan, Song, and Lu]{shi2018high}
Chengchun Shi, Alin Fan, Rui Song, and Wenbin Lu.
\newblock High-dimensional \textsc{A}-learning for optimal dynamic treatment regimes.
\newblock \emph{Annals of Statistics}, 46\penalty0 (3):\penalty0 925, 2018.

\bibitem[Steinwart and Christmann(2008)]{steinwart2008support}
Ingo Steinwart and Andreas Christmann.
\newblock \emph{Support \textsc{v}ector \textsc{m}achines}.
\newblock Springer Science, 2008.

\bibitem[Tang and Ishwaran(2017)]{tang2017random}
Fei Tang and Hemant Ishwaran.
\newblock Random forest missing data algorithms.
\newblock \emph{Statistical Analysis and Data Mining: The ASA Data Science Journal}, 10\penalty0 (6):\penalty0 363--377, 2017.

\bibitem[Tao et~al.(2018)Tao, Wang, and Almirall]{tao2018tree}
Yebin Tao, Lu~Wang, and Daniel Almirall.
\newblock Tree-based reinforcement learning for estimating optimal dynamic treatment regimes.
\newblock \emph{The Annals of Applied Statistics}, 12\penalty0 (3):\penalty0 1914, 2018.

\bibitem[Taylor et~al.(2015)Taylor, Cheng, and Foster]{taylor2015reader}
Jeremy~MG Taylor, Wenting Cheng, and Jared~C Foster.
\newblock Reader reaction to “a robust method for estimating optimal treatment regimes” by zhang et al.(2012).
\newblock \emph{Biometrics}, 71\penalty0 (1):\penalty0 267--273, 2015.

\bibitem[Tetzlaff et~al.(2021)Tetzlaff, Schmiedek, and Brod]{tetzlaff2021developing}
Leonard Tetzlaff, Florian Schmiedek, and Garvin Brod.
\newblock Developing personalized education: A dynamic framework.
\newblock \emph{Educational Psychology Review}, 33:\penalty0 863--882, 2021.

\bibitem[Tsiatis et~al.(2019)Tsiatis, Davidian, Holloway, and Laber]{tsiatis2019dynamic}
Anastasios~A Tsiatis, Marie Davidian, Shannon~T Holloway, and Eric~B Laber.
\newblock \emph{Dynamic \textsc{t}reatment \textsc{r}egimes: Statistical \textsc{m}ethods for \textsc{p}recision \textsc{m}edicine}.
\newblock Chapman and Hall/CRC, 2019.

\bibitem[Tsoumakas and Katakis(2007)]{tsoumakas2007multi}
Grigorios Tsoumakas and Ioannis Katakis.
\newblock Multi-label classification: An overview.
\newblock \emph{International Journal of Data Warehousing and Mining (IJDWM)}, 3\penalty0 (3):\penalty0 1--13, 2007.

\bibitem[{University of \textsc{C}alifornia Health}()]{uccords}
{University of \textsc{C}alifornia Health}.
\newblock University of \textsc{C}alifornia \textsc{H}ealth creates centralized data set to accelerate \textsc{COVID}-19 research.

\bibitem[Van~de Geer(2000)]{van2000empirical}
Sara Van~de Geer.
\newblock \emph{Empirical Processes in M-estimation}, volume~6.
\newblock Cambridge university \textsc{p}ress, 2000.

\bibitem[Vaswani et~al.(2017)Vaswani, Shazeer, Parmar, Uszkoreit, Jones, Gomez, Kaiser, and Polosukhin]{vaswani2017attention}
Ashish Vaswani, Noam Shazeer, Niki Parmar, Jakob Uszkoreit, Llion Jones, Aidan~N Gomez, {\L}ukasz Kaiser, and Illia Polosukhin.
\newblock Attention is all you need.
\newblock \emph{Advances in \textsc{N}eural \textsc{I}nformation \textsc{P}rocessing \textsc{S}ystems}, 30, 2017.

\bibitem[Vesanen and Raulas(2006)]{vesanen2006building}
Jari Vesanen and Mika Raulas.
\newblock Building bridges for personalization: A process model for marketing.
\newblock \emph{Journal of Interactive marketing}, 20\penalty0 (1):\penalty0 5--20, 2006.

\bibitem[Waterer and Rello(2020)]{waterer2020steroids}
Grant~W Waterer and Jordi Rello.
\newblock Steroids and \textsc{COVID}-19: \textsc{W}e need a precision approach, not one size fits all.
\newblock \emph{Infectious \textsc{D}iseases and \textsc{T}herapy}, 9\penalty0 (4):\penalty0 701--705, 2020.

\bibitem[Watkins and Dayan(1992)]{watkins1992q}
Christopher~JCH Watkins and Peter Dayan.
\newblock Q-learning.
\newblock \emph{Machine \textsc{L}earning}, 8\penalty0 (3):\penalty0 279--292, 1992.

\bibitem[Xue et~al.(2022)Xue, Zhang, Zhou, Fu, and Qu]{xue2022multicategory}
Fei Xue, Yanqing Zhang, Wenzhuo Zhou, Haoda Fu, and Annie Qu.
\newblock Multicategory angle-based learning for estimating optimal dynamic treatment regimes with censored data.
\newblock \emph{Journal of the American Statistical Association}, 117\penalty0 (539):\penalty0 1438--1451, 2022.

\bibitem[Zhang and Zhang(2018)]{zhang2018c}
Baqun Zhang and Min Zhang.
\newblock C-learning: A new classification framework to estimate optimal dynamic treatment regimes.
\newblock \emph{Biometrics}, 74\penalty0 (3):\penalty0 891--899, 2018.

\bibitem[Zhang et~al.(2012{\natexlab{a}})Zhang, Tsiatis, Davidian, Zhang, and Laber]{zhang2012estimating}
Baqun Zhang, Anastasios~A Tsiatis, Marie Davidian, Min Zhang, and Eric Laber.
\newblock Estimating optimal treatment regimes from a classification perspective.
\newblock \emph{Stat}, 1\penalty0 (1):\penalty0 103--114, 2012{\natexlab{a}}.

\bibitem[Zhang et~al.(2012{\natexlab{b}})Zhang, Tsiatis, Laber, and Davidian]{zhang2012robust}
Baqun Zhang, Anastasios~A Tsiatis, Eric~B Laber, and Marie Davidian.
\newblock A robust method for estimating optimal treatment regimes.
\newblock \emph{Biometrics}, 68\penalty0 (4):\penalty0 1010--1018, 2012{\natexlab{b}}.

\bibitem[Zhang et~al.(2013)Zhang, Tsiatis, Laber, and Davidian]{zhang2013robust}
Baqun Zhang, Anastasios~A Tsiatis, Eric~B Laber, and Marie Davidian.
\newblock Robust estimation of optimal dynamic treatment regimes for sequential treatment decisions.
\newblock \emph{Biometrika}, 100\penalty0 (3):\penalty0 681--694, 2013.

\bibitem[Zhang et~al.(2015)Zhang, Laber, Tsiatis, and Davidian]{zhang2015using}
Yichi Zhang, Eric~B Laber, Anastasios Tsiatis, and Marie Davidian.
\newblock Using decision lists to construct interpretable and parsimonious treatment regimes.
\newblock \emph{Biometrics}, 71\penalty0 (4):\penalty0 895--904, 2015.

\bibitem[Zhang et~al.(2018)Zhang, Laber, Davidian, and Tsiatis]{zhang2018estimation}
Yichi Zhang, Eric~B Laber, Marie Davidian, and Anastasios~A Tsiatis.
\newblock Estimation of optimal treatment regimes using lists.
\newblock \emph{Journal of the American Statistical Association}, pages 1541--1549, 2018.

\bibitem[Zhao et~al.(2015)Zhao, Zeng, Laber, and Kosorok]{SOWL}
Ying-Qi Zhao, Donglin Zeng, Eric~B Laber, and Michael~R Kosorok.
\newblock New statistical learning methods for estimating optimal dynamic treatment regimes.
\newblock \emph{Journal of the American Statistical Association}, 110\penalty0 (510):\penalty0 583--598, 2015.

\bibitem[Zhao et~al.(2019)Zhao, Laber, Ning, Saha, and Sands]{zhao2019efficient}
Ying-Qi Zhao, Eric~B Laber, Yang Ning, Sumona Saha, and Bruce~E Sands.
\newblock Efficient augmentation and relaxation learning for individualized treatment rules using observational data.
\newblock \emph{The Journal of Machine Learning Research}, 20\penalty0 (1):\penalty0 1821--1843, 2019.

\bibitem[Zhao et~al.(2012)Zhao, Zeng, Rush, and Kosorok]{OWL}
Yingqi Zhao, Donglin Zeng, A~John Rush, and Michael~R Kosorok.
\newblock Estimating individualized treatment rules using outcome weighted learning.
\newblock \emph{Journal of the American Statistical Association}, 107\penalty0 (499):\penalty0 1106--1118, 2012.

\bibitem[Zhao et~al.(2009)Zhao, Kosorok, and Zeng]{zhao2009reinforcement}
Yufan Zhao, Michael~R Kosorok, and Donglin Zeng.
\newblock Reinforcement learning design for cancer clinical trials.
\newblock \emph{Statistics in \textsc{M}edicine}, 28\penalty0 (26):\penalty0 3294--3315, 2009.

\bibitem[Zhou et~al.(2022)Zhou, Zhu, and Qu]{zhou2022estimating}
Wenzhuo Zhou, Ruoqing Zhu, and Annie Qu.
\newblock Estimating optimal infinite horizon dynamic treatment regimes via pt-learning.
\newblock \emph{Journal of the American Statistical Association to appear}, 2022.

\bibitem[Zhou et~al.(2017)Zhou, Mayer-Hamblett, Khan, and Kosorok]{zhou2017residual}
Xin Zhou, Nicole Mayer-Hamblett, Umer Khan, and Michael~R Kosorok.
\newblock Residual weighted learning for estimating individualized treatment rules.
\newblock \emph{Journal of the American Statistical Association}, 112\penalty0 (517):\penalty0 169--187, 2017.

\end{thebibliography}
}


\renewcommand \thepart{}
\renewcommand \partname{}

\newpage
\appendix
\allowdisplaybreaks

\addcontentsline{toc}{section}{Appendix} 
\part{Appendix} 
\parttoc 


\section{Theoretical proof}


\subsection{Derivation of K matching potential outcome} \label{S:K-est}
In this appendix we prove the following proposition from Section~\ref{sec:K-IPWE}:

\vspace{1em}
\noindent
{\bf Proposition \ref{eq:k-IPWE}}  {\it  Under the \textit{SUTVA} and \textit{no unmeasured confounding} assumptions, the expected total reward under the target regime $\mathscr{D}$ with k number of matching stages equals
\[
	\mathbb{E}^{\mathscr{D}_{(k)}}[R] =  \mathbb{E}\left\{ \frac{R \cdot \mathbb{I}(\lvert\mathbf{A} \cap \mathscr{D}\rvert = k)}{\prod_{j=1}^T \pi_j(A_j | H_j)}\right\},
\]
and the corresponding maximizing regime $\Tilde{\mathscr{D}}_{(k)}$ is defined as 
\[
	\Tilde{\mathscr{D}}_{(k)} = \argmax_{\mathscr{D}} \mathbb{E}^{\mathscr{D}_{(k)}}\{R\}.
\]
} 

\vspace{1em}
\noindent
{\bf Proof}. Let $\tau = (x_1, a_1, x_2, a_2, ..., x_T, a_T, R) \sim P$ be the observed sequence. We want to evaluate the expected rewards under the restricted measure $P^{\mathscr{D}_{(k)}}$, where there are k number of stages matched with the target regime $\mathscr{D}$. Suppose we have a set of indices $\mathcal{K}$ that contains the K indices of the matching stage and given the past information $\Bar{x_t} = (x_1,...,x_t)$ and $\Bar{a}_{t-1} = (a_1,...,a_{t-1})$, the probability of assigning $a_t$ according to policy $\mathscr{D}_{(k)}$ is equal to 
\begin{equation}
    P^{\mathscr{D}_{(k)}}_t (a_t; \Bar{x_t}, \Bar{a_{t-1}}) \;|\; \mathcal{K} = \mathbb{I}(t \in \mathcal{K}) \cdot \mathbb{I}(a_t = d_t(\Bar{x_t}, \Bar{a_{t-1}})) + \mathbb{I}(t \notin \mathcal{K}) \cdot \mathbb{I}(a_t \neq d_t(\Bar{x_t}, \Bar{a_{t-1}}))
\end{equation}

\vspace{1em}
\noindent By Radon-Nikodym:
\begin{align}
     &\left.\frac{dP^{\mathscr{D}_{(k)}}(\tau)}{dP(\tau)}\middle| {\mathcal{K}}\right. \nonumber \\
     &=\frac{f_0(x_1) P^{\mathscr{D}_{(k)}}_1 (a_1 | x_1) P(x_2|x_1,a_1)P^{\mathscr{D}_{(k)}}_2 (a_2 | x_1, a_1)\cdot\cdot\cdot P(x_T|\Bar{X_{T-1}}, \Bar{a_{T-1}}) P^{\mathscr{D}_{(k)}}_T (a_T; \Bar{x_T}, \Bar{a_{t-1}})}{f_0(x_1)\pi_1(a_1|x_1)P(x_2|x_1,a_1)\pi_2(a_2|x_1,a_1,x_2)\cdot\cdot\cdot P(x_T|\Bar{X_{T-1}}, \Bar{a_{T-1}}) \pi_T(a_T|\Bar{x_T}, \Bar{a_{T-1}})} \nonumber \\
     &= \frac{\prod_{j=1}^T P^{\mathscr{D}_{(k)}}_j (a_j; \Bar{x_j}, \Bar{a_{j-1}})}{\prod_{j=1}^T \pi_j (a_j; \Bar{x_j}, \Bar{a_{j-1}})} \nonumber \\
     &= \frac{\prod_{j=1}^T  \left[\mathbb{I}(j \in \mathcal{K}) \cdot \mathbb{I}(a_j = d_j(\Bar{x_j}, \Bar{a_{j-1}})) + \mathbb{I}(j \notin \mathcal{K}) \cdot \mathbb{I}(a_j \neq d_j(\Bar{x_j}, \Bar{a_{j-1}}))\right]}{\prod_{j=1}^T \pi_j (a_j; \Bar{x_j}, \Bar{a_{j-1}})} \nonumber \\
     &= \frac{\prod_{j \in \mathcal{K}} \mathbb{I}(a_j = d_j(\Bar{x_j}, \Bar{a_{j-1}})) \cdot \prod_{j \notin \mathcal{K}}\mathbb{I}(a_j \neq d_j(\Bar{x_j}, \Bar{a_{j-1}}))}{\prod_{j=1}^T \pi_j (a_j; \Bar{x_j}, \Bar{a_{j-1}})} \nonumber 
\end{align}

\vspace{1em}
\noindent Applying the importance weight correction:
{\small
\begin{align}
    &\mathbb{E}_{\tau \sim P^{\mathcal{D}_{(k)}}} [R] \nonumber \\
    &= \mathbb{E}_{\mathcal{K}} \left[ \mathbb{E}_{\tau \sim P^{\mathcal{D}_{(k)}}} [R \;\middle|\; \mathcal{K}] \right] \nonumber \\
    &= \mathbb{E}_{\mathcal{K}} \left[\mathbb{E}_{\tau \sim P} \left[ \frac{dP^{\mathscr{D}_{(k)}}(\tau)}{dP(\tau)} \cdot R \;\middle|\; a_{1:T} \sim \pi, \mathcal{K}\right] \right]\nonumber \\
    &= \mathbb{E}_{\mathcal{K}} \left[\mathbb{E}_{\tau \sim P} \left[ \frac{\prod_{j \in \mathcal{K}} \mathbb{I}(a_j = d_j(\Bar{x_j}, \Bar{a_{j-1}})) \cdot \prod_{j \notin \mathcal{K}}\mathbb{I}(a_j \neq d_j(\Bar{x_j}, \Bar{a_{j-1}}))}{\prod_{j=1}^T \pi_j (a_j; \Bar{x_j}, \Bar{a_{j-1}})} \cdot R \;\middle|\; a_{1:T} \sim \pi, \quad \mathcal{K}\right] \right]\nonumber \\
    &= {{T}\choose{k}}^{-1} \sum_{\mathcal{K}_j} \mathbb{E}_{\tau \sim P} \left[ \frac{\prod_{j \in \mathcal{K}_j} \mathbb{I}(a_j = d_j(\Bar{x_j}, \Bar{a_{j-1}})) \cdot \prod_{j \notin \mathcal{K}_j}\mathbb{I}(a_j \neq d_j(\Bar{x_j}, \Bar{a_{j-1}}))}{\prod_{j=1}^T \pi_j (a_j; \Bar{x_j}, \Bar{a_{j-1}})} \cdot R \;\middle|\; a_{1:T} \sim \pi\right] \nonumber \\
    &={{T}\choose{k}}^{-1} \cdot \mathbb{E}_{\tau \sim P} \left[ \frac{\sum_{\mathcal{K}_j} \left\{\prod_{j \in \mathcal{K}_j} \mathbb{I}(a_j = d_j(\Bar{x_j}, \Bar{a_{j-1}})) \cdot \prod_{j \notin \mathcal{K}_j}\mathbb{I}(a_j \neq d_j(\Bar{x_j}, \Bar{a_{j-1}})) \right\}} {\prod_{j=1}^T \pi_j (a_j; \Bar{x_j}, \Bar{a_{j-1}})} \cdot R \;\middle|\; a_{1:T} \sim \pi\right] \nonumber \\
    &= {{T}\choose{k}}^{-1} \cdot \mathbb{E}_{\tau \sim P} \left[ \frac{ \mathbb{I}(\sum_{j=1}^T \mathbb{I}(a_j = d_j(\Bar{x}_j, \Bar{a}_{j-1})) = k)} {\prod_{j=1}^T \pi_j (a_j; \Bar{x_j}, \Bar{a_{j-1}})} \cdot R \;\middle|\; a_{1:T} \sim \pi\right] \label{eq:equivalence}\\
    &= c_0 \cdot \mathbb{E}_{\tau \sim P} \left[ \frac{\mathbb{I}(\mathbf{a} \cap \mathbf{d} = k)}{\prod_{j=1}^T \pi_j (a_j; \Bar{x_j}, \Bar{a_{j-1}})} \cdot R \;\middle|\; a_{1:T} \sim \pi\right]
\end{align}
}
\vspace{1em}
\noindent Notice that \eqref{eq:equivalence} satisfies because there is only one set of K indices ($\mathcal{K}^*$) making the product $\prod_{j \in \mathcal{K}^*} \cdot \prod_{j \notin \mathcal{K}^{*}}$ equal to 1. Then the if and only if relationship is obvious.


\vspace{3em}

\subsection{Derivation of SAL estimators} \label{S:SAL}\
In this appendix we prove the following SAL value function from Section~\ref{sec:SAL}:
\[
     V^{SA}(\mathscr{D})=  \sum_{k=0}^T   \frac{k}{T} \cdot \mathbb{E}^{\mathscr{D}_{(k)}} \{R\} = \mathbb{E} \left\{ \frac{R \cdot \frac{1}{T} \sum_{j=1}^T \mathbb{I}(A_j = D_j(H_j))}{\prod_{j=1}^T \pi_j(A_j | H_j)}\right\},
\]

\vspace{1em}
\noindent
{\bf Proof}. We first take the iterated expectation of each level of $k$-IPWE:

\begin{align}
    \mathbb{E}_{K} \left[\mathbb{E}^{\mathscr{D}_{(k)}} [R] \right] &= \sum_{k=0}^T \mathbb{E}^{\mathscr{D}_{(k)}} [R] \cdot  P(K = k)  \nonumber \\
    &= \sum_{k=0}^T \mathbb{E}\left[\frac{R \cdot \mathbb{I}(K = k)}{\prod \pi_j(A_j | H_j)} \right] \cdot  P(K = k)   \nonumber\\
    &= \sum_{k=0}^T \mathbb{E} \left[\frac{R}{\prod \pi_j(A_j | H_j)} \;\middle|\; K=k \right] \cdot P(K = k) \cdot P(K = k)  \nonumber \\
    &= \sum_{k=0}^T \mathbb{E} \left[\frac{R}{\prod \pi_j(A_j | H_j)} \;\middle|\; K=k \right] \cdot \phi(k) \cdot P(K = k)  \nonumber \\
    &= \sum_{k=0}^T \mathbb{E} \left[\frac{R \cdot \phi(k) }{\prod \pi_j(A_j | H_j)} \;\middle|\; K=k \right] \cdot P(K = k)  \nonumber \\
    &= \sum_{k=0}^T \mathbb{E} \left[\frac{R \cdot \phi(K)}{\prod \pi_j(A_j | H_j)} \;\middle|\; K=k \right] \cdot P(K = k)  \nonumber \\
    &= \mathbb{E} \left[\frac{R \cdot \phi(K)}{\prod \pi_j(A_j | H_j)} \right]  \nonumber \\
    &= \mathbb{E} \left[\frac{R \cdot \phi \left( \sum_{j=1}^T \mathbb{I}(A_j = D_j) \right)}{\prod \pi_j(A_j | H_j)} \right]
    \nonumber
\end{align}

Since the SAL takes a linear weighting function $\phi(k) = k / T$ with respect to the matching number $k$, it is obvious to show:
\begin{equation}
     V^{SA}(\mathscr{D})= \sum_{k=0}^T \mathbb{E}^{\mathscr{D}_{(k)}} [R] \cdot  P(K = k) = \sum_{k=0}^T \mathbb{E}^{\mathscr{D}_{(k)}} [R] \cdot  \frac{k}{T} = \mathbb{E} \left[\frac{R \cdot \frac{1}{T} \sum_{j=1}^T \mathbb{I}(A_j = D_j)}{\prod \pi_j(A_j | H_j)} \right]
    \nonumber
\end{equation}


\vspace{3em}
\subsection{Proof of Theorem 1 - Fisher consistency} \label{S:Thm-1}
In this appendix we prove the following Theorem from Section~\ref{sec:fisher-1}:

\vspace{1em}
\noindent
{\bf Theorem \ref{thm:fisher-1}}  {\it Let $V_1$, $V_2$, and $\mathcal{F}$ be three functions mapping  $\mathbf{H}$ onto $\mathbb{R}$. Besides, $\mathcal{F}$ is a functional space closed under complement. Assume the surrogate function $\psi(a, f; \theta): \mathcal{A} \times \mathcal{F} \mapsto \mathbb{R}$ satisfies that $\psi(a, f; \theta) = \psi(-a, -f; \theta)$ and $sign(\psi(1, f; \theta) - \psi(-1, f; \theta)) = sign(f)$. Then if $f^* \in \mathcal{F}$ maximizes $T(f) = \psi(1,f)V_1 + \psi(-1,f) V_2$, the sign of $f^*$ aligns with the sign of $V_1 - V_2$ (i.e., $\text{sign}(f^*) = \text{sign}(V_1 - V_2)$).
} \hfill

\vspace{1em}
\noindent
{\bf Proof}.  The objective of this theorem is consisted of two steps. First, we want to find a sequence of functions $(f_{\psi 1}^*,...,f_{\psi T}^*)$ that maximizes the surrogate version of SWL value function; and second, verify that the sign of obtained optimizers is aligned with the treatment decision $(f^*_1,...,f^*_T)$ that maximizes $V^{SW}$. We will illustrate the proof on any arbitrary stage t,  where $1 \le t \le T$. To begin with, consider the maximization results of the proposed SWL algorithm on the $t^{th}$ stage given the history information $H_t \in \mathcal{H}_t$. Notice that as $H_t$ summarizes all past information up to the $t^{th}$ stage, the treatment results $\{A_j\}_{j=1}^t$ and covariates information $\{H_j\}_{j=1}^y$ from past stages are treated as known constants. Thus,

\begin{align}
    f_{\psi t}^*&= \argmax_{(f_1,...,f_T) \in \mathcal{F}^{|T|}} \mathbb{E} \left[ \frac{R \cdot \sum_{j=1}^T \omega_j \cdot \psi(A_j, f_j(H_j))}{\prod_{j=1}^T \pi_j (a_j | H_j)} \;\middle|\; H_t, \{\omega_j\}_{j=1}^T = \{\hat{\omega}_j\}_{j=1}^T \right] \nonumber \\
    &= \argmax_{f_t \in \mathcal{F}} \mathbb{E} \left[ \frac{R \cdot \hat{\omega}_t \cdot \psi(A_t, f_t(H_t))}{\prod_{j=1}^T \pi_j (a_j | H_j ) } \;\middle|\; H_t \right] \label{eq:fisher-breakdown} \\
    &= \argmax_{f_t \in \mathcal{F}} \sum_{a_t \in \{-1, 1\}} \mathbb{E} \left[ \frac{R \cdot \hat{\omega}_t \cdot \psi(A_t, f_t(H_t))}{\prod_{j=t+1}^T \pi_j (a_j | H_j )} \;\middle|\; H_t, A_t=a_t\right] \nonumber \\ 
    &= \argmax_{f_t \in \mathcal{F}} \sum_{a_t \in \{-1, 1\}} \psi(a_t, f_t(H_t)) \cdot \mathbb{E} \left[ R \;\middle|\; H_t, A_t=a_t\right] \label{eq:fisher-randomTrial} \\ 
    &= \argmax_{f_t \in \mathcal{F}} \psi(1, f_t(H_t)) \cdot V_{t1}(H_t) + \psi(-1, f_t(H_t)) \cdot V_{t2}(H_t), \label{eq:fisher-lemmaform}
\end{align}

\noindent where $V_{t1}(H_t) = \mathbb{E} \left[ R \;\middle|\; H_t, A_t=1 \right]$ and $V_{t2}(H_t) = \mathbb{E} \left[ R \;\middle|\; H_t, A_t=-1 \right] \nonumber$. An important property of the summation operator in our proposed SWL method is that optimizing $f_t$ at one stage will not affect the final results of decision rules at other stages. Thus, we could break down the optimation steps into T-individual sub-tasks as could be seen from the second equation \eqref{eq:fisher-breakdown}. Besides, according to the random trial property inherited from the SMART study design, the propensities are independent from the covariate information $\mathbf{H}$ and thus could be dropped as constants for all future steps in the fourth equation \eqref{eq:fisher-randomTrial}. Lastly, continued from the last equation \eqref{eq:fisher-lemmaform}, we could utilize Lemma \ref{lemma:fisher} to verify the sign of the obtained maximizer $f_{\psi t}^*$ at the $t^{th}$ stage. Assume the surrogate function $\psi$ satisfies the conditions of Lemma \ref{lemma:fisher}, we obtained

\begin{align}
    &\text{sign}(f_{\psi t}^*(H_t)) \nonumber \\
    &= \text{sign} \left\{\mathbb{E} \left[\sum_{j=1}^T r_j \;\middle|\; H_t, A_t=1 \right]  -\mathbb{E} \left[\sum_{j=1}^T r_j \;\middle|\; H_t, A_t=-1 \right] \right\} \nonumber \\
    &= \text{sign} \left\{\mathbb{E} \left[r_t + \sum_{j=t+1}^T r_j \;\middle|\; H_t, A_t=1 \right] - \mathbb{E} \left[r_t + \sum_{j=t+1}^T r_j \;\middle|\; H_t, A_t=-1\right] \right\} \nonumber \\
    &= \argmax_{a_t \in \{-1,1\}} \mathbb{E} \left[r_t + \sum_{j=t+1}^T r_j \;\middle|\; H_t, A_t=a_t \right] \label{eq:fisher-sign}
\end{align}

\noindent The result from equation \eqref{eq:fisher-sign} once again utilizes the fact that the information of past rewards is encapsulated in the history variable $H_t$. Now, since the indicator function $\mathbb{I}(af > 0)$ also satisfies the conditions of Lemma \ref{lemma:fisher}, we are able to establish the fisher consistency between $V^{SW}_\psi$ and $V^{SW}$ on any arbitrary $t^{th}$ stage as:

\begin{align}
    &\text{sign}(f_t^*(H_t)) \nonumber \\
    &= \text{sign} \left\{\mathbb{E} \left[r_t + \sum_{j=t+1}^T r_j \;\middle|\; H_t, A_t=1 \right]  -  \mathbb{E} \left[r_t + \sum_{j=t+1}^T r_j \;\middle|\; H_t, A_t=-1\right] \right\} \nonumber \\
    &= \text{sign}(f_{\psi t}^*(H_t)) 
\end{align}


\vspace{2em}
\begin{lemma}
Let $V_1$, $V_2$, and $\mathcal{F}$ be three functions mapping  $\mathbf{H}$ onto $\mathbb{R}$. Besides, $\mathcal{F}$ is a functional space closed under complement. Assume the surrogate function $\psi(a, f; \theta): \mathcal{A} \times \mathcal{F} \mapsto \mathbb{R}$ satisfies that $\psi(a, f; \theta) = \psi(-a, -f; \theta)$ and $sign(\psi(1, f; \theta) - \psi(-1, f; \theta)) = sign(f)$. Then if $f^* \in \mathcal{F}$ maximizes $T(f) = \psi(1,f)V_1 + \psi(-1,f) V_2$, the sign of $f^*$ aligns with the sign of $V_1 - V_2$ (i.e., $\text{sign}(f^*) = \text{sign}(V_1 - V_2)$).
\label{lemma:fisher}
\end{lemma}

\vspace{1em}
\noindent
{\bf Proof of Lemma \ref{lemma:fisher}}. To find the $f^* \in \mathcal{F}$ that maximizes $T(f)$, we consider a reduced function space $\Tilde{\mathcal{F}} = \{\argmax_{\Tilde{f} \in \{f, -f\}} T(\Tilde{f})\}_{f \in \mathcal{F}}$. Since $\mathcal{F}$ is closed under complement, the contrast function $-f$ exists in $\mathcal{F}$. Hence, we first compare $T(f)$ with $T(-f)$ and construct $\Tilde{\mathcal{F}}$ by adding one of the $f$ and $-f$ that yields larger $T$ value. By construction, $|\Tilde{F}| = \frac{|\mathcal{F}|}{2}$, and the following comes naturally: 

\begin{equation}
    f^* = \argmax_{f \in \mathcal{F}} T(f) = \argmax_{\Tilde{f} \in \Tilde{\mathcal{F}}} T(\Tilde{f}) \nonumber
\end{equation}

\noindent Next, for each $f \in \mathcal{F}$, we find its corresponding $\Tilde{f} = \argmax_{\Tilde{f} \in \{f, -f\}} T(\Tilde{f})$ from $T(f) - T(-f)$:
\begin{align}
    T(f) - T(-f) &= \left[ \psi(1, f)V_1 + \psi(-1. f)V_2\right] - \left[ \psi(1, -f)V_1 + \psi(-1. -f)V_2\right] \nonumber \\
    &= \psi(1, f)(V_1 - V_2) - \psi(1, -f)(V_1-V_2) \nonumber \\
    &=  \left[ \psi(1, f) - \psi(1, -f)\right] (V_1 - V_2) \nonumber
\end{align}

\noindent which implies, $\text{sign}( T(f) - T(-f) ) = \text{sign}(f) * \text{sign}(V_1 - V_2)$. Based on the sign of $f$ and $V_1 - V_2$, following two cases can be discussed:
\begin{equation}
    \argmax_{\Tilde{f} \in \{f, -f\}} T(\Tilde{f}) = \begin{cases}
        f & \text{if }  \text{sign}(f) = \text{sign}(V_1 - V_2)\\
        -f & \text{if } \text{sign}(f) \ne \text{sign}(V_1 - V_2)\\ 
    \end{cases}\nonumber
\end{equation}

\noindent In either cases, 
\begin{equation}
    \text{sign}(\Tilde{f}) = \text{sign}(\argmax_{\Tilde{f} \in \{f, -f\}} T(\Tilde{f})) = \text{sign}(V_1 - V_2) \nonumber
\end{equation}

\noindent Since $\forall \Tilde{f} \in \Tilde{\mathcal{F}}$ and $\forall H \in \mathbf{H}$, $\text{sign}(\Tilde{f}) = \text{sign}(V_1(H) - V_2(H))$ and $f^* \in \Tilde{\mathcal{F}}$, we have shown that $\text{sign}(f^*) = \text{sign}(V_1(H) - V_2(H))$.  

\vspace{1em}
\noindent In the following, we present that the indicator function, logistic function, as well as binary cross-entropy all satisfy the $\psi$ function assumption.

\vspace{1em}
\noindent $\mathbf{1}.\quad \psi(a, f) = \mathbb{I}(af>0)$ 
\begin{flalign*}
&\psi(-a,-f) = \mathbb{I}((-a)(-b) > 0) = \mathbb{I}(af>0) = \psi(a,b) & \nonumber\\
&\text{sign}(\psi(1,f) - \psi(-1,f)) = \text{sign}(\mathbb{I}(f>0) - \mathbb{I}(f<0)) = \text{sign}(f) \nonumber
\end{flalign*}

\vspace{1em}
\noindent $\mathbf{2}.\quad \psi(a,f; \lambda) = (e^{-\lambda af} + 1)^{-1}$ 
\begin{flalign*}
&\psi(-a,-f) = (e^{-\lambda (-a)(-f)} + 1)^{-1} = (e^{-\lambda af} + 1)^{-1} = \psi(a,b) &\nonumber \\
&\text{sign}(\psi(1,f) - \psi(-1,f)) = \text{sign}((e^{-\lambda f)} + 1)^{-1} - (e^{\lambda f} + 1)^{-1}) = \text{sign}\left(\frac{e^{\lambda f} - 1}{e^{\lambda f} + 1}\right) = \text{sign}(f) \nonumber
\end{flalign*}

\vspace{1em}
\noindent $\mathbf{3}.\quad \psi(a,f; \lambda) = -\frac{a+1}{2} log(e^{-f}+1) - (1-\frac{a+1}{2})log(e^f+1)$
\begin{flalign*}
    &\psi(-a,-f) = -\frac{-a+1}{2} log(e^{f}+1) - (1-\frac{-a+1}{2})log(e^{-f}+1) & \nonumber \\
    &\qquad \qquad \quad =  -\frac{a+1}{2}log(e^{-f}+1) - (1-\frac{a+1}{2})log(e^f+1) = \psi(a,b) & \nonumber \\
    &\text{sign}(\psi(1,f) - \psi(-1,f)) = \text{sign}(-log(e^{-f}+1)+log(e^f+1)) = \text{sign}(log(e^f)) = \text{sign}(f) \nonumber
\end{flalign*}

\vspace{3em}
\subsection{Proof of Theorem 2 - Fisher consistency with optimal} 
\label{S:FC-2}

In this appendix we prove the following Theorem from Section~\ref{sec:FC-2}:

\vspace{1em}
\noindent
{\bf Theorem \ref{thm: optimal-fisher}}  {\it For all stages $t=1,...,T$ and $H_t \in \mathscr{H}_t$, $\text{sign}(f_{\psi t}^*) = \text{sign}(f_{t}^*) = d_t^*$ if and only if Stage $t$ is dominated by the optimal treatment effect.
} \hfill

\vspace{2em}
\noindent
{\bf Proof}.  
After expanding equation \eqref{eq:fisher-sign}, $\text{sign}(f_{\psi t}^*)$ at stage t can be rewritten as,
{\small
\begin{align}
    &\text{sign}(f_{\psi t}^*) \nonumber \\
    &= \argmax_{a_t \in \{-1, 1\}} \mathbb{E} \left[ r_t + \sum_{j=t+1}^T r_j \;\middle|\; H_t, A_t = a_t\right] \nonumber \\
    &= \argmax_{a_t \in \{-1, 1\}} \mathbb{E} \left[ r_t + \sum_{j=t+1}^T r_j \;\middle|\; H_t, A_t = a_t, A_{(t+1):T} = d^*_{(t+1):T}\right] \cdot P(A_{(t+1):T} = d^*_{(t+1):T} \;|\; H_t, A_t=a_t) + \nonumber \\
    & \hspace{4em} \mathbb{E} \left[ r_t + \sum_{j=t+1}^T r_j \;\middle|\; H_t, A_t = a_t, A_{(t+1):T} \neq d^*_{(t+1):T}\right] \cdot P(A_{(t+1):T} \neq d^*_{(t+1):T} \;|\; H_t, A_t=a_t) \nonumber \\ 
    &= \argmax_{a_t \in \{-1, 1\}}  \mathbb{E} \left[ r_t + \sum_{j=t+1}^T r_j \;\middle|\; H_t, A_t = a_t, A_{(t+1):T} =  d^*_{(t+1):T}\right] \cdot \left( 1 - P(A_{(t+1):T} \neq d^*_{(t+1):T} \;|\; H_t, A_t=a_t) \right) + \nonumber \\
    & \hspace{4em} \mathbb{E} \left[ r_t + \sum_{j=t+1}^T r_j \;\middle|\; H_t, A_t = a_t, A_{(t+1):T} \neq d^*_{(t+1):T}\right] \cdot P(A_{(t+1):T} \neq d^*_{(t+1):T} \;|\; H_t, A_t=a_t) \\ 
    &= \argmax_{a_t \in \{-1, 1\}} R^*(A_t=a_t, H_t) + \left( R^*(A_t=a_t, H_t) -  R^\dagger(A_t=a_t, H_t) \right) \cdot P^\dagger(A_t=a_t, H_t), \label{eq:31}
\end{align}
}
\hspace{-1em} where in step \eqref{eq:31}, we denote $\mathbb{E} \left[ r_t + \sum_{j=t+1}^T r_j \;\middle|\; H_t, A_t = a_t, A_{(t+1):T} =  d^*_{(t+1):T}\right]$ to be $R^*(A_t=a_t, H_t)$, which represents the optimal expected reward achievable after assigning treatment $a_t$ given health variable $H_t$ at the current stage $t$ while assuming all future treatments are assigned optimally. Additionally, we let $R^\dagger(A_t=a_t, H_t)$ denote $ \mathbb{E} \left[ r_t + \sum_{j=t+1}^T r_j \;\middle|\; H_t, A_t = a_t, A_{(t+1):T} \neq d^*_{(t+1):T}\right]$ to represent the expected total rewards if some of the future assignments are not optimal after assigning $a_t$ at the current stage. Correspondingly, $P^\dagger(A_t=a_t, H_t)$ denotes the possibility of assigning any sub-optimal treatment in future stages. Given the notations, to let $\text{sign}(f_{\psi t}^*) = d^*_t$, it is trivial to show that the optimal $d^*_t$ must maximize Equation \eqref{eq:31} and fulfill this condition.

\vspace{3em}
\subsection{Proof of Theorem 3 - Finite-sample performance error bound} 
\label{S:Thm-3}

In this appendix we prove the following Theorem from Section~\ref{sec:thm-3}:

\vspace{1em}
\noindent
{\bf Theorem \ref{theorem:performance-bound}}  {\it Under Assumptions \ref{A-reward}-\ref{assumption:Metric-Entropy}, there exist constants $C_1 > 0$ and $0 < \alpha < 1$ such that for any $\delta \in (0, 1)$, w.p. at least $1-\delta$, the performance error is upper-bounded by:
\[
\left|V^{SW}(\mathbf{f}^*) - \hat{V}^{SW}_\psi(\hat{\mathbf{f}}_n) \right| \le  \underbrace{\frac{M}{c_0^T} \sum_{j=1}^T \omega_j \epsilon_{n,j}}_{\text{Surrogate error}}  + \underbrace{\frac{6(\alpha+1)}{\alpha} \left[\alpha C_1 \sqrt{\frac{T}{n}} \left( \frac{\lambda M}{4c_0^T}\right)^\alpha \right]^{\frac{1}{\alpha + 1}} + \frac{9M}{c_0^T} \sqrt{\frac{\log 2 / \delta}{2n}}}_{\text{Empirical estimation error}},
\]
where $\epsilon_{n,j} = \sup_{A_j, H_j}\left| \mathbb{I}(A_jf_j(H_j) > 0) - \psi(A_jf_j(H_j); \lambda_n) \right|$.

} \hfill

\vspace{1em}
\noindent {\bf Proof}.  
To show the performance bond between $V^{SW}(\mathbf{f}^*)$ and $\hat{V}^{SW}_\psi(\hat{\mathbf{f}}_n)$, we first notice that the performance bound could be separated into two bounds: the concentration bounds between $V^{SW}$ and $V^{SW}_\psi$, as well as the empirical estimation error bound of $V^{SW}_\psi$.
\begin{align}
    &\left|V^{SW}(\mathbf{f}^*) - \hat{V}^{SW}_\psi(\hat{\mathbf{f}}_n) \right| \nonumber \\
    &\le \left| V^{SW}(\mathbf{f}^*) - V^{SW}_\psi(\mathbf{f}^*)\right| + \left|V^{SW}_\psi(\mathbf{f}^*) - \hat{V}^{SW}_\psi(\hat{\mathbf{f}}_n) \right| \nonumber \\
    &\le \left| V^{SW}(\mathbf{f}^*) - V^{SW}_\psi(\mathbf{f}^*)\right| + \left|V^{SW}_\psi(\mathbf{f}^*) - V^{SW}_\psi(\hat{\mathbf{f}}_n) \right| + \left|V^{SW}_\psi(\hat{\mathbf{f}}_n) - \hat{V}^{SW}_\psi(\hat{\mathbf{f}}_n) \right| \nonumber \\
    &\le \left| V^{SW}(\mathbf{f}^*) - V^{SW}_\psi(\mathbf{f}^*)\right| + \left|V^{SW}_\psi(\mathbf{f}^*) -  V^{SW}_\psi(\hat{\mathbf{f}}_n) + \hat{V}^{SW}_\psi(\hat{\mathbf{f}}_n) - \hat{V}^{SW}_\psi(\mathbf{f}^*)\right| + \nonumber \\
    &\quad \; \left|V^{SW}_\psi(\hat{\mathbf{f}}_n) - \hat{V}^{SW}_\psi(\hat{\mathbf{f}}_n) \right| \\
    &\le \left| V^{SW}(\mathbf{f}^*) - V^{SW}_\psi(\mathbf{f}^*)\right|  +  2\sup_{\mathbf{f} \in \mathcal{F}^{|T|}} |V^{SW}_\psi(\mathbf{f}) - \hat{V}^{SW}_\psi(\mathbf{f})| + \sup_{\mathbf{f} \in \mathcal{F}^{|T|}} |V^{SW}_\psi(\mathbf{f}) - \hat{V}^{SW}_\psi(\mathbf{f})| \nonumber \\
    &=   \underbrace{\left| V^{SW}(\mathbf{f}^*) - V^{SW}_\psi(\mathbf{f}^*)\right|}_{\text{Surrogate error bound}} + \underbrace{3\sup_{\mathbf{f} \in \mathcal{F}^{|T|}} \left|V^{SW}_\psi(\mathbf{f}) - \hat{V}^{SW}_\psi(\mathbf{f})\right|}_{\text{Empirical estimation error bound}} 
\label{eq:performance-bound-breakdown}
\end{align}
 
\noindent According to the results \eqref{eq:performance-bound-breakdown}, the performance bound could be obtained once the concentration and estimation error bounds are estimated. We will bound each component in the following two subsections.

\subsubsection{Surrogate error bound} 

First, we show that for any regime $\mathbf{f} \in \mathcal{F}^{|T|}$ and surrogate function $\psi(x;\lambda)$, the concentration bound is upper bounded by,
\begin{align}
&\left|V^{SW}(\mathbf{f}) - V^{SW}_\psi (\mathbf{f})\right| \nonumber \\
&= \left|\mathbb{E} \left[ \frac{R}{\prod_{j=1}^T \pi_j} \sum_{j=1}^T \omega_j \left(\mathbb{I}(A_jf_j(H_j) > 0) - \psi(A_jf_j(H_j); \lambda) \right) \right]\right| \nonumber \\
    &\le \mathbb{E} \left[ \frac{R}{\prod_{j=1}^T \pi_j} \sum_{j=1}^T \omega_j \left| \mathbb{I}(A_jf_j(H_j) > 0) - \psi(A_jf_j(H_j); \lambda) \right| \right] \nonumber \\
    &\le \frac{M}{c_0^T} \sum_{j=1}^T \omega_j \sup_{A_j, H_j}\left| \mathbb{I}(A_jf_j(H_j) > 0) - \psi(A_jf_j(H_j); \lambda) \right|.
\label{proof:concentration-bound}
\end{align}

\subsubsection{Empirical estimation error bound}

Next, we would like to show the empirical estimation error bound. To begin with, we denote the observed data to be $\Omega_N = \{(\underline{X}_i, \underline{A}_i, R_i, \underline{\pi}_i)\}_{i=1}^N$. Besides, based on the observed data, we consider a class of functions $\mathcal{G}$ such that $\hat{V}^{SW}_{\psi}(\Omega_N) = \frac{1}{N} \sum_{i=1}^N g(\Omega_{Ni})$ for $g \in \mathcal{G}$. To be specific, the connection between $g$ and the decision rules $\mathbf{f}$ is that $g(\underline{X}, \underline{A}, \underline{\pi}, R; \mathbf{f}) = \frac{R}{\prod_{j=1}^T \pi_j} \sum_{j=1}^T \omega_j \cdot (e^{-\lambda \cdot A_j f_j(H_j)} + 1)^{-1}$. Then, according to the Lemma \ref{Lemma:Rademacher} and Lemma \ref{Lemma:Metric-Entropy}, with probability at least $1 - \delta$, the estimation error bound could be upper bounded by,
\begin{equation}
    \sup_{\mathbf{f} \in \mathcal{F}^{|T|}} \left|V^{SW}_\psi(\mathbf{f}) - \hat{V}^{SW}_\psi(\mathbf{f})\right| \le 2 \cdot (\alpha^{-1} + 1) \cdot \left[\alpha C_1 \sqrt{\frac{T}{N}} \left( \frac{\lambda M}{4c_0^T}\right)^\alpha \right]^{1/(\alpha + 1)} + \frac{3M}{c_0^T} \sqrt{\frac{1}{2N} \log \frac{2}{\delta}}
\label{proof:estimation-error-bound}
\end{equation}

\noindent At last, combining \eqref{proof:concentration-bound} and \eqref{proof:estimation-error-bound}, we are able to present the performance bound.

\begin{lemma}
Let $\{(\underline{X}_i, \underline{A}_i, R_i, \underline{\pi}_i)\}_{i=1}^N$ be iid random draws from $\Omega=\mathcal{X}^{T} \times \mathcal{A}^{T} \times \mathbb{R}_{(0, M]} \times \mathbb{R}_{[c_0, c_1]}^T$ that compose the observed data $\Omega_N$. Consider a class of functions $\mathcal{G} = \{g \in \mathcal{G}: \Omega \mapsto \mathbb{R}\}$ such that $\hat{V}^{SW}_{\psi}(\Omega_N) = \frac{1}{N} \sum_{i=1}^N g(\Omega_{Ni})$. Then for any $\delta \in (0, 1)$, with probability at least $1 - \delta$, we have:
\begin{equation}
    \sup_{\mathbf{f} \in \mathcal{F}^{|T|}} |V^{SW}_\psi(\mathbf{f}) - \hat{V}^{SW}_\psi(\mathbf{f}; \Omega_N)| \le 2\hat{\mathcal{R}}_n(\mathcal{G}; \Omega_N) + \frac{3M}{c_0^T} \sqrt{\frac{1}{2N} \log \frac{2}{\delta}}
\end{equation}
\label{Lemma:Rademacher}
\end{lemma}

\noindent \textbf{Proof of Lemma \ref{Lemma:Rademacher}.} This Lemma aims to upper bound the empirical estimation error bound with a Rademacher generalization bound. For ease of denotations, we let $\eta(\Omega_N) = \sup_{\mathbf{f} \in \mathcal{F}^{|T|}} |V^{SW}_\psi(\mathbf{f}) - \hat{V}^{SW}_\psi(\mathbf{f}; \Omega_N)|$. Besides, we define $\Omega_N^{'}$ as a duplicate of $\Omega_N$ except one sample k with observed values $(\underline{X}_k^{'}, \underline{A}_k^{'}, R_k^{'}, \underline{\pi}_k^{'})$. Assume we have estimated the summary function $\{\hat{\mathbf{S}}_j\}_{j=1}^T$ and obtained the history information $\{\hat{H}_j\}_{j=1}^T = \{\hat{\mathbf{S}}_j(X_1,A_1,...,A_{j-1},X_j)\}_{j=1}^T$, then we can first obtain an upper bound of $\eta$ based on its expected value $\mathbb{E}\eta$ as,

\begin{align}
&|\eta(\Omega_N) - \eta(\Omega_N^{'})| \nonumber \\
&= \left |\sup_{\mathbf{f} \in \mathcal{F}^{|T|}}|V^{SW}_\psi(\mathbf{f}) - \hat{V}^{SW}_\psi(\mathbf{f}, \Omega_N)| - \sup_{\mathbf{f} \in \mathcal{F}^{|T|}}|V^{SW}_\psi(\mathbf{f}) - \hat{V}^{SW}_\psi(\mathbf{f}, \Omega_N^{'})| \right| \nonumber\\
&\le \sup_{\mathbf{f} \in \mathcal{F}^{|T|}} \left| |V^{SW}_\psi(\mathbf{f}) - \hat{V}^{SW}_\psi(\mathbf{f}, \Omega_N)| - |V^{SW}_\psi(\mathbf{f}) - \hat{V}^{SW}_\psi(\mathbf{f}, \Omega_N^{'})| \right|\nonumber \\
&\le \sup_{\mathbf{f} \in \mathcal{F}^{|T|}} \left| V^{SW}_\psi(\mathbf{f}) - \hat{V}^{SW}_\psi(\mathbf{f}, \Omega_N) - V^{SW}_\psi(\mathbf{f}) + \hat{V}^{SW}_\psi(\mathbf{f}, \Omega_N^{'})| \right|\nonumber \\
&= \left |V^{SW}_\psi(\Tilde{\mathbf{f}}) - \hat{V}^{SW}_\psi(\Tilde{\mathbf{f}}, \Omega_N) - V^{SW}_\psi(\Tilde{\mathbf{f}}) + \hat{V}^{SW}_\psi(\Tilde{\mathbf{f}}, \Omega_N^{'}) \right| \nonumber \\
&= \left | \frac{1}{N}\sum_{i=1}^N \frac{R_i \cdot  \sum_{j=1}^T \omega_j (e^{-\lambda A_j \Tilde{f}_j(\hat{H}_{ij})} + 1)^{-1}}{\prod_{j=1}^T \pi_{ij}}  - \frac{1}{N}\sum_{i=1}^N \frac{R_i^{'} \cdot \sum_{j=1}^T \omega_j (e^{-\lambda A_j^{'} \Tilde{f}_j(\hat{H}_{ij}^{'})}+ 1)^{-1}}{\prod_{j=1}^T \pi_{ij}^{'}} \right | \nonumber \\
&= \frac{1}{N} \left| \frac{R_k \cdot \sum_{j=1}^T \omega_j (e^{-\lambda A_j \Tilde{f}_j(\hat{H}_{kj})} + 1)^{-1}}{\prod_{j=1}^T \pi_{kj}}   - \frac{R_k^{'} \cdot \sum_{j=1}^T \omega_j (e^{-\lambda A_j^{'} \Tilde{f}_j(\hat{H}_{kj}^{'})} + 1)^{-1}}{\prod_{j=1}^T \pi_{kj}^{'}}\right | \nonumber \\
&\le \frac{1}{N} \sum_{j=1}^T \omega_j \left| \frac{R_k}{\prod_{j=1}^T \pi_{kj}} \cdot \frac{1}{e^{-\lambda A_j \Tilde{f}_j(\hat{H}_{kj})} + 1}  - \frac{R_k^{'}}{\prod_{j=1}^T \pi_{kj}^{'}} \cdot \frac{1}{e^{-\lambda A_j^{'} \Tilde{f}_j(\hat{H}_{kj}^{'})} + 1} \right| \nonumber \\
&\le  \frac{1}{N} \sum_{j=1}^T \omega_j \cdot \text{max} \left( \frac{R_k}{\prod_{j=1}^T \pi_k}, \frac{R_k^{'}}{\prod \pi_k^{'}} \right) \nonumber \\
&\le \frac{1}{N} \sum_{j=1}^T \omega_j \frac{M}{c_0^T}  = \frac{M}{N \cdot c_0^T}.
\end{align}
By McDiarmid’s inequality, the following inequality holds:
\begin{equation}
    Pr(\eta - \mathbb{E}\eta \ge t) \le exp(-\frac{2t^2}{\sum_{i=1}^N c_i^2}) = exp(-\frac{2t^2}{\frac{M^2}{N\cdot c_0^{2T}}}) = exp(-\frac{c_0^{2T} \cdot 2t^2}{M^2 / N}). \nonumber
\end{equation}
Thus, by setting $\delta =  exp(-\frac{c_0^{2T} \cdot 2t^2}{M^2/N})$, with probability at least $1-\delta$, we obtained:
\begin{equation}
    \eta \le \mathbb{E} \eta + \frac{M}{c_0^T} \sqrt{\frac{\log 1/\delta}{2N}}
\label{eq:MC-1}
\end{equation}

\noindent The problem remained is to bound the expectation of the estimated error bound, $\mathbb{E}\eta$. Here, we utilize the Rademacher generalization bound of the function class $\mathcal{G}$. Additionally, we denote $\Tilde{\Omega}_N$ to be a ghost sample that follows the same distribution as $\Omega_N$. Then we can show that
\begin{align}
&\mathbb{E}[\eta(\Omega_N)] \nonumber \\
&= \mathbb{E}_{\Omega_N}\left[ \sup_{\mathbf{f} \in \mathcal{F}^{|T|}} \left|V^{SW}_\psi(\mathbf{f}) - \hat{V}^{SW}_\psi(\mathbf{f}, \Omega_N)\right|\right] \nonumber \\
&= \mathbb{E}_{\Omega_N}\left[ \sup_{\mathbf{f} \in \mathcal{F}^{|T|}} \left|\mathbb{E}_{\Tilde{\Omega}_N}\left[\hat{V}^{SW}_\psi(\mathbf{f}, \Tilde{\Omega}_N)\right] - \hat{V}^{SW}_\psi(\mathbf{f}, \Omega_N)\right|\right] \nonumber \\
&= \mathbb{E}_{\Omega_N}\left[ \sup_{\mathbf{f} \in \mathcal{F}^{|T|}} \left|\mathbb{E}_{\Tilde{\Omega}_N} \left[\hat{V}^{SW}_\psi(\mathbf{f}, \Tilde{\Omega}_N) - \hat{V}^{SW}_\psi(\mathbf{f}, \Omega_N) \right]\right|\right] \nonumber \\
& \le \mathbb{E}_{\Omega_N, \Tilde{\Omega}_N} \left[ \sup_{\mathbf{f} \in \mathcal{F}^{|T|}} \left|\hat{V}^{SW}_\psi(\mathbf{f}, \Tilde{\Omega}_N) - \hat{V}^{SW}_\psi(\mathbf{f}, \Omega_N)\right| \right] \nonumber \\
&= \mathbb{E}_{\Omega_N, \Tilde{\Omega}_N} \left[ \sup_{\mathbf{g} \in \mathcal{G}} \left| \frac{1}{N} \sum_{i=1}^N g(\Tilde{\Omega}_{Ni}; \mathbf{f}) - g(\Omega_{Ni}; \mathbf{f})\right| \right] \nonumber \\
&= \mathbb{E}_{\Omega_N, \Tilde{\Omega}_N, \sigma} \left[ \sup_{\mathbf{g} \in \mathcal{G}} \left| \frac{1}{N} \sum_{i=1}^N \sigma_i \left(g(\Tilde{\Omega}_{Ni}; \mathbf{f}) - g(\Omega_{Ni}; \mathbf{f}) \right)\right| \right] \quad \text{(where $\sigma_i$ are iid Rademacher r.v.)} \nonumber \\
&\le 2 \mathcal{R}_n(\mathcal{G})
\label{eq: MC-2}
\end{align}

\noindent To relate the expected Rademacher average to the empirical Rademacher average, we take a further step. Similarly to the previous application of McDiarmid’s inequality on $\eta$ \eqref{eq:MC-1}, we let $\gamma(\Omega_N) = \mathcal{R}_n(\mathcal{G}) - \hat{\mathcal{R}}_n(\mathcal{G}; \Omega_N)$. Then,
\begin{align}
&|\gamma(\Omega_N) - \gamma(\Omega_N^{'})| \nonumber \\
&= \mathcal{R}_n(\mathcal{G}) - \hat{\mathcal{R}}_n(\mathcal{G}; \Omega_N) - \mathcal{R}_n(\mathcal{G}) + \hat{\mathcal{R}}_n(\mathcal{G}; \Omega_N^{'}) \nonumber \\
&\le |\mathcal{R}_n(\mathcal{G}) - \hat{\mathcal{R}}_n(\mathcal{G}; \Omega_N) - \mathcal{R}_n(\mathcal{G}) + \hat{\mathcal{R}}_n(\mathcal{G}; \Omega_N^{'})| \nonumber \\
&= \frac{1}{N}\left| \mathbb{E}_{\sigma}\left[ \sup_{g \in \mathcal{G}} \sum_{i=1}^N \sigma_i \left(g(\Omega_{Ni}) - g(\Omega_{Ni}^{'}) \right)\right]\right| \nonumber \\
&= \frac{1}{N}\left| \mathbb{E}_{\sigma}\left[ \sup_{g \in \mathcal{G}} \sigma_k \left(g(\Omega_{Nk}) - g(\Omega_{Nk}^{'}) \right)\right]\right| \nonumber \\
&\le \frac{1}{N} \sup_{g \in \mathcal{G}} \mathbb{E}_{\sigma}\left[ \left| g(\Omega_{Nk}) - g(\Omega_{Nk}^{'}) \right|\right] \nonumber \\
&= \frac{1}{N} \sup_{\mathbf{f} \in \mathcal{f}^{|T|}} \mathbb{E}_{\sigma}\left[ \left| \frac{R_k \cdot \sum_{j=1}^T \omega_j (e^{-\lambda A_jf_j(\hat{H}_{kj})} + 1)^{-1}}{\prod_{j=1}^T \pi_{kj}} - \frac{R_k^{'} \cdot \sum_{j=1}^T \omega_j (e^{-\lambda A_j^{'} f_j(\hat{H}_{kj}^{'})} + 1)^{-1}}{\prod_{j=1}^T \pi_{kj}^{'}} \right|\right] \nonumber \\
&\le \frac{M}{N \cdot c_0^T}
\end{align}

\noindent Besides, note that $\mathbb{E} \gamma = 0$. As a result, we applied the McDiarmid’s inequality once again. By setting $\delta/2 =  exp(-\frac{c_0^{2T} \cdot 2t^2}{M^2/N})$, with probability at least $1-\delta/2$, we obtained:
\begin{equation}
    \mathcal{R}_n(\mathcal{G}) \le \hat{\mathcal{R}}_n(\mathcal{G}; \Omega_N) + \frac{M}{c_0^T} \sqrt{\frac{\log 2/\delta}{2N}}
\label{eq:MC-3}
\end{equation}

\noindent Combined the results from \eqref{eq:MC-1}, \eqref{eq: MC-2} and \eqref{eq:MC-3}, with probability at least $1-\delta$, we obtained the final lemma results:
\begin{equation}
    \sup_{\mathbf{f} \in \mathcal{F}^{|T|}} |V^{SW}_\psi(\mathbf{f}) - \hat{V}^{SW}_\psi(\mathbf{f}; \Omega_N)| \le 2\hat{\mathcal{R}}_n(\mathcal{G}; \Omega_N) + \frac{3M}{c_0^T} \sqrt{\frac{\log 2 / \delta}{2N}}
\end{equation}

\vspace{1em}
\begin{lemma}
Under the Assumption \ref{assumption:Metric-Entropy}, there exists a constant $C_1 > 0$ such that the Rademacher complexity of function class $\mathcal{G}$ is upper bounded as
\begin{equation}
\hat{\mathcal{R}}_n(\mathcal{G}) \le (\alpha^{-1} + 1) \cdot \left[\alpha C_1 \sqrt{\frac{T}{n}} \left( \frac{\lambda M}{4c_0^T}\right)^\alpha \right]^{1/(\alpha + 1)}.
\end{equation}
\label{Lemma:Metric-Entropy}
\end{lemma}

\vspace{1em}
\noindent \textbf{Proof of Lemma \ref{Lemma:Metric-Entropy}}. This Lemma aims to upper bound the empirical Rademacher average via the pre-defined metric entropy of the maximizing functional space. Let $g_1, g_2\in \mathcal{G}$, corresponding to $(f_1,...,f_T), (h_1,...,h_T) \in \mathcal{F}^{T}$. We have

\begin{align}
&\frac{1}{N} \sum_{i=1}^N \left|g_1(\Omega_{Ni}; \mathbf{f}) - g_2(\Omega_{Ni}; \mathbf{h}) \right| ^2 \nonumber \\
&= \frac{1}{N} \sum_{i=1}^N \left| \frac{R_i}{\prod \pi_i} \sum_{j=1}^T \omega_j \frac{1}{e^{-\lambda A_{ij}f_j(\hat{H}_{ij})}+1}  - \frac{R_i}{\prod \pi_i} \sum_{j=1}^T \omega_j \frac{1}{e^{-\lambda A_{ij}h_j(\hat{H}_{ij})}+1}\right|^2 \nonumber \\
&= \frac{1}{N} \sum_{i=1}^N \left| \frac{R_i}{\prod \pi_i} \sum_{j=1}^T \omega_j \left( \frac{1}{e^{-\lambda A_{ij}f_j(\hat{H}_{ij})}+1} - \frac{1}{e^{-\lambda A_{ij}h_j(\hat{H}_{ij})}+1}\right)\right|^2 \nonumber \\
&\le \frac{M^2}{N \cdot c_0^{2T}} \sum_{i=1}^N \left[ \sum_{j=1}^T \omega_j \cdot \left|\frac{1}{e^{-\lambda A_{ij}f_j(\hat{H}_{ij})}+1} - \frac{1}{e^{-\lambda A_{ij}h_j(\hat{H}_{ij})}+1}\right| \right]^2\nonumber \\
&\le \frac{M^2}{N \cdot c_0^{2T}} \sum_{i=1}^N \left[ \sum_{j=1}^T \omega_j^2 \cdot \sum_{j=1}^T \left|\frac{1}{e^{-\lambda A_{ij}f_j(\hat{H}_{ij})}+1} - \frac{1}{e^{-\lambda A_{ij}h_j(\hat{H}_{ij})}+1}\right|^2 \right] \label{step:Cauchy-Schwarz} \\
&\le \frac{M^2}{N \cdot c_0^{2T}} \sum_{i=1}^N \left[ \left(\sum_{j=1}^T \omega_j \right)^2 \cdot \sum_{j=1}^T \left|\frac{1}{e^{-\lambda A_{ij}f_j(\hat{H}_{ij})}+1} - \frac{1}{e^{-\lambda A_{ij}h_j(\hat{H}_{ij})}+1}\right|^2 \right] \nonumber \\
&\le \frac{M^2}{N \cdot c_0^{2T}} \sum_{i=1}^N \sum_{j=1}^T \left|\frac{1}{e^{-\lambda A_{ij}f_j(\hat{H}_{ij})}+1} - \frac{1}{e^{-\lambda A_{ij}h_j(\hat{H}_{ij})}+1}\right|^2  \nonumber \\
&\le \left(\frac{\lambda M}{4 c_0^T}\right)^2 \frac{1}{N}\sum_{i=1}^n \sum_{j=1}^T |f_j(\hat{H}_{ij}) - h_j(\hat{H}_{ij})|^2 \label{eq:step-MVT}
\end{align}

\noindent Note that step \eqref{step:Cauchy-Schwarz} holds with the Cauchy–Schwarz inequality, and the last step \eqref{eq:step-MVT} can be shown below by applying the mean value theorem on the surrogate logistic function $\psi(z;\lambda) = \frac{1}{e^{-\lambda \cdot z}+1}$, where $z = A_jf_j(H_j)$. Specifically, since $\psi$ is continuous w.r.t. z, there exists a value $c$ between $A_{ij}f_j(\hat{H}_{ij})$ and $A_{ij}h_j(\hat{H}_{ij})$, s.t.,
\begin{equation}
    \psi^{'}(c) = \frac{\psi(A_{ij}f_j(\hat{H}_{ij})) - \psi(A_{ij}h_j(\hat{H}_{ij}))}{{A_{ij}f_j(\hat{H}_{ij})-A_{ij}h_j(\hat{H}_{ij})}}.
\end{equation}
Then naturally, we can upper bound the surrogate function distances as,
\begin{align}
&|\psi(A_{ij}f_j(\hat{H}_{ij}); \lambda) - \psi(A_{ij}h_j(\hat{H}_{ij}); \lambda)| \nonumber \\
&= |\psi^{'}(c)||{{A_{ij}f_j(\hat{H}_{ij})-A_{ij}h_j(\hat{H}_{ij})}}| \nonumber \\
&=\left|\frac{\lambda e^{\lambda c}}{(e^{\lambda c}+1)^2} \right| \cdot |A_{ij}| \cdot |{f_j(\hat{H}_{ij})-h_j(\hat{H}_{ij})}| \nonumber \\
&\le \frac{\lambda }{4}|{f_j(\hat{H}_{ij})-h_j(\hat{H}_{ij})}| \quad \text{($\psi^{'}(c)$ reaches maximum when $c=0$)} \nonumber
\end{align}
As a result, we obtained 
\begin{equation}
\left( \psi(A_{ij}f_j(\hat{H}_{ij}); \lambda) - \psi(A_{ij}h_j(\hat{H}_{ij}); \lambda) \right)^2 \le \frac{\lambda^2}{16} |{f_j(\hat{H}_{ij})-h_j(\hat{H}_{ij})}|^2
\end{equation}

\noindent Based on the inequality results from \eqref{eq:step-MVT}, we have shown that $u$-covers on $f_j, h_j \in \mathcal{F}$ for $j=1,..,T$ w.r.t. the empirical l2-norm $\|\cdot\|_{H_{1:N}}$ define an $\frac{\lambda M}{4c_0^T}$ $u$-covers on $\mathcal{G}$ w.r.t. $\|\cdot\|_{\Omega_{1:N}}$. Thus,
\begin{equation}
\mathcal{N}_2(\frac{\lambda M}{4c_0^T} u, \mathcal{G}, \Omega_{1:N}) \le \mathcal{N}_2 (u, \mathcal{F}, H_{1:N})^T, \nonumber
\end{equation}
which could also be represented as metric entropy,
\begin{equation}
    \log \mathcal{N}_2(u, \mathcal{G}, \Omega_{1:n}) \le T \cdot \log \mathcal{N}_2 (\frac{4c_0^T}{\lambda M} u, \mathcal{F}, H_{1:n}) \le T \cdot C \left( \frac{\lambda M}{4c_0^T u}\right)^{2\alpha}
\end{equation}
Finally, based on the Discretization theorem, we could relate the empirical rademacher averages with the function metric entropy
\begin{align}
    \hat{\mathcal{R}}_n(\mathcal{G}) &\le \inf_{u>0} \left\{ u + c \sqrt{\frac{\log N_2(u, \mathcal{G}, \|\cdot\|_2)}{N}}\right\}  \nonumber  \\
    & \le \inf_{u>0} \left\{ u + c \sqrt{\frac{T \cdot C\left(\frac{\lambda M}{4c_0^T u}\right)^{2\alpha}}{N}}\right\} \nonumber \\
    &\le (\alpha^{-1} + 1) \cdot \left[\alpha C_1 \sqrt{\frac{T}{N}} \left( \frac{\lambda M}{4c_0^T}\right)^\alpha \right]^{1/(\alpha + 1)},
\end{align}
which reaches its minimum when $u = \left[\alpha C_1 \sqrt{\frac{T}{N}} \left(\frac{\lambda M}{4C_0^T} \right)^\alpha \right]^{1/(\alpha + 1)}$.

\section{K-IPWE nested treatments and carryover effects}
\label{S:nested}

The proposed K-IPWE is designed to break the \textit{curse of full-matching} by allowing treatment mismatches. According to the treatment matching number $K \doteq \sum_{t=1}^T \mathbb{I}\{A_t = D_t(H_t)\}$, it may first appear that the indicator function could be evaluated independently at each stage due to the summation operator, which seems to ignore treatment carryover effects and fail to account for the nested structure of treatments across decision stages. However, it is important to note that our treatment recommendation at each stage are based on subjects' summarized history of health variables, which effectively captures past treatments. In the following, we elaborate the evaluation process of K-IPWE and explain how our methods can incorporate carried over effects and nested treatments. 

For illustration purpose, we consider a subject who undergoes a three-stage treatment session, with their treatment trajectory observed as $(X_1, A_1,X_2,A_2,X_3,A_3,R)$. At each stage, we summarize the subjects' evolving health information into a single variable: $H_1=S_1(X_1)$, $H_2=S_2(X_1,A_1,X_2)$, and $H_3=S_3(X_1,A_1,X_2,A_2,X_3)$, using the summary functions $S_1, S_2$, and $S_3$. These summarized health variables reflect the cumulative effects of past treatments.

\subsection{Evaluation process of K-IPWEs}
The evaluation of the random treatment matching number $K$  involves a sequential assessment of each stage, from the first to the last. At each stage, we determine whether the assigned treatment aligns with the treatment regime decisions informed by all previous assignments ($A_1, \ldots, A_{t-1}$). For example, at the first stage, we compare if $A_1$ is aligned with $D_1(H_1)$. At the second stage, we take into account the treatment received in the previous stage $A_1$ and its effects on the subject's health status $X_2$. We then recommend a treatment for the second stage, $D_2(H_2) = D_2(S_2(X_1,A_1,X_2))$, and compare this with the assigned treatment, $A_2$. This process is repeated $T$ times until all treatments have been compared.

From this evaluation process of K-IPWE, we can further show how K-IPWE can improve sample efficiency and stabilize IPWEs.  Specifically, if this subject received treatments $\{A_1=1, A_2=1, A_3=1\}$, and the regime's recommendations were $\{D_1(H_1)=1, D_2(H_2)=1, D_3(H_3)=-1\}$, there would be two alignments (i.e., $K=2$). However, if the optimal decisions for this subject were $\{D_1^{*}(H_1)=1, D_2^{*}(H_2)=-1, D_3^{*}(H_3)=1\}$, there would still be two alignments, but since there is no exact match (i.e., $K < T$), this subject would be excluded from the estimating procedure of IPWEs and would not contribute to the estimation of the optimal regime. Given that the number of patients with optimal decisions across all stages is small, each collected patient data point becomes valuable. Therefore, considering $K=2<T=3$, we estimate the optimal regime from a group of regimes which have two matching stages with the actual treatment assignment. In the given example, this subject qualifies for our estimation procedure (i.e., $\sum_{t=1}^3 \mathbb{I}(A_t = D_t(H_t)) = 2$)  and the optimal regime is also qualified because two of the actual treatment assignments are assigned optimally at first and third stages  (i.e., $\sum_{t=1}^3 \mathbb{I}(A_t = D^* _t(H_t)) = 2$). This facilitates the estimation of the optimal regime using this additional instance. 

\subsection{Carryover effects from a summary of history}

Carryover effects capture the impact of a past treatment on patient's response to future treatments, which is essential in DTRs. The evaluation process of K-IPWEs above demonstrates how historical information is summarized and used to inform the treatment regime to make a new treatment recommendation for the patient at a decision stage. To provide a more concrete example, we focus on the decision to be made at the second stage.

Suppose the subject first received treatment $A_1=1$. Given the available information $\{X_1, A_1=1, X_{2, A_1=1}\}$, we summarize the up-to-date historical health information $H_2=S_2(X_1, A_1=1, X_{2, A_1=1})$ and make a treatment recommendation for the second stage, which turns out to be treatment 1, i.e., $D_2(H_2)=1$. Now, suppose the subject received treatment $A_1=-1$ instead. Similarly, we can construct the health summary $\Tilde{H}_2=S_2(X_1, A_1=-1, X_{2, A_1=-1})$, but the recommended treatment based on the updated $\Tilde{H}_2$ becomes -1, i.e., $D_2(\Tilde{H}_2)=-1$. As shown, due to the change in the treatment $A_1$ at first stage, the first-stage treatment effect is carried over to the second stage, resulting into two health status ($H_2$ v.s. $\Tilde{H}_2$). Meanwhile, our estimating regime, which is based on these summaries of historical health status, incorporates this carryover effect and dynamically update treatment recommendations accordingly.

\subsection{Nested interpretation of treatments}

The interpretation of treatments are typically considered nested in DTRs. For example, consider a subject participating in a ADHD study. At the first stage, the subject is assigned to drug treatment if $A_1=1$, or to cognitive behavioral therapy (CBT) if $A_1=-1$. At the second stage, if the subject received drug treatment at the first stage (i.e., $A_1=1$), they will then either be assigned to a new drug if $A_2=1$ or switch to CBT if $A_2=-1$. On the other hand, if the subject was initially assigned to CBT (i.e. $A_1=-1$), they may either combine their existing CBT session with a drug treatment if $A_2=1$ or add an additional CBT session if $A_2=-1$. As shown, the treatment options at second stage have the same encoding $\{-1, 1\}$, but their meaning depends on the treatment received at the first stage. 

In our estimation framework, we incorporate this nested interpretation of treatments by similarly leveraging the summary of historical health status. Specifically, when recommending future treatments, the history variable—which encapsulates all previously assigned treatments—allows for varied interpretations of the resulting recommendations. This approach enables us to effectively construct the nested structure of treatments.

\section{Generalized K-IPWE learning framework}
\label{S:KIPWL}

The proposed SAL method introduces a more general K-IPWE learning (K-IPWL) framework defined in Equation \eqref{eq:EE}. Depending on one's prior knowledge of the matching status between assigned treatments and optimal regime, this framework can be more effectively recover the optimal regime in theory compared to SAL, which assumes a linear relationship. However, there are two major optimization difficulties associated with it. First, one need to find a smooth convex surrogate for the matching number, which can be decomposed into two indicator functions, i.e., $\mathbb{I}(\lvert\mathbf{A} \cap \mathscr{D}\rvert = k) = \mathbb{I}(\sum_{t=1}^T \mathbb{I}(a_t \cdot D_t > 0) = k)$. Here, we consider a logistic function as a surrogate to the inner indicator $\mathbb{I}(x > 0)$, i.e., $\psi_1(x;\lambda)=(e^{-\lambda x}+1)^{-1}$, and a Gaussian function as a surrogate to the outer indicator $\mathbb{I}(x = 0)$, i.e., $\psi_2(x;\sigma)=e^{-x^2 / \sigma}$, where $\lambda$ and $\sigma$ are two growth-rate hyper=parameters. An illustration of the surrogates can be found below in Figures \ref{fig:surr-1} and \ref{fig:surr-2},
\begin{figure}[H]
    \centering
    \includegraphics[width=0.3\textwidth]{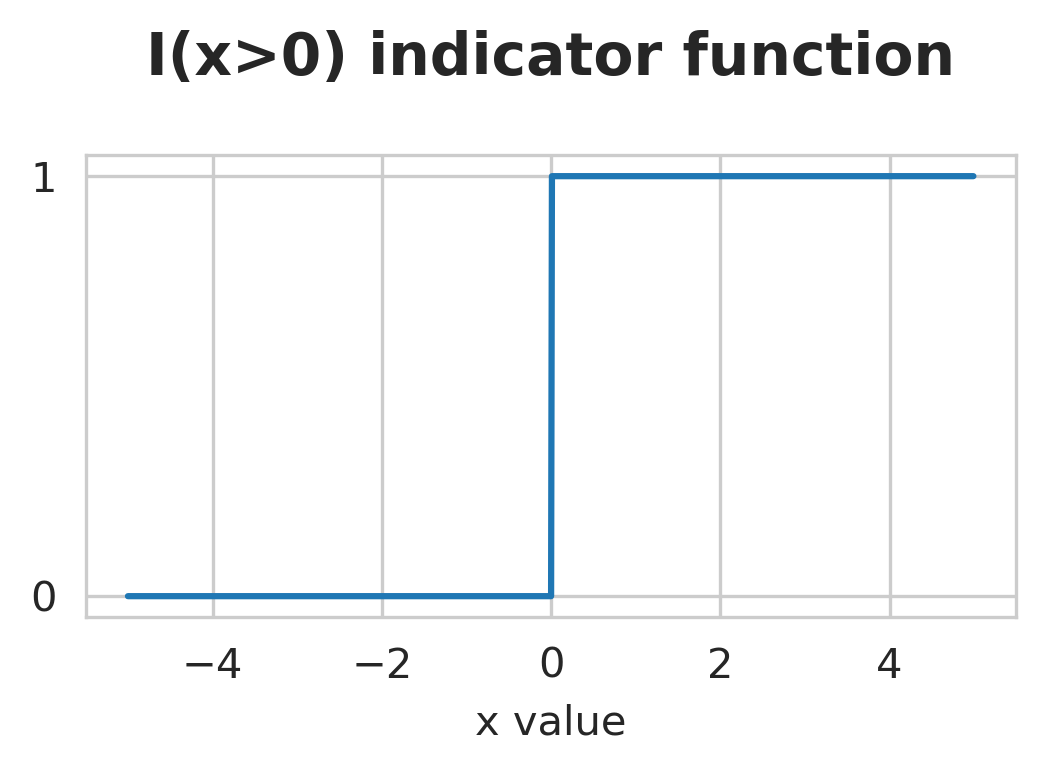}
    \includegraphics[width=0.6\textwidth]{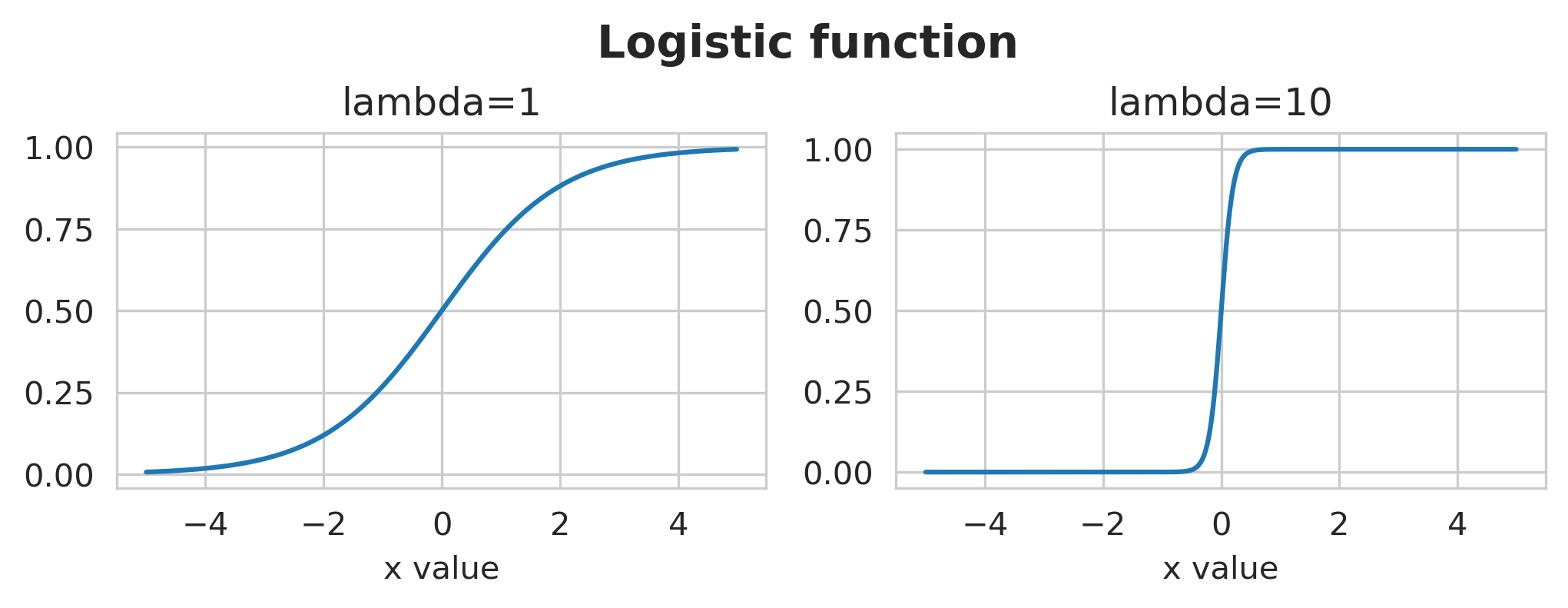}
    \caption{Logistic function as a smooth surrogate to $\mathbb{I}(x>0)$}
    \label{fig:surr-1}
\end{figure}
\begin{figure}[H]
    \centering
    \includegraphics[width=0.3\textwidth]{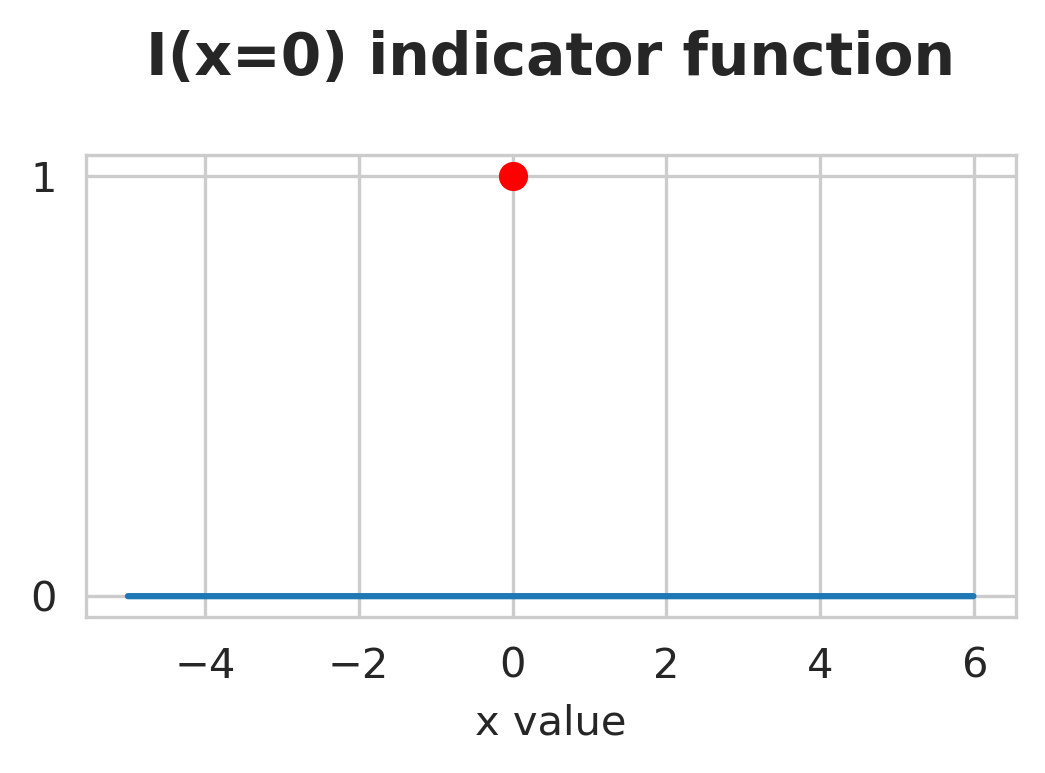}
    \includegraphics[width=0.6\textwidth]{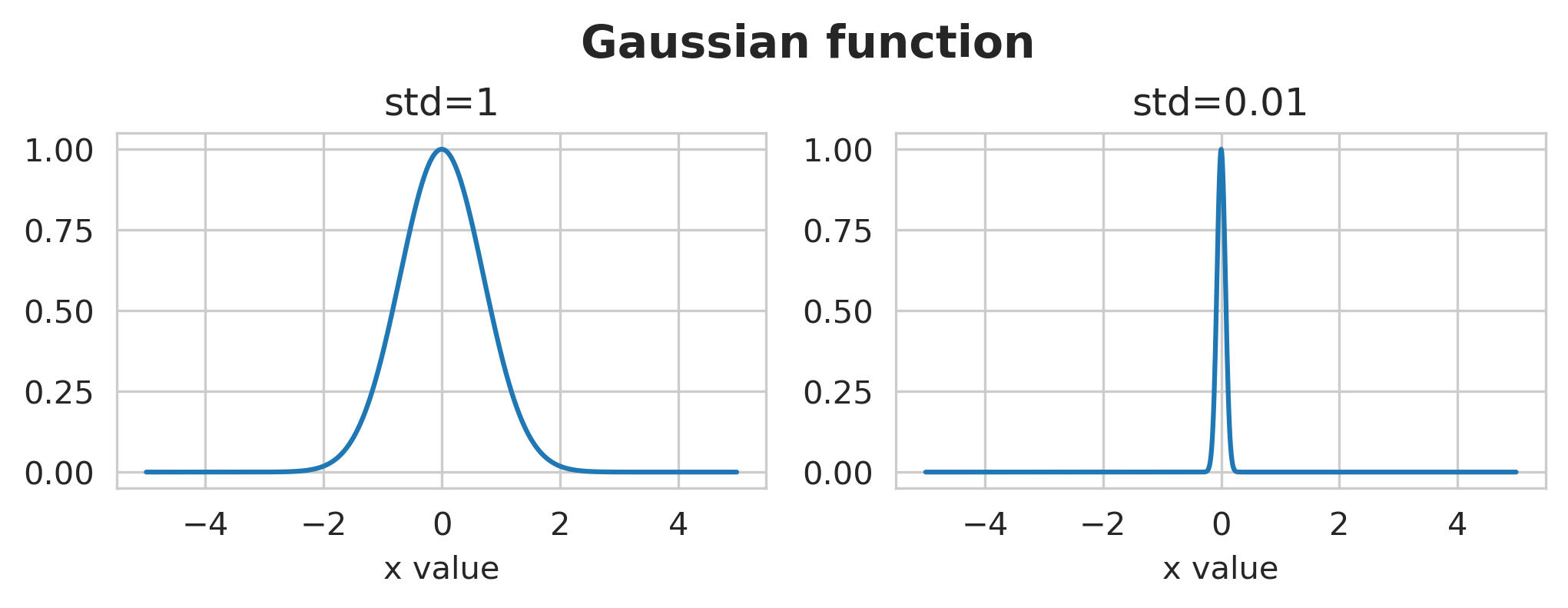}
    \caption{Gaussian function as a smooth surrogate to $\mathbb{I}(x=0)$}
    \label{fig:surr-2}
\end{figure}
\noindent With the smooth surrogates, we reformulate the objective function K-IPWE \eqref{eq:EE} into,
\begin{equation}
     \Tilde{\mathscr{D}} = \argmax_{\mathscr{D}} \mathbb{E} \left\{  \sum_{k=0}^T \phi(k) \cdot \mathbb{E}\left( \frac{R \cdot \psi_2(\sum_{t=1}^T \psi_1 (A_t \cdot D_t(H_t)) - k)}{\prod_{t=1}^T \pi_t(A_t | H_t)} \right) \right\},
\label{eq:KIPWE}
\end{equation}
which leads to an additional challenge -- determining the underlying probability, $\phi(k)$, of matching $k$ number of stages between assigned treatments and the optimal regime. However, this quantity is typically unobservable in real applications. To address this, we propose first applying existing DTR methods, such as Q-learning or BOWL, to estimate the optimal regime and then obtain empirical matching probabilities between the estimated regime and the observed assignments. These empirical probabilities serve as approximations for $\phi(k)$, completing the final component of the value function in Equation \eqref{eq:KIPWE} and allows the K-IPWE learning framework to be optimized via standard gradient descent methods.

In the following, we present simulation results for K-IPWL in Table \ref{table: phi} from the same setting specified in Section \ref{sim::6-2}. Notice that, apart from the procedure described above, we also incorporate the ground-true $\phi(\cdot)$, which are known from the simulation design, to estimate K-IPWL. optimization, and demonstrates robustness in practice.

\begin{table}[hbt!]
    \centering
    \bgroup
    \def\arraystretch{1.3}%
    \resizebox{1\textwidth}{!}{
\begin{tabular}{|c|ccccc|cccc|cc|c|}
\hline
$\phi(K)$ & BOWL & Q-learning & AIPW & RWL & C-learning & \begin{tabular}{@{}c@{}} \textbf{K-IPWL} \\ (Pred. $\phi$)\end{tabular}  & \begin{tabular}{@{}c@{}} \textbf{K-IPWL} \\ (True $\phi$)\end{tabular} & \textbf{SAL} & \textbf{SWL} & Observed & Oracle & \begin{tabular}{@{}c@{}} \textbf{Imp-rate} \\ (to BOWL) \end{tabular} \\ \hline
Binomial (0.7) & \begin{tabular}{@{}c@{}} 0.578 \\ (0.022) \end{tabular} & \begin{tabular}{@{}c@{}} 0.165 \\ (0.070) \end{tabular} & \begin{tabular}{@{}c@{}} 0.222 \\ (0.027) \end{tabular} & \begin{tabular}{@{}c@{}} 0.547 \\ (0.024) \end{tabular} & \begin{tabular}{@{}c@{}} 0.524 \\ (0.027) \end{tabular} & \begin{tabular}{@{}c@{}} 0.648 \\ (0.060) \end{tabular} & \begin{tabular}{@{}c@{}} 0.620 \\ (0.019) \end{tabular} & \begin{tabular}{@{}c@{}} 0.657 \\ (0.020) \end{tabular} & \begin{tabular}{@{}c@{}} \textbf{0.658}  \\ (0.021) \end{tabular} & \begin{tabular}{@{}c@{}} 0.400 \\ (0.011) \end{tabular} & \begin{tabular}{@{}c@{}} 0.999 \\ (0.009) \end{tabular} & 13.780\% \\ \hline
Binomial (0.6) & \begin{tabular}{@{}c@{}} 0.455 \\ (0.034) \end{tabular} & \begin{tabular}{@{}c@{}} 0.146 \\ (0.075) \end{tabular} & \begin{tabular}{@{}c@{}} 0.218 \\ (0.028) \end{tabular} & \begin{tabular}{@{}c@{}} 0.428 \\ (0.028) \end{tabular} & \begin{tabular}{@{}c@{}} 0.530 \\ (0.023) \end{tabular} & \begin{tabular}{@{}c@{}} 0.620 \\ (0.056) \end{tabular} & \begin{tabular}{@{}c@{}} 0.602 \\ (0.046) \end{tabular} & \begin{tabular}{@{}c@{}} \textbf{0.635}  \\ (0.022) \end{tabular} & \begin{tabular}{@{}c@{}} 0.633 \\ (0.023) \end{tabular} & \begin{tabular}{@{}c@{}} 0.200 \\ (0.010) \end{tabular} & \begin{tabular}{@{}c@{}} 0.999 \\ (0.009) \end{tabular} & 39.603\% \\ \hline
Binomial (0.5) & \begin{tabular}{@{}c@{}} 0.347 \\ (0.026) \end{tabular} & \begin{tabular}{@{}c@{}} 0.115 \\ (0.072) \end{tabular} & \begin{tabular}{@{}c@{}} 0.213 \\ (0.027) \end{tabular} & \begin{tabular}{@{}c@{}} 0.382 \\ (0.034) \end{tabular} & \begin{tabular}{@{}c@{}} 0.518 \\ (0.026) \end{tabular} & \begin{tabular}{@{}c@{}} 0.311 \\ (0.031) \end{tabular} & \begin{tabular}{@{}c@{}} 0.308 \\ (0.034) \end{tabular} & \begin{tabular}{@{}c@{}} \textbf{0.624}  \\ (0.021) \end{tabular} & \begin{tabular}{@{}c@{}} 0.624 \\ (0.021) \end{tabular} & \begin{tabular}{@{}c@{}} -0.001 \\ (0.012) \end{tabular} & \begin{tabular}{@{}c@{}} 0.999 \\ (0.009) \end{tabular} & 79.694\% \\ 
\hline 
\end{tabular}
}
\egroup
\caption{\small  Estimated total rewards when $N=5000$, $T=10$, the optimal regime is nonlinear, treatment matching $\sum_{t=1}^T \mathbb{I}(A_t= d^*_t) \sim \phi(\cdot)$ and no important stage. Standard errors are listed next to the estimated means. The Oracle stands for the best estimated total rewards if all treatments are assigned optimally.}
\label{table: phi}
\end{table}

According to the results, while K-IPWL outperforms competing methods in most cases, there remains a noticeable performance gap compared to SAL/SWL, primarily due to the complexity of the nested surrogate functions. This gap persists even when the ground-truth matching distribution, $\phi(\cdot)$, is provided, which effectively reduces variance and enhances performance compared to its empirically estimated counterpart. Additionally, in the Binomial case with $p=0.5$, K-IPWL experiences a significant performance loss. Possible reason is that the symmetry of the Binomial distribution at $p=0.5$, where, for example, matching 1 and $T-1$ stages carries the same weight, makes it irrelevant whether a recommended decision shall be aligned with the optimal regime. These numerical results collectively indicate that K-IPWL is estimable and the linear assumption utilized by SAL encourages identifying more optimal decisions across the treatment stages, improves sample efficiency, simplifies optimization, and demonstrates robustness in practice.

\section{Simulation} \label{S:simulation}

In this appendix, we further compare our proposed methods with different DTR methods, specifically non-parametric regression models, infinite-horizon framework, Fisher-consistent surrogate estimators, and shared-parameter DTRs. Supportive model performance results are provided in each comparison.

\subsection{Non-parametric outcome model}
\label{S:non-parametric}

One of the major requirements of regression-based methods is the correct specification of the outcome models. Among the existing literature, non-parametric approaches  \citep{NPQL-1, NPQL-2, zhao2009reinforcement, zhang2015using, zhang2018c, zhang2018estimation} have been proposed to alleviate the model specification challenges. In this simulation, we implement the regression-based outcome component of C-learning with the non-parametric random forest. The results are summarized as follows in Table \ref{table: non-parametric}.

Notably, the non-parametric outcome model can indeed improve model performance and is aligned with the findings presented in \cite{taylor2015reader}. However, there still exists a noticeable performance gap between the non-parametric C-learning and our proposed methods. This suggests that the additional computational burden from the outcome models could deteriorate model performance. In contrast, our methods, which retain the simplicity of IPWEs, directly utilizes the observed total rewards, and fundamentally address the \textit{curse of full-matching}, have substantially improved sample efficiency and brought optimization convenience, leading to superior model performance

\begin{table}[H]
    \centering
    \bgroup
    \def\arraystretch{1.5}%
    \resizebox{1\textwidth}{!}{
\begin{tabular}{|l|l|cccccc|cc|cc|c|}
\hline 
N & T  & BOWL & Q-learning & AIPWE  & RWL &  \begin{tabular}{@{}c@{}} C-learning \\ (Parametric)\end{tabular} & \begin{tabular}{@{}c@{}} C-learning \\ (Non-Parametric)\end{tabular} & \textbf{SAL} & \textbf{SWL}  &  Observed & Oracle & \begin{tabular}{@{}c@{}} \textbf{Imp-rate} \\ (To Best)\end{tabular}\\
\hline
5000 & 5 & 0.157 (0.063) & 0.580 (0.030) & 0.380 (0.040) & 0.611 (0.024) & 0.604 (0.024) & 0.606 (0.029) & \textbf{0.673} (0.025) & 0.671 (0.026) & 0.000 (0.017) & 0.999 (0.007) & 10.081\% \\
 & 8 & 0.131 (0.069) & 0.420 (0.037) & 0.266 (0.028) & 0.466 (0.026) & 0.550 (0.025) & 0.552 (0.027) & \textbf{0.647} (0.019) & \textbf{0.647} (0.019) & 0.000 (0.012) & 1.000 (0.008) & 17.551\% \\
 & 10 & 0.115 (0.072) & 0.347 (0.026) & 0.213 (0.027) & 0.382 (0.034) & 0.518 (0.026) & 0.524 (0.027) & \textbf{0.624} (0.021) & \textbf{0.624} (0.021) & -0.001 (0.012) & 0.999 (0.009) & 20.541\% \\
\hline
1000 & 5 & 0.145 (0.064) & 0.418 (0.050) & 0.215 (0.046) & 0.463 (0.044) & 0.421 (0.049) & 0.432 (0.046) & \textbf{0.533} (0.045) & \textbf{0.533} (0.045) & -0.008 (0.035) & 0.995 (0.015) & 14.979\% \\
 & 8 & 0.112 (0.070) & 0.264 (0.050) & 0.156 (0.038) & 0.314 (0.047) & 0.341 (0.053) & 0.364 (0.043) & \textbf{0.459} (0.037) & \textbf{0.459} (0.037) & -0.009 (0.032) & 1.001 (0.013) & 34.660\% \\
 & 10 & 0.072 (0.051) & 0.214 (0.038) & 0.124 (0.037) & 0.247 (0.041) & 0.308 (0.046) & 0.316 (0.043) & 0.438 (0.041) & \textbf{0.440} (0.042) & -0.012 (0.028) & 1.000 (0.016) & 42.969\% \\
\hline
500 & 5 & 0.162 (0.061) & 0.316 (0.068) & 0.148 (0.067) & 0.363 (0.054) & 0.336 (0.075) & 0.350 (0.069) & 0.464 (0.063) & \textbf{0.465} (0.063) & 0.008 (0.054) & 0.995 (0.021) & 27.924\% \\
 & 8 & 0.130 (0.057) & 0.209 (0.053) & 0.117 (0.049) & 0.247 (0.057) & 0.262 (0.055) & 0.294 (0.062) & \textbf{0.410} (0.062) & 0.409 (0.062) & 0.011 (0.045) & 1.001 (0.020) & 56.281\% \\
 & 10 & 0.100 (0.068) & 0.157 (0.062) & 0.089 (0.048) & 0.199 (0.052) & 0.224 (0.067) & 0.244 (0.049) & \textbf{0.378} (0.063) & 0.376 (0.064) & 0.013 (0.041) & 1.000 (0.027) & 68.250\% \\
\hline
\end{tabular}
}
\egroup
\caption{\small Estimated total rewards when the optimal regime is nonlinear (heterogeneous), assigned treatment $A_t \sim \text{Bernoulli(0.5)} \cdot d^*_t$ and no important stage. Standard errors are listed next to the estimated means. The Oracle stands for the best estimated total rewards if all treatments are assigned optimally. The improvement rate compares SAL/SWL against the best competing methods.}
\label{table: non-parametric}
\end{table}

\subsection{Infinite-horizon DTRs}
\label{S:sim-infinite}
Our proposed methods demonstrate superior performance, especially when dealing with a large number of decision stages. However, as the number of stages continues to increase, infinite-horizon methods may become a viable option. In this simulation, we explore this possibility for scenarios with a large number of stages.

Specifically, consistent with the generation process specified in Section \ref{sim:6-1}, we extend our simulation settings to include scenarios with up to 30 stages. Additionally, we implement the deep Q-network (DQN) \citep{mnih2013playing} as a classic infinite-horizon algorithm. The results are summarized in Table \ref{table: infinite-horizon}.

\begin{table}[hbt!]
    \centering
    \bgroup
    \def\arraystretch{1.3}%
    \resizebox{1\textwidth}{!}{
\begin{tabular}{|c|cccccc|cc|cc|c|}
\hline 
T  & BOWL & Q-learning  & AIPW  & RWL  & C-learning & DQN & SAL & SWL &  Observed & Oracle & \begin{tabular}{@{}c@{}} \textbf{Imp-rate} \\ (To Best)\end{tabular}\\
\hline
 10 & 0.214 (0.038) & 0.072 (0.051) & 0.124 (0.037) & 0.247 (0.041) & 0.316 (0.043) & 0.246 (0.049) & 0.438 (0.041) & \textbf{0.440}  (0.042) & -0.012 (0.028) & 1.000 (0.016) & 39.243\% \\ \hline
15 & 0.159 (0.041) & 0.041 (0.059) & 0.095 (0.032) & 0.182 (0.037) & 0.279 (0.038) & 0.194 (0.047) & 0.403 (0.034) & \textbf{0.403}  (0.034) & -0.011 (0.018) & 1.003 (0.013) & 44.448\% \\ \hline
20 & 0.132 (0.032) & 0.035 (0.040) & 0.075 (0.027) & 0.139 (0.028) & 0.241 (0.035) & 0.170 (0.048) & 0.385 (0.038) & \textbf{0.386}  (0.037) & -0.009 (0.019) & 1.001 (0.013) & 60.341\% \\ \hline
30 & 0.082 (0.032) & -0.013 (0.048) & 0.038 (0.022) & 0.085 (0.040) & 0.190 (0.030) & 0.097 (0.048) & \textbf{0.341}  (0.030) & \textbf{0.341}  (0.030) & -0.011 (0.016) & 0.993 (0.010) & 78.958\% \\ 
\hline
\end{tabular}
}
\egroup
\caption{\small Estimated total rewards when $N=1000$, the optimal regime is nonlinear, assigned treatment $A_t \sim \text{Bernoulli(0.5)} \cdot d^*_t$ and no important stage. Standard errors are listed next to the estimated means. The Oracle stands for the best estimated total rewards if all treatments are assigned optimally. The improvement rate compares SAL/SWL against the best competing methods.}
\label{table: infinite-horizon}
\end{table}

As shown, infinite-horizon DQN can outperform Q-learning, IPWE (BOWL), and several of its robust variants, including AIPW and RWL, in long-stage settings which these methods are not designed for. However, DQN starts to underperform compared to stronger baseline methods such as C-learning. One key reason is that the infinite-horizon literature commonly assumes the Markov decision process (MDP), which states that the future state $S^{t+1}$ only depends on the current state $S^{t}$ and the action taken $A^{t}$, not on the sequence of past states and actions (i.e., $S^{t+1} \perp (S^1, A^1,...,S^{t-1}, A^{t-1}) \;|\; (S^t, A^t)$). However, in a finite-horizon setting, future health variables depend on all past treatments, and the full treatment trajectory can be used to model temporal dependencies among health variables, making the MDP assumption less suitable.

 In contrast, our proposed methods summarize the entire treatment history, consistently achieving the best performance scores. Although the expected total rewards for all methods decline as the number of stages increases, our methods show a growing improvement margin, considering the level of heterogeneity involved over many stages. Lastly, in real-case scenarios, having 30 stages is sufficiently large in a finite-horizon setting. Further increasing the number of stages may fundamentally change the nature of the problem and one probably should proceed with the infinite-horizon methods instead.

\subsection{Fisher-consistent surrogate models}
\label{S:sim-fisher}

BOWL is a pioneering work which views IPWE as a weighted classification error and employs surrogate methods from large-margin classification to estimate the optimal DTR. However, as noted in later work \citep{laha2024finding}, the surrogate approach used by BOWL might not achieve Fisher consistency. In response, \cite{laha2024finding} discusses a class of Fisher consistent surrogate functions. In this simulation, we select the function $\phi(x)= 1 + 2/\pi \cdot \arctan (\pi x / 2)$ (DTRESLO) to examine whether using a Fisher-consistent surrogate can improve BOWL's performance and how it compares to our proposed method.

\begin{table}[hbt!]
    \centering
    \bgroup
    \def\arraystretch{1.3}%
    \resizebox{0.9\textwidth}{!}{
\begin{tabular}{|c|cccc|cc|cc|c|}
\hline
T &      BOWL & SOWL & DTRESLO & C-learning & \textbf{SAL} & \textbf{SWL} &   Observed & Oracle & \begin{tabular}{@{}c@{}} \textbf{Imp-rate} \\ (To Best)\end{tabular}\\
\hline
5 & 0.418 (0.050) & 0.010 (0.067) & 0.416 (0.055) & 0.423 (0.055) & \textbf{0.533}  (0.045) & \textbf{0.533}  (0.045) & -0.008 (0.035) & 0.995 (0.015) & 26.082\% \\
8 & 0.264 (0.050) & 0.015 (0.055) & 0.324 (0.041) & 0.354 (0.053) & \textbf{0.459}  (0.037) & \textbf{0.459}  (0.037) & -0.009 (0.032) & 1.001 (0.013) & 29.600\% \\
10 & 0.214 (0.038) & 0.001 (0.035) & 0.273 (0.039) & 0.316 (0.043) & 0.438 (0.041) & \textbf{0.440}  (0.042) & -0.012 (0.028) & 1.000 (0.016) & 39.243\% \\
\hline
\end{tabular}
}
\egroup
\caption{\small Estimated total rewards when $n=1000$, the optimal regime is nonlinear, assigned treatment $A_t \sim \text{Bernoulli(0.5)} \cdot d^*_t$ and no important stage. Standard errors are listed next to the estimated means. The Oracle stands for the best estimated total rewards if all treatments are assigned optimally. The improvement rate compares SAL/SWL against the best performer of competing methods.}
\label{table: Leha}
\end{table}

Based on the simulation results in Table \ref{table: Leha}, we observe that DTRESLO improves upon BOWL's performance due to the use of a Fisher-consistent estimator. However, our methods, SAL and SWL, still achieve the highest performance, with the improvement rate increasing as the number of stages grows. This indicates that, while DTRESLO benefits from a Fisher consistent surrogate function, it is still an IPWE-based method and thus affected by the \textit{curse of full-matching}. Since our methods are specifically designed to overcome this limitation, we achieve the best performance scores.

\subsection{DTRs with shared parameters}
\label{S:sim-shared}

There are scenarios where the treatment decisions exhibit similarities across different stages. In these cases, DTRs with shared parameters seem to be a preferable option as they reduce the number of estimators and better align with the underlying decision process. In this simulation, we investigate the robustness of proposed method against shared-parameter DTRs when the underlying treatment rules differ at every stage (heterogeneous) or are shared and remain the same across all stages (homogeneous). 

Specifically, we choose the shared Q-learning method  \citep{chakraborty2016estimating} for demonstration purpose. In addition, to mimic the shared-parameter DTR methods, we enforce DTRESLO, SAL, and SWL to use the same set of parameters for their decision rules across all stages. This can be viewed as a special case of the shared-parameter DTR method where the shared parameter set equals the entire parameter set. The results are summarized in Table \ref{table: shared-parameter}.

\begin{table}[hbt!]
    \centering
    \bgroup
    \def\arraystretch{1.3}%
    \resizebox{0.8\textwidth}{!}{
\begin{tabular}{|l|l|l|ll|ll|l|}
\hline 
 \begin{tabular}{@{}c@{}} Underlying \\ decision rule \end{tabular} &  \begin{tabular}{@{}c@{}} Model \\ parameters \end{tabular} &  T   & Q-learning &  DTRESLO & \textbf{SAL} & \textbf{SWL}   & \begin{tabular}{@{}c@{}} \textbf{Imp-rate} \\ (To Best)\end{tabular}\\
\hline
Heterogeneous & Un-shared & 5 & 57.155 (3.429) & 71.025 (2.600) & \textbf{76.845}  (2.064) & \textbf{76.845}  (2.064) & \textbf{8.194\%} \\
 &  & 8 & 55.230 (3.238) & 66.150 (2.037) & \textbf{72.898}  (1.755) & \textbf{72.898}  (1.755) & \textbf{10.202\%} \\
 &  & 10 & 53.258 (2.469) & 63.590 (1.738) & 71.861 (1.908) & \textbf{71.959}  (1.957) & \textbf{13.160\%} \\
\cline{2-8}
 & Shared & 5 & 65.480 (2.712) & 67.582 (2.962) & \textbf{70.882}  (2.339) & 70.733 (2.771) & 4.883\% \\
 &  & 8 & 63.019 (2.654) & 63.323 (3.217) & 69.114 (2.291) & \textbf{69.184}  (2.911) & 9.256\% \\
 &  & 10 & 61.224 (2.853) & 60.794 (3.664) & \textbf{68.816}  (2.072) & 68.157 (2.298) & 12.401\% \\
\hline
Homogeneous & Un-shared & 5 & 54.960 (3.732) & 70.388 (2.663) & \textbf{76.175}  (2.934) & \textbf{76.175}  (2.934) & 8.222\% \\
 &  & 8 & 54.400 (3.096) & 64.952 (2.307) & \textbf{71.848}  (2.950) & \textbf{71.848}  (2.950) & 10.618\% \\
 &  & 10 & 52.898 (3.574) & 63.424 (1.804) & \textbf{70.964}  (2.682) & \textbf{70.964}  (2.682) & 11.888\% \\
\cline{2-8}
& Shared & 5 & 75.305 (3.686) & 73.957 (3.827) & \textbf{82.650}  (2.595) & 82.445 (2.769) & \textbf{9.754\%} \\
 &  & 8 & 72.341 (4.329) & 65.537 (4.871) & \textbf{81.373}  (3.376) & 80.964 (3.165) & \textbf{12.486\%} \\
 &  & 10 & 69.398 (4.426) & 62.378 (5.212) & \textbf{80.050}  (3.441) & 80.001 (3.365) & \textbf{15.350\%} \\
\hline
\end{tabular}
}
\egroup
\caption{\small Matching accuracy with the optimal regime when $N=1000$, the optimal regime is nonlinear, assigned treatment $A_t \sim \text{Bernoulli(0.5)} \cdot d^*_t$ and no important stage. Standard errors are listed next to the estimated means. The Oracle stands for the best estimated total rewards if all treatments are assigned optimally. The improvement rate compares SAL/SWL against the best performer of competing methods.}
\label{table: shared-parameter}
\end{table}

Based on the simulation results summarized in Table \ref{table: shared-parameter}, we notice that our model outperforms the others in these scenarios and the improvement margin increases with the number of stages, which demonstrates the estimation efficiency of our proposed method. Additionally, we show that correctly aligning the model's shared property with the underlying decision rule (i.e., heterogeneous-unshared and homogeneous-shared) can effectively enhance model performances across DTRESLO, SAL, and SWL. In the case of shared Q-learning, while it consistently outperforms regular un-shared Q-learning, shared Q-learning demonstrates even greater performance gain when the underlying decision rule is homogeneous. 

In conclusion, DTRs with shared parameters indeed can improve model performance when the underlying treatment decisions are similar. From the simulations, we show that our methods are still able to reach the highest empirical performance in a shared-decision rule setting.  In addition, our method demonstrates the potential of enabling the shared parameters properties, and we will leave the extensions of allowing varying parameters to be shared in future endeavors.


\section{Real data application} \label{S:real-data}

The raw COVID-19 data from UC CORDS is imbalanced and inappropriate for direct application of DTR methods. To properly formulate the problem under the framework of our proposed model, we need to find the total rewards, determine the treatment stages, and pre-process patients' covariates and assignments at each stage. First of all, we consider the reverse-scaled number of days stayed at hospitals and ICU as the total rewards for each inpatient. For instance, a patient, who stays the longest at hospitals, will receive 0 total reward, but s/he would have the largest total reward if s/he spent the least number of days at hospitals. Next, since the actual $j^{th}$ drug administration timing could vary among patients due to the observational data, we choose a consistent time space for each individual where the meaning of $j^{th}$ treatment timestamp is transformed into the $j^{th}$ number of treatment visits. For instance, at stage 1, all patients have been made the first decision whether to take a drug or not, but the timining for the decisions could vary among individuals. Then, under the newly determined time-space, the treatment stages are consistent across individuals, but the length of time intervals between two treatment stages can still be different, To resolve the inconsistency of time intervals, we applied kernel smoothing to project the recorded patients' covariates information during each time interval onto the consistent time-space, i.e., the $j^{th}$ treatment stage. In particular, we select a total of 38 covariates, covering patients' demographic features, basic vital measurements, blood test results, bio-markers, and comorbidity history, to parameterize the treatment regime. A descriptive summary of the selected covariates can be found in Table \ref{tab:descriptive_statistics}. In the last step of data preprocessing, we impute the missing covariate values by random forests \citep{tang2017random}.

\begin{table}[hbt!]
    \centering
        \bgroup
    \def\arraystretch{1.05}%
    \resizebox{0.55\textwidth}{!}{
    \begin{tabular}{lllllll}
        \hline
        & Count & Mean & Median & Min & Max & S.D. \\
        \hline
        \textbf{Demographics} & & & & & & \\ \hline
        Female & 2333 & 0.421 & 0 & 0 & 1 & 0.494 \\
        Race:Caucasian & 2333 & 0.396 & 0 & 0 & 1 & 0.489 \\
        Race:Asian & 2333 & 0.108 & 0 & 0 & 1 & 0.311 \\
        Race: Hispanic/Latino & 2333 & 0.412 & 0 & 0 & 1 & 0.492 \\
        Age & 2333 & 53.18 & 57.00 & 1 & 87  & 22.97 \\
        BMI & 2333 & 35.22 & 31.59 & 5 & 50 & 12.35 \\ \hline
    \textbf{Comorbidities} & & & & & & \\ \hline
    Diabetes  & 2333 & 0.403 & 0 & 0 & 1 & 0.491 \\
    Hypertension & 2333 & 0.649 & 1 & 0 & 1 & 0.491 \\
    Asthma & 2333 & 0.159 & 0 & 0 & 1 & 0.366 \\
    Obesity & 2333 & 0.604 & 1 & 0 & 1 & 0.489 \\
    Coronary artery disease  & 2333 & 0.269 & 0 & 0 & 1 & 0.443 \\
    Cardiovascular diseases & 2333 & 0.076 & 0 & 0 & 1 & 0.266 \\
    Chronic kidney disease & 2333 & 0.323 & 0 & 0 & 1 & 0.468 \\ \hline
    \textbf{Basic measurements} & & & & & & \\ \hline
    Heart rate & 2290 & 89.15 & 86.00 & 44.71& 179.7 & 20.55 \\
    Body temperature & 2290 & 36.83 & 36.78 & 31.58 & 40.44 & 0.760 \\
    Oxygen saturation & 2291 & 96.06 & 96.97 & 56.00 & 100.0 & 3.500 \\
    Respiratory rate & 2266 & 20.82 & 18.74 & 8.110 & 97.00 & 6.240  \\
    Diastolic blood pressure & 2281 & 72.09 & 71.75 & 27.00 & 131.1 & 12.55 \\
    Systolic blood pressure & 2281 & 123.6 & 122.0 & 93.52 & 160.1 & 15.97 \\ \hline
    \textbf{Blood tests} & & & & & & \\ \hline
    Albumin & 1700 & 3.470 & 3.600 & 0.450 & 5.400 & 0.770  \\
    Alkaline phosphates & 2093 & 112.2 & 84.00 & 18.50 & 2411 & 112.8  \\
    Aspartate aminotransferase & 2090 & 43.84 & 34.00 & 14.00 & 194.0 & 31.04  \\
    Bilirubin ($10^3$)& 2086 & 1.210 & 0.500 & 0.100 & 2.200 & 0.379  \\ 
    Calcium ($10^3$) & 2170 & 8.84 & 8.800 & 5.400 & 19.10 & 0.860  \\
    Carbon dioxide & 1781 & 22.78 & 23.00 & 15.25 & 29.70 & 3.197  \\
    Erythrocytes & 2169 & 4.140 & 4.210 & 1.010 & 8.610 & 0.880  \\
    Glucose & 2166 & 0.150 & 0.120 & 0.030 & 1.210 & 0.090  \\ 
    Lymphocytes leukocytes & 1405 & 15.32 & 12.70 & 1.50 & 52.10 & 10.74  \\
    MCHC & 2169 & 32.84 & 32.95 & 25.40 & 37.00 & 1.490  \\
    Neutrophils leukocytes & 1778 & 73.75 & 76.30 & 32.00 & 95.20 & 13.66  \\
    Potassium & 2127 & 4.056 & 4.10 & 2.365 & 5,400 & 0.590  \\
    Protein & 2018 & 6.737 & 6.800 & 5.033 & 8.700 & 0.727  \\
    Urea nitrogen ($10^3$) & 2172 & 26.27 & 19.00 & 45.00 & 131.5 & 20.70  \\ \hline
    \textbf{Biomarkers} & & & & & & \\ \hline
    C-reactive protein & 1334 & 3.986 & 0.012 & 0.0003 & 92.78 & 14.41  \\
    Creatinine ($10^3$) & 2006 & 2.597 & 0.980 & 0.400 & 47.30 & 5.87  \\
    Lactate dehydrogenase  & 585 & 490.1 & 394.0 & 139.0 & 1264 & 202.5  \\
    Platelets & 2125 & 221.6 & 212.0 & 35.00 & 544.0 & 94.55  \\
    Troponin-i & 1066 & 9.260 & 0.110 & 0.025 & 129.0 & 20.31 \\ \hline
    \end{tabular}
}
\egroup
    \caption{Summary statistics of all obtained covariates for ICU patients from UC CORDS datasets}
    \label{tab:descriptive_statistics}
\end{table}

\end{document}